\documentclass[robotics,perspective,accept,moreauthors,pdftex]{Definitions/mdpi}

\usepackage[normalem]{ulem} % For using the command \uuline{} in the \abstract 
\usepackage{amsfonts} % Use for some special math mark
\usepackage{wasysym} % Use for some special mark
\usepackage{mathcomp} % For permille mark
\usepackage{CJKutf8} % For Chinese font
\usepackage{pifont} % For some special mark 
\usepackage{bm} % Forr math enviroment bold format
\usepackage{bbm} % For some math fancy characters
\graphicspath{{./Definitions/}} % Use for import path the figures paper used

\makeatletter
\def\T@n@@nc@d@ngM@cr@M@d{}
\def\LY@n@@nc@d@ngM@cr@M@d{}
\makeatother

\let\orignewcommand\newcommand  % Store the original \newcommand
\let\newcommand\providecommand  % Make \newcommand behave like \providecommand
\usepackage{verse}
\let\newcommand\orignewcommand  % Use the original `\newcommand` in future

\newsavebox\foobox

\renewcommand{\dagger}{\mathchar"2279}
\renewcommand{\ddagger}{\mathchar"227A}

\newcommand{\mmathit}[1]{
  \ifthenelse{\equal{#1}{\ln}}{\mathit{ln}}{
    \ifthenelse{\equal{#1}{\max}}{\mathit{max}}{\mathit{#1}}
  }
}
\makeatother
\robustify{\footnote} % I don't know what it is for? Maybe for Chicago journal footnote?
  
\DeclareUnicodeCharacter{1E45}{\.{n}}
\DeclareUnicodeCharacter{1E41}{\.{m}}
\DeclareUnicodeCharacter{2003}{\quad}
\DeclareUnicodeCharacter{2009}{\thinspace}
\DeclareUnicodeCharacter{2002}{\enspace{}}
\DeclareUnicodeCharacter{2005}{\thinspace}
\DeclareUnicodeCharacter{0263}{\textipa{G}}
\DeclareUnicodeCharacter{A0}{~}
\DeclareUnicodeCharacter{2460}{\textcircled{\scriptsize{1}}}
\DeclareUnicodeCharacter{2461}{\textcircled{\scriptsize{2}}}
\DeclareUnicodeCharacter{2462}{\textcircled{\scriptsize{3}}}
\DeclareUnicodeCharacter{2463}{\textcircled{\scriptsize{4}}}
\DeclareUnicodeCharacter{2464}{\textcircled{\scriptsize{5}}}
\DeclareUnicodeCharacter{2465}{\textcircled{\scriptsize{6}}}
\DeclareUnicodeCharacter{2466}{\textcircled{\scriptsize{7}}}
\DeclareUnicodeCharacter{2467}{\textcircled{\scriptsize{8}}}
\DeclareUnicodeCharacter{2468}{\textcircled{\scriptsize{9}}}
\DeclareUnicodeCharacter{2070}{\textsuperscript{0}}
\DeclareUnicodeCharacter{2074}{\textsuperscript{4}}
\DeclareUnicodeCharacter{2075}{\textsuperscript{5}}
\DeclareUnicodeCharacter{2076}{\textsuperscript{6}}
\DeclareUnicodeCharacter{2077}{\textsuperscript{7}}
\DeclareUnicodeCharacter{2078}{\textsuperscript{8}}
\DeclareUnicodeCharacter{2079}{\textsuperscript{9}}
\DeclareUnicodeCharacter{02C2}{<}
\DeclareUnicodeCharacter{2033}{\relax\ifmmode '' \else $''$\fi}
\DeclareUnicodeCharacter{2034}{\relax\ifmmode ''' \else $'''$\fi}
\DeclareUnicodeCharacter{2026}{\relax\ifmmode … \else $\ldots$\fi}
\DeclareUnicodeCharacter{0229}{\c{e}}
\DeclareUnicodeCharacter{016F}{\r{u}}
\DeclareUnicodeCharacter{127}{\relax\ifmmode\rm\hbar\else $\rm\hbar$\fi}
\DeclareUnicodeCharacter{3AC}{\relax\ifmmode\acute{\alpha}\else $\acute{\alpha}$\fi}
\DeclareUnicodeCharacter{3AD}{\relax\ifmmode\acute{\varepsilon}\else $\acute{\varepsilon}$\fi}
\DeclareUnicodeCharacter{3AE}{\relax\ifmmode\acute{\eta}\else $\acute{\eta}$\fi}
\DeclareUnicodeCharacter{3AF}{\relax\ifmmode\acute{\iota}\else $\acute{\iota}$\fi}
\DeclareUnicodeCharacter{3CC}{\relax\ifmmode\acute{o}\else $\acute{o}$\fi}
\DeclareUnicodeCharacter{3CD}{\relax\ifmmode\acute{\upsilon}\else $\acute{\upsilon}$\fi}
\DeclareUnicodeCharacter{3CE}{\relax\ifmmode\acute{\omega}\else $\acute{\omega}$\fi}
\DeclareUnicodeCharacter{391}{A}
\DeclareUnicodeCharacter{392}{B}
\DeclareUnicodeCharacter{395}{E}
\DeclareUnicodeCharacter{396}{Z}
\DeclareUnicodeCharacter{397}{H}
\DeclareUnicodeCharacter{399}{I}
\DeclareUnicodeCharacter{39A}{K}
\DeclareUnicodeCharacter{39C}{M}
\DeclareUnicodeCharacter{39D}{N}
\DeclareUnicodeCharacter{39F}{O}
\DeclareUnicodeCharacter{3A1}{P}
\DeclareUnicodeCharacter{3A4}{T}
\DeclareUnicodeCharacter{3A7}{X}

\DeclareUnicodeCharacter{27E6}{\relax\ifmmode \llbracket \else $\llbracket$\fi}
\DeclareUnicodeCharacter{27E7}{\relax\ifmmode \rrbracket \else $\rrbracket$\fi}

\DeclareUnicodeCharacter{1D434}{\relax\ifmmode A \else $A$\fi}
\DeclareUnicodeCharacter{1D435}{\relax\ifmmode B \else $B$\fi}
\DeclareUnicodeCharacter{1D436}{\relax\ifmmode C \else $C$\fi}
\DeclareUnicodeCharacter{1D437}{\relax\ifmmode D \else $D$\fi}
\DeclareUnicodeCharacter{1D438}{\relax\ifmmode E \else $E$\fi}
\DeclareUnicodeCharacter{1D439}{\relax\ifmmode F \else $F$\fi}
\DeclareUnicodeCharacter{1D43A}{\relax\ifmmode G \else $G$\fi}
\DeclareUnicodeCharacter{1D43B}{\relax\ifmmode H \else $H$\fi}
\DeclareUnicodeCharacter{1D43C}{\relax\ifmmode I \else $I$\fi}
\DeclareUnicodeCharacter{1D43D}{\relax\ifmmode J \else $J$\fi}
\DeclareUnicodeCharacter{1D43E}{\relax\ifmmode K \else $K$\fi}
\DeclareUnicodeCharacter{1D43F}{\relax\ifmmode L \else $L$\fi}
\DeclareUnicodeCharacter{1D440}{\relax\ifmmode M \else $M$\fi}
\DeclareUnicodeCharacter{1D441}{\relax\ifmmode N \else $N$\fi}
\DeclareUnicodeCharacter{1D442}{\relax\ifmmode O \else $O$\fi}
\DeclareUnicodeCharacter{1D443}{\relax\ifmmode P \else $P$\fi}
\DeclareUnicodeCharacter{1D444}{\relax\ifmmode Q \else $Q$\fi}
\DeclareUnicodeCharacter{1D445}{\relax\ifmmode R \else $R$\fi}
\DeclareUnicodeCharacter{1D446}{\relax\ifmmode S \else $S$\fi}
\DeclareUnicodeCharacter{1D447}{\relax\ifmmode T \else $T$\fi}
\DeclareUnicodeCharacter{1D448}{\relax\ifmmode U \else $U$\fi}
\DeclareUnicodeCharacter{1D449}{\relax\ifmmode V \else $V$\fi}
\DeclareUnicodeCharacter{1D44A}{\relax\ifmmode W \else $W$\fi}
\DeclareUnicodeCharacter{1D44B}{\relax\ifmmode X \else $X$\fi}
\DeclareUnicodeCharacter{1D44C}{\relax\ifmmode Y \else $Y$\fi}
\DeclareUnicodeCharacter{1D44D}{\relax\ifmmode Z \else $Z$\fi}
\DeclareUnicodeCharacter{1D44E}{\relax\ifmmode a \else $a$\fi}
\DeclareUnicodeCharacter{1D44F}{\relax\ifmmode b \else $b$\fi}
\DeclareUnicodeCharacter{1D450}{\relax\ifmmode c \else $c$\fi}
\DeclareUnicodeCharacter{1D451}{\relax\ifmmode d \else $d$\fi}
\DeclareUnicodeCharacter{1D452}{\relax\ifmmode e \else $e$\fi}
\DeclareUnicodeCharacter{1D453}{\relax\ifmmode f \else $f$\fi}
\DeclareUnicodeCharacter{1D454}{\relax\ifmmode g \else $g$\fi}
\DeclareUnicodeCharacter{1D456}{\relax\ifmmode i \else $i$\fi}
\DeclareUnicodeCharacter{1D457}{\relax\ifmmode j \else $j$\fi}
\DeclareUnicodeCharacter{1D458}{\relax\ifmmode k \else $k$\fi}
\DeclareUnicodeCharacter{1D459}{\relax\ifmmode l \else $l$\fi}
\DeclareUnicodeCharacter{1D45A}{\relax\ifmmode m \else $m$\fi}
\DeclareUnicodeCharacter{1D45B}{\relax\ifmmode n \else $n$\fi}
\DeclareUnicodeCharacter{1D45C}{\relax\ifmmode o \else $o$\fi}
\DeclareUnicodeCharacter{1D45D}{\relax\ifmmode p \else $p$\fi}
\DeclareUnicodeCharacter{1D45E}{\relax\ifmmode q \else $q$\fi}
\DeclareUnicodeCharacter{1D45F}{\relax\ifmmode r \else $r$\fi}
\DeclareUnicodeCharacter{1D460}{\relax\ifmmode s \else $s$\fi}
\DeclareUnicodeCharacter{1D461}{\relax\ifmmode t \else $t$\fi}
\DeclareUnicodeCharacter{1D462}{\relax\ifmmode u \else $u$\fi}
\DeclareUnicodeCharacter{1D463}{\relax\ifmmode v \else $v$\fi}
\DeclareUnicodeCharacter{1D464}{\relax\ifmmode w \else $w$\fi}
\DeclareUnicodeCharacter{1D465}{\relax\ifmmode x \else $x$\fi}
\DeclareUnicodeCharacter{1D466}{\relax\ifmmode y \else $y$\fi}
\DeclareUnicodeCharacter{1D467}{\relax\ifmmode z \else $z$\fi}

\DeclareUnicodeCharacter{1E67}{\.{\v s}}
\DeclareUnicodeCharacter{1E11}{\relax\ifmmode \c{d} \else $\c{d}$\fi}
\DeclareUnicodeCharacter{1ECB}{\relax\ifmmode \d{i} \else $\d{i}$\fi}
\DeclareUnicodeCharacter{1D8D}{\relax\ifmmode \textlhookx \else $\textlhookx$\fi}
\DeclareUnicodeCharacter{104}{\relax\ifmmode \k{A} \else $\k{A}$\fi}
\DeclareUnicodeCharacter{211E}{\relax\ifmmode \textrecipe \else $\textrecipe$\fi}
\DeclareUnicodeCharacter{29D}{\relax\ifmmode \textctj \else $\textctj$\fi}

\DeclareUnicodeCharacter{1E2E}{\'{\"I}}
\DeclareUnicodeCharacter{23F}{\textrts}

\DeclareUnicodeCharacter{2C73}{\varw}

\DeclareUnicodeCharacter{2127}{\mho}

\DeclareUnicodeCharacter{28C}{\textturnv}
\DeclareUnicodeCharacter{252}{\textturnscripta}
\DeclareUnicodeCharacter{259}{\schwa}
\DeclareUnicodeCharacter{25B}{\m{e}}
\DeclareUnicodeCharacter{266}{\m{h}}
\DeclareUnicodeCharacter{127}{\B{h}}
\DeclareUnicodeCharacter{27E}{\textfishhookr}
\DeclareUnicodeCharacter{281}{\textinvscr}

\firstpage{1}
  \articlenumber{112}
\pubvolume{13}
\issuenum{8}
\pubyear{2024}
\copyrightyear{2024}
\externaleditor{Academic Editor: Po-Yen~Chen}
\datereceived{30 May 2024}
\daterevised{6 July 2024}
\dateaccepted{18 July 2024}
\datepublished{23 July 2024}
\hreflink{https://doi.org/10.3390/\linebreak robotics13080112}%please use \linebreak command if need to do line break
\usepackage[OT1,OT2,T2A,T2B,T2C,T3,T5,T1]{fontenc}
\usepackage[russian,english]{babel}

\Title{The Future of Intelligent Healthcare: A Systematic Analysis and Discussion on the Integration and Impact of Robots Using Large Language Models for Healthcare}

\TitleCitation{The Future of Intelligent Healthcare: A Systematic Analysis and Discussion on the Integration and Impact of Robots Using Large Language Models for Healthcare}

\Author{Souren~Pashangpour~\textsuperscript{1}\textsuperscript{,}*\href{https://orcid.org/0009-0006-9672-1911}{\orcidicon} and Goldie~Nejat~\textsuperscript{1}\textsuperscript{,}\textsuperscript{2}\textsuperscript{,}\textsuperscript{3}\textsuperscript{,}*\href{https://orcid.org/0000-0002-7080-6857}{\orcidicon}}

\AuthorNames{Souren Pashangpour and Goldie Nejat}

\AuthorCitation{Pashangpour, S.; Nejat, G.}

\address{\textsuperscript{1} \quad Autonomous Systems and Biomechatronics Laboratory~(ASBLab), Department of Mechanical and Industrial Engineering, University of Toronto, Toronto, ON M5S 3G8, Canada

\textsuperscript{2} \quad KITE, Toronto Rehabilitation Institute, University Health Newtork~(UHN), Toronto, ON M5G 2A2, Canada

\textsuperscript{3} \quad Rotman Research Institute, Baycrest Health Sciences, North York, ON M6A 2E1, Canada}

\corres{Correspondence: souren.pashangpour@mail.utoronto.ca (S.P.); nejat@mie.utoronto.ca (G.N.)}

\abstract{The potential use of large language models (LLMs) in healthcare robotics can help address the significant demand put on healthcare systems around the world with respect to an aging demographic and a shortage of healthcare professionals. Even though LLMs have already been integrated into medicine to assist both clinicians and patients, the integration of LLMs within healthcare robots has not yet been explored for clinical settings. In this perspective paper, we investigate the groundbreaking developments in robotics and LLMs to uniquely identify the needed system requirements for designing health-specific LLM-based robots in terms of multi-modal communication through human--robot interactions (HRIs), semantic reasoning, and task planning. Furthermore, we discuss the ethical issues, open challenges, and potential future research directions for this emerging innovative field.}


\makeatletter
\DeclareRobustCommand*\textsubscript[1]{%
  \@textsubscript{\selectfont#1}}
\def\@textsubscript#1{%
  {\m@th\ensuremath{_{\mbox{\fontsize\sf@size\z@#1}}}}}
\makeatother

\usepackage{sansmath}

\begin{document}
\section{Introduction \label{sect:sec1-robotics-3060247}}

In healthcare, the need for new technology to maintain the quality and efficiency of care is paramount. This demand has been amplified by an increase in the overall older population of the world. Namely, by 2050, 22\% of the global population is expected to be over 65 years old~\cite{B1-robotics-3060247}. This demographic shift leads to a rising prevalence of chronic diseases such as dementia, diabetes, and heart disease, which require continuous monitoring and long-term management, further straining healthcare resources~\cite{B2-robotics-3060247,B3-robotics-3060247}. In 2022, Canada alone had 143,695 job vacancies for healthcare professionals~\cite{B4-robotics-3060247}. In the U.S., it is predicted that by 2026 there will be a shortage of up to 3.2 million healthcare workers~\cite{B5-robotics-3060247}, highlighting the staggering workforce shortage~\cite{B5-robotics-3060247}. Furthermore, the vast amounts of health data generated in this sector require efficient management and use to improve patient outcomes and reduce healthcare costs, a task well suited for generative AI and deep learning models~\cite{B6-robotics-3060247,B7-robotics-3060247,B8-robotics-3060247}. For example, generative AI models, such as large language models (LLMs), can use the large EHR (Electronic Health Record) datasets for training to learn patterns between symptoms, diagnoses, and recommendations in order to help in healthcare with the management of data and the retrieval of information, as well as decision-making processes~\cite{B9-robotics-3060247}.

The need for new technologies in healthcare is multi-faceted, using the following:\linebreak (1) classical and deep learning methods to facilitate medical imaging analysis~\cite{B10-robotics-3060247}, (2) deep neural networks (DNNs) to perform automated disease detection and prediction~\cite{B11-robotics-3060247}, and (3) LLMs for clinical decision-making and teleconsultation~\cite{B12-robotics-3060247}. In particular, LLMs have already been integrated into medicine to assist with clinical note-taking by rewriting and summarizing clinicians’ notes for clarity and have also been leveraged through chatbots to assist patients with reminders and general questions about medications~\cite{B12-robotics-3060247}. Furthermore, LLMs have also been used to personalize treatment plans for individual patients by assisting with EHR data extraction~\cite{B13-robotics-3060247}. The use of deep learning and LLMs have many benefits and advancements in the delivery of healthcare.

In addition to the aforementioned software technologies, innovations in robotics have allowed their integration into healthcare, including the following: (1) surgical robots being used to perform various minimally invasive surgical procedures for brain, spine, and neck surgeries, where accuracy and reliability are paramount~\cite{B14-robotics-3060247,B15-robotics-3060247}, (2) rehabilitation robots in the form of robotic assistive wheelchairs~\cite{B16-robotics-3060247}, prostheses~\cite{B17-robotics-3060247}, and robotic arms and exoskeletons for lower and upper limb rehab~\cite{B18-robotics-3060247,B19-robotics-3060247}, (3) mobile medication delivery robots utilized as automated medication carts~\cite{B20-robotics-3060247}, and (4) humanoid robots monitoring patient vital signs~\cite{B21-robotics-3060247}. The need for robots to assist in healthcare was especially evident during the COVID-19 pandemic, where the healthcare sector turned to robotics in response to public health emergencies for various tasks to minimize person-to-person contact and the spread of the virus~\cite{B22-robotics-3060247}. This included autonomous disinfecting robots using UV-C irradiation for surface decontamination in hospitals~\cite{B23-robotics-3060247} and service robots used to facilitate social distancing measures in hospitals and long-term care homes by conducting initial screening of COVID-19 symptoms and detecting face masks~\cite{B23-robotics-3060247,B24-robotics-3060247,B25-robotics-3060247,B26-robotics-3060247}. Social robots have also been used to provide companionship to older adults in healthcare environments (i.e., long-term homes) by engaging in conversations, such as sharing stories or telling jokes, thus creating a more personal and comforting presence by interacting with users~\cite{B27-robotics-3060247,B28-robotics-3060247}. In general, the integration of robots into healthcare aims to enhance patient experiences and outcomes, support skill augmentation, and improve the overall quality of care while reducing the workload of care providers~\cite{B29-robotics-3060247}.

Despite the use of both healthcare robots and LLMs in healthcare, the integration and deployment of these two technologies still remains unexplored. To the authors’ knowledge, there have only been three instances where robots and LLMs have been applied directly for healthcare applications: two instances involving social robots~\cite{B30-robotics-3060247,B31-robotics-3060247} and one instance involving a surgical robot~\cite{B32-robotics-3060247}. However, this potential marriage of robotics and LLMs for healthcare presents an untapped opportunity, as the combination of a robot’s physical capabilities with the understanding and generative abilities of LLMs has the potential to provide person-centered care, as well as streamline operational workflows (e.g., logistical tasks), and reduce the workload of healthcare professionals. This integration will result in the utilization of vast healthcare data to further refine diagnostic, therapeutic, and predictive healthcare services. The potential of healthcare robots using LLMs to provide such services underlines the urgent need for research and development in this emerging~area.

In this paper, we present the first investigation into the emerging field of healthcare robots using LLMs. Namely, we explore the innovative developments which aim to address the challenge of enhancing the quality and efficiency of patient care during a time where the demographic is shifting towards an older population and there is a shortage of healthcare professionals. Our objective is to identify the needed system requirements for multi-modal communication through human--robot interactions (HRIs), systematic reasoning, and task planning for designing health-specific LLM-based robotic solutions. We will also discuss the ethical issues associated with the potential development and utilization of healthcare robots leveraging LLMs and the open challenges and potential future research directions for this field.

\section{Large Language Models (LLMs) for Healthcare \label{sect:sec2-robotics-3060247}}

The versatility of generative AI is apparent in its ability to be trained on an array of data types from textual (i.e., EHRs)~\cite{B33-robotics-3060247} and visual content (i.e., medical imaging data)~\cite{B34-robotics-3060247} to genetic sequences~\cite{B29-robotics-3060247} in order to learn and capture the underlying patterns and distributions from such data. This makes it especially valuable for healthcare tasks that require adaptability and continuous learning~\cite{B30-robotics-3060247}. LLMs represent a significant advancement in generative AI. They primarily utilize the transformer architecture~\cite{B35-robotics-3060247} and contain key features including multi-head attention for parallel processing, positional encodings for sequence awareness, layer normalization, and feedforward networks for data refinement~\cite{B36-robotics-3060247}. The encoder--decoder structure in many LLMs facilitates complex tasks such as language translation and content generation~\cite{B37-robotics-3060247}. The transformer architecture provides LLMs with the potential to be instrumental in healthcare robotic development due to the proficiency in generating human-like text~\cite{B38-robotics-3060247}, understanding of the lexical semantics of the physical world~\cite{B36-robotics-3060247,B39-robotics-3060247}, and making decisions regarding appropriate robot behaviors to implement~\cite{B40-robotics-3060247}.

\tabref{tabref:robotics-3060247-t001} presents an overview of five specialized LLMs designed for healthcare:\linebreak (1) BiomedBERT~\cite{B41-robotics-3060247}, (2) MED-PaLM 2~\cite{B42-robotics-3060247}, (3) DRG-LLaMA~\cite{B43-robotics-3060247}, (4) GPT2-BioPt~\cite{B44-robotics-3060247}, and (5) Clinical-T5~\cite{B45-robotics-3060247}. It also consists of seven general LLMs which have been incorporated into robotic tasks for HRIs, semantic reasoning, and planning: (1) GPT-3~\cite{B46-robotics-3060247}, \mbox{(2) GPT-3.5~\cite{B46-robotics-3060247},} (3) GPT-4~\cite{B47-robotics-3060247}, (4) T5~\cite{B48-robotics-3060247}, (5) DialogLED~\cite{B49-robotics-3060247}, (6) PaLM~\cite{B50-robotics-3060247}, and (7)~the Stanford Alpaca~\cite{B51-robotics-3060247}. Furthermore, \tabref{tabref:robotics-3060247-t001} includes two popular general-use LLMs:\linebreak (1) BERT~\cite{B52-robotics-3060247} and (2) PaLM 2~\cite{B53-robotics-3060247}. To date, the specialized LLMs have not yet been deployed in robotics for healthcare applications. However, they have the potential to be applied to tasks such as diagnostic assistance, patient data analysis, and treatment and procedure recommendations. These specialized LLMs are created by fine-tuning the model parameters of their respective base models, namely, BERT~\cite{B52-robotics-3060247}, PaLM~\cite{B50-robotics-3060247}, PaLM 2~\cite{B53-robotics-3060247}, LLaMA 2~\cite{B54-robotics-3060247}, GPT2~\cite{B55-robotics-3060247}, and T5~\cite{B48-robotics-3060247}, on healthcare datasets such as MedQA~\cite{B56-robotics-3060247}, MedMCQA~\cite{B57-robotics-3060247}, and PubMedQA~\cite{B58-robotics-3060247}. It is important to note that these base models are considered foundational models, which can be either open-source or closed-source. Open-source models are those whose underlying code and training procedures are freely available for use, modification, and distribution by anyone, facilitating transparency and collaboration in model development~\cite{B59-robotics-3060247}. In contrast, closed-source models are proprietary, with their code, methodologies, and data often kept confidential by the organization that developed them, limiting access and modification by external entities~\cite{B60-robotics-3060247}. This distinction is crucial, as it influences the breadth of application and customization potential of each model in healthcare settings. Moreover, these base models are built from the ground up, having unique architectures, and they are trained on expertly curated datasets to give them general knowledge about the world~\cite{B61-robotics-3060247}. The training data size is often reported in gigabytes, while the context length (prompt length) is reported as tokens which represent the variable numbers of characters depending on the tokenization process used. The size and the source of the training data (i.e., chat forums, scholarly journals) are what gives breadth (general knowledge) to the model, while the context length dictates the size of the input to the model and constitutes the context for what the model generates~\cite{B62-robotics-3060247,B63-robotics-3060247}. The general LLMs reported in \tabref{tabref:robotics-3060247-t001} are the models that will be discussed in the following sections in this paper with respect to their current use in robotics and their potential implications for healthcare. We will present two design studies in \sect{sect:sec8-robotics-3060247} showcasing how these three types of LLMs can be potentially used for healthcare robotics applications.

There are several common prompting techniques that are essential for the interaction between robotic systems and LLMs which can also be extended to healthcare settings. The prompts serve as the conduit through which queries or tasks are communicated to LLMs through agents (people or robots), with the model generating responses based on the given prompt structure. Outputs from LLMs are probabilistic, leading to ‘prompt engineering’, where different prompt structures are used to bias the output of the LLM towards a desired outcome~\cite{B64-robotics-3060247}. Moreover, the context window of an LLM refers to the contiguous sequence of tokens considered by the LLM in a forward pass (i.e., generation time)~\cite{B65-robotics-3060247}. This window, typically limited by memory constraints, determines the amount of preceding and succeeding text data the model utilizes to maintain coherence and accuracy in tasks such as word prediction~\cite{B66-robotics-3060247}. This is evident in few-shot prompting, where input--output pairs provided to the model fill its context window with consistent structures, reducing variations in the input text~\cite{B67-robotics-3060247,B68-robotics-3060247}. This uniformity in the context window influences the attention mechanism to focus more narrowly, enhancing the relevance of certain parts of the input to the generated output~\cite{B69-robotics-3060247,B70-robotics-3060247}. Therefore, the effectiveness of a prompt and, consequently, the utility of the model’s response are highly contingent upon the prompt’s design. A prompting method is critical for achieving specific objectives. \tabref{tabref:robotics-3060247-t002} describes key prompting strategies, detailing their operational frameworks, exemplary applications we have identified within healthcare, and their respective advantages and limitations. It is worth noting that prompts can be multi-modal, including text, images, or audio. In order for LLMs to respond to a variety of informational inputs~\cite{B71-robotics-3060247}, the non-textual components are converted into textual descriptions through preprocessing models such as CLIP which aligns images and text in a joint embedding space~\cite{B72-robotics-3060247} or SCANREF which aligns point clouds and sentences in a joint embedding space~\cite{B73-robotics-3060247}. This multi-modal integration broadens the scope of LLM applications, enhancing their adaptability and effectiveness in complex healthcare environments where diverse data types are prevalent, such as medical imaging data, recorded conversations between patients and healthcare professionals, and dedicatory images such as maps of healthcare environments or medication labels. The prompting techniques presented in \tabref{tabref:robotics-3060247-t002} integrate LLMs into robotics, as discussed in the literature in the subsequent sections. The example prompts provided by us in \tabref{tabref:robotics-3060247-t002} highlight their potential applications in healthcare.

    \begin{table}[H]
    \tablesize{\footnotesize}
    \caption{LLMs with potential applications in healthcare.}
    \label{tabref:robotics-3060247-t001}

\begin{adjustwidth}{-\extralength}{0cm}
%\centering %% If there is a figure in wide page, please release command \centering
\setlength{\cellWidtha}{\fulllength/6-2\tabcolsep-0in}
\setlength{\cellWidthb}{\fulllength/6-2\tabcolsep-0.2in}
\setlength{\cellWidthc}{\fulllength/6-2\tabcolsep-0.2in}
\setlength{\cellWidthd}{\fulllength/6-2\tabcolsep--0.8in}
\setlength{\cellWidthe}{\fulllength/6-2\tabcolsep-0.2in}
\setlength{\cellWidthf}{\fulllength/6-2\tabcolsep-0.2in}
\scalebox{1}[1]{\begin{tabularx}{\fulllength}{>{\centering\arraybackslash}m{\cellWidtha}>{\centering\arraybackslash}m{\cellWidthb}>{\centering\arraybackslash}m{\cellWidthc}>{\centering\arraybackslash}m{\cellWidthd}>{\centering\arraybackslash}m{\cellWidthe}>{\centering\arraybackslash}m{\cellWidthf}}
\toprule

\textbf{Model Name} & \textbf{Parameter Size} & \textbf{Context Length} & \textbf{Type of Training Data} & \textbf{Foundational Model} & \textbf{Papers That Use the Model}\\
\cmidrule{1-6}

MED-PaLM-2 {\cite{B42-robotics-3060247}} & 540 B & 8196 & MedQA, MedMCQA, HealthSearchQA, LiveQA, and MedicationQA  & PaLM 2 {\cite{B53-robotics-3060247}} &  \\
\cmidrule{1-6}
DRG-LLaMA {\cite{B43-robotics-3060247}} & 7 B, 13 B, 70 B & 4096 & 236,192 MIMIC-IV discharge summaries & LLaMa 2 {\cite{B54-robotics-3060247}} &  \\
\cmidrule{1-6}
GPT2-BioPT {\cite{B44-robotics-3060247}}  & 124 M, 770 M,  & 1024 & PorTuguese-2 with biomedical literature {\cite{B44-robotics-3060247}} & GPT-2 {\cite{B55-robotics-3060247}} &  \\
\cmidrule{1-6}
Clinical-T5 {\cite{B45-robotics-3060247}} & 220 M, 770 M, 3 B, 11 B & Variable, memory-constrained  & Approximately 2 million textual notes from MIMIC-III & T5 {\cite{B48-robotics-3060247}} &  \\
\cmidrule{1-6}
BiomedBERT {\cite{B41-robotics-3060247}} & 110 M, 340 M & 512 & BREATHE containing research articles and abstracts from different sources (\emph{BMJ}, arXiv, medRxiv, bioRxiv, CORD-19, Springer Nature, NCBI, \emph{JAMA}, and BioASQ) & BERT {\cite{B52-robotics-3060247}} &  \\
\cmidrule{1-6}
BERT {\cite{B52-robotics-3060247}} & 110 M, 340 M & 512 & BookCorpus (800 M words) and\linebreak  English Wikipedia (2500 M words) & NA &  \\
\cmidrule{1-6}
T5 {\cite{B48-robotics-3060247}} & 60 M, 220 M, 770 M, 3 B, 11 B & Variable Length & 750 GB of Colossal Clean Crawled Corpus (C4) & NA & {\cite{B31-robotics-3060247,B74-robotics-3060247,B75-robotics-3060247}}\\
\cmidrule{1-6}
GPT-3 {\cite{B46-robotics-3060247}} & 175 B  & 4096 & Not provided  & NA & {\cite{B76-robotics-3060247,B77-robotics-3060247,B78-robotics-3060247,B79-robotics-3060247,B80-robotics-3060247}}\\
\cmidrule{1-6}
GPT-3.5 {\cite{B46-robotics-3060247}} & 175 B & 4096 & Not provided  & NA & {\cite{B31-robotics-3060247,B32-robotics-3060247,B81-robotics-3060247,B82-robotics-3060247,B83-robotics-3060247,B84-robotics-3060247}}\\
\cmidrule{1-6}
GPT-4 {\cite{B47-robotics-3060247}} & 1.8 T & 128,000 & Not provided & NA & {\cite{B74-robotics-3060247,B85-robotics-3060247,B86-robotics-3060247}}\\
\cmidrule{1-6}
PaLM {\cite{B50-robotics-3060247}} & 8 B, 62 B, 540 B & 2048 & Social media conversations (multilingual): 50\%;\linebreak  filtered webpages (multilingual): 27\%;\linebreak books (English): 13\%;\linebreak GitHub (code): 5\%;\linebreak Wikipedia (multilingual): 4\%;\linebreak news (English): 1\% & NA & {\cite{B87-robotics-3060247}}\\
\cmidrule{1-6}
PaLM 2 {\cite{B53-robotics-3060247}} & Not \linebreak available & Not available & Not available & NA &  \\
\cmidrule{1-6}
Stanford Alpaca {\cite{B51-robotics-3060247}} & 7 B & 4096 & A mix of publicly available online data and synthetic data generated by GPT-3 & NA & {\cite{B88-robotics-3060247}}\\
\cmidrule{1-6}
DialogLED {\cite{B49-robotics-3060247}} & 41 M & 4096 & Books, English Wikipedia, real news, and stories & NA & {\cite{B31-robotics-3060247}}\\

\bottomrule
\end{tabularx}}

\end{adjustwidth}
    
    \end{table}
    \vspace{-6pt}
    
    \begin{table}[H]
    \tablesize{\scriptsize}
    \caption{Prompting techniques for LLMs for healthcare.}
    \label{tabref:robotics-3060247-t002}

\begin{adjustwidth}{-\extralength}{0cm}
%\centering %% If there is a figure in wide page, please release command \centering
\setlength{\cellWidtha}{\fulllength/6-2\tabcolsep-0in}
\setlength{\cellWidthb}{\fulllength/6-2\tabcolsep-0in}
\setlength{\cellWidthc}{\fulllength/6-2\tabcolsep-0in}
\setlength{\cellWidthd}{\fulllength/6-2\tabcolsep-0in}
\setlength{\cellWidthe}{\fulllength/6-2\tabcolsep-0in}
\setlength{\cellWidthf}{\fulllength/6-2\tabcolsep-0in}
\scalebox{1}[1]{\begin{tabularx}{\fulllength}{>{\centering\arraybackslash}m{\cellWidtha}>{\centering\arraybackslash}m{\cellWidthb}>{\centering\arraybackslash}m{\cellWidthc}>{\centering\arraybackslash}m{\cellWidthd}>{\centering\arraybackslash}m{\cellWidthe}>{\centering\arraybackslash}m{\cellWidthf}}
\toprule

\textbf{Prompting Method} & \textbf{Prompt Description} & \textbf{Example Prompt} & \textbf{Advantages} & \textbf{Disadvantages} & \textbf{Papers That Use Method}\\
\cmidrule{1-6}

Direct questioning & Direct questioning about a topic of interest. & ‘What are the primary symptoms of Type 2 diabetes?’ & Straightforward, clear, and easy to understand. Effective for factual inquiries. & May not elicit detailed or nuanced responses; limited to the user’s knowledge to ask the right questions. &  \\
\cmidrule{1-6}
Chain of thought & A problem is presented, followed by a step-by-step reasoning process to solve it. & ‘To determine the Body Mass Index (BMI), first divide the weight in kilograms by the height in meters squared.’ & Breaks down complex processes into understandable steps; useful for teaching and clarification. This approach can help the model in complex problem-solving tasks. & Can be time-consuming; requires accurate initial logic to be effective. & {\cite{B75-robotics-3060247,B86-robotics-3060247,B87-robotics-3060247}}\\
\cmidrule{1-6}
Zero-shot & Providing little to no context (zero-shot) to guide the LLM on how to respond or what format to follow. & ‘Describe the process of cellular respiration in human cells.’ & Tests the model’s ability to respond based on its pre-trained knowledge. & Responses may lack context or specificity; dependent on the model’s existing knowledge. & {\cite{B31-robotics-3060247,B74-robotics-3060247,B75-robotics-3060247,B81-robotics-3060247,B84-robotics-3060247,B85-robotics-3060247}}\\
\cmidrule{1-6}
Few-shot learning & Giving a few examples (few-shot) to guide the LLM on how to respond or what format to follow. & ‘{[}Example 1: ‘An apple is a fruit that can help with digestion’.{]}\linebreak {[}Example 2: ‘A treadmill is a device used for physical exercise’.{]}\linebreak What is an ultrasound?’ & Provides context through examples; improves the accuracy and relevance of responses. & The quality of the response depends on the quality of the examples provided. & {\cite{B31-robotics-3060247,B32-robotics-3060247,B76-robotics-3060247}}\\
\cmidrule{1-6}
EmotionPrompt {\cite{B89-robotics-3060247}} & Incorporating emotional cues to prompts and/or asking the LLM to emphasize emotion stimulus in its output. & ‘It’s crucial for my family’s well-being. Can you provide advice on maintaining a balanced diet for heart health?’ & Result in more engaging and less generic LLM outputs. & Overexaggeration in emotional stimuli and indication to excessive gestures. & {\cite{B82-robotics-3060247,B83-robotics-3060247,B86-robotics-3060247}}\\
\cmidrule{1-6}
Multi-modal prompting & Incorporating more than just text in the prompts, like images or data, especially in models that can process multi-modal inputs. This is useful for tasks that require interpretation across different types of information. & ‘Here is an MRI image of a knee. Can you explain the common injuries indicated by this type of scan?’ & Incorporates different data types for a more holistic understanding; useful for diagnostics and treatment planning. & Requires LLM models capable of processing and interpreting multiple data modes effectively. & {\cite{B77-robotics-3060247,B85-robotics-3060247}}\\
\cmidrule{1-6}
Task-oriented prompting & Combination of previous prompting methods with the addition of primitive robot actions and feedback from the robot and its operating environment, as well user request. & ‘Your role is to generate robotic plans in a X embodied robot capable of \textless{}primitive actions: moveTO (location, grasp (Object), scan()\textgreater{} \linebreak The current state of the robot is \$\{state\}, generate robotic plans by generating pythonic code with the use of primitive action functions. \linebreak The user is requestioned \$\{userRequest).’ & Enables LLMs for integration with robotic systems. Enables LLMs to be used to generate robotic plans taking into consideration the abilities of a robot and the conditions in the environment. & Requires expert programming to integrate into an autonomous system. Limited by context length of LLM models. & {\cite{B32-robotics-3060247,B76-robotics-3060247,B78-robotics-3060247,B79-robotics-3060247,B80-robotics-3060247,B84-robotics-3060247,B86-robotics-3060247,B87-robotics-3060247,B88-robotics-3060247}}\\

\bottomrule
\end{tabularx}}
\end{adjustwidth}

    \end{table}
    \vspace{-6pt}

\section{Human--Robot Interaction (HRI) and Communication \label{sect:sec3-robotics-3060247}}

The field of HRI for healthcare is focused on the development of appropriate design and implementation strategies for robots to assist with different tasks for a wide range of users from healthcare professionals~\cite{B23-robotics-3060247} to patients~\cite{B89-robotics-3060247}. Therefore, HRI approaches require robotic systems to understand and adapt to the needs of users. The types of HRIs used in healthcare applications include the following: (1) teleoperation, e.g., through graphical user interfaces for socially assistive robots to provide therapy and cognitive interactions~\cite{B90-robotics-3060247,B91-robotics-3060247}; (2) social interactions using natural communication modes, for example, for companionship~\cite{B92-robotics-3060247}; and (3) the use of mechanical interfaces used, for example, for surgical robots~\cite{B93-robotics-3060247}. Surgical robots require mechanical interfaces (joysticks, switches, etc.) which can result in high cognitive workloads for surgeons~\cite{B94-robotics-3060247}. On the other hand, social HRI utilizes natural language processing (NLP) to recognize user verbal requests and commands in order to provide reminders and to engage in conversations~\cite{B95-robotics-3060247}. These requests and commands are dependent specifically on pronunciation and word choices, which can potentially result in incorrect command calls being detected~\cite{B96-robotics-3060247}.

In general, it is important that interactive healthcare robots have intelligent communication abilities using multiple modalities such as spoken natural language, gaze, facial expressions, and illustrative gestures in order to be able to recognize the intent of a user and effectively convey their own intent using these modes. Current healthcare robots are capable of both single- and multi-modal interactions; however, they still have not yet been widely adopted in hospitals, clinics, and/or long-term care homes due mainly to their linguistic limitations, which can lead to critical procedural mistakes (misinterpretation of user requests) and/or frustration and lack of trust in the technology~\cite{B97-robotics-3060247}.

LLMs can address the aforementioned challenges in order to improve user experiences and the intent to use the technology~\cite{B98-robotics-3060247}. The ability of LLMs to generate dynamic multi-modal communication and detect and utilize emotional cues through the use of emotional prompting techniques such as EmotionPrompt~\cite{B99-robotics-3060247} choreographed by an LLM will result in natural HRIs that will be similar to human--human interactions~\cite{B100-robotics-3060247}. The introduction of LLMs in healthcare robots will aim to improve HRI via the cohesion between multiple communication modes; however, to date, the use of LLMs in HRI has mainly only considered a single mode~\cite{B101-robotics-3060247}.

\subsection{Single-Modal Communication \label{sect:sec3dot1-robotics-3060247}}

Single-modal robot communication in healthcare applications has mainly consisted of either textual or verbal information exchange~\cite{B102-robotics-3060247,B103-robotics-3060247}. A primary challenge for such single-mode communication is ensuring that a robot does not become monotonous or repetitive, as this can reduce user engagement and trust and also negatively impact the adoption of healthcare robots~\cite{B104-robotics-3060247}.

In~\cite{B76-robotics-3060247}, GPT-3 was integrated into the ‘Mini’ small character-like social robot to provide companionship to older adults with mild cognitive impairment. The robot engaged in cognitive stimulation games and general conversations with the older adults. The Babbage and Davinci versions of GPT-3 were used to generate user-adapted semantic descriptions and to paraphrase prewritten texts in Spanish. This integration was tailored to achieve a single consistent mode of communication, enhancing the robot’s capabilities in natural and adaptive dialogues. The research highlights the potential of streamlined adaptable interactions in social robotics made possible by the capabilities of GPT-3 to create tailored responses. The use of GPT-3, despite requiring a translation step from English, was justified by its high performance and adaptability to the specific requirements of the application compared to other models including T5 multilingual~\cite{B105-robotics-3060247}, PEGASUS~\cite{B106-robotics-3060247}, and BERT2BERT~\cite{B107-robotics-3060247}.

In~\cite{B32-robotics-3060247}, a natural language interface using GPT-3.5 was implemented into a daVinci surgical robot~\cite{B108-robotics-3060247} to provide a user-friendly interface for surgeons. The aim was to minimize the cognitive load of surgeons and improve efficiency. A surgeon inputs a verbal command using a microphone, which is then preprocessed using an off-the-shelf text-to-speech model and prompted to GPT-3.5. This prompt also includes a dictionary of possible actions that the daVinci robot can perform. GPT-3.5 is asked to match the natural language input to a robot action through generating an output based on the contents of its context window. The specific actions for the robot to execute include camera position settings, video and picture recording, and finding and tracking surgical tools based on the surgeon’s command. A Robot Operating System (ROS)~\cite{B109-robotics-3060247} node structure was used to directly provide execution commands to the daVinci robot. The system usability was tested in a laboratory where 275 natural language commands were given to the system. It was able to correctly identify and execute the intended robot action with a success rate of 94.2\%. A time delay existed between the command request and the execution of the robotic action, which was attributed to computation time to respond of the GPT-3.5 model.

In~\cite{B31-robotics-3060247}, GPT-3.5-Turbo was used to identify the intent of individuals in multi-party conversations between the social robot ARI~\cite{B110-robotics-3060247} and patients and their companions in a memory clinic in a hospital. ARI provided directions and responded to visitor/patient questions. Namely, a dataset of multi-party interactions was obtained and annotated for the intention and goal of each speaker. This dataset was created using a Wizard of Oz procedure where the operator would choose the response of the ARI robot. The \mbox{T5-large}, DialogLED, and GPT-3.5-Turbo LLMs were tested on these tasks using the dataset. The models were evaluated using zero-shot and few-shot approaches due to the limited dataset of patients. The ANOVA and Tukey HSD statistical significance tests were used to evaluate the model performance. Namely, the ANOVA test determined a significant difference in the percentage of correct annotations for intent slot recognition between GPT-3.5-Turbo and both T5-Large and DialogLED. The Tukey HSD validated that GPT-3.5-Turbo significantly outperformed all other models in terms of exact match and partial correctness of annotations in few-shot settings. It was found that GPT-3.5-Turbo used in a few-shot setting with a ‘reasoning’-style prompt, where the reasoning for the desired output was explained, had the highest recognition and tracking correctness. Namely, it was correct in 69.57\% of intent recognition and 62.3\% of goal tracking. However, all models were prone to hallucination when the prompt was phrased as a story. This can be an issue in healthcare settings, as incorrect information can be provided to patients and healthcare professionals.

In~\cite{B81-robotics-3060247}, the ChatGPT web interface was used for speaker diarization of HRIs between people and the social robot Furhat~\cite{B111-robotics-3060247}, which has human-like expressions and conversational capabilities, to investigate the capabilities of LLMs for improving HRI experience. Namely, diarization concepts such as ‘who’ spoke and ‘when’ they spoke were explored in multi-party human interactions. The robot would be able to address specific users in group interactions and allow for a less resource-intensive system which is capable of diarization. The developed system used ChatGPT in zero-shot settings to determine if the model could identify different speakers within a transcribed conversation based on the linguistic patterns of each participant involved in the conversation. A test dataset was created by using the whisper model of OpenAI~\cite{B112-robotics-3060247} to transcribe a video where a discussion between two hiring managers, a candidate, and a Furhat robot took place. The model achieved 77\% in exact matches, 0.88 in sentence level annotation accuracy, 0.92 in word-level annotation accuracy, and 0.18 in Jaccard similarity, demonstrating that large language models can be used for diarization. Currently, the time required to generate speaker labels is too slow for real-world implementation. However, ChatGPT integrated into healthcare robots for diarization tasks has the potential to reduce the administrative burden on caregivers by assuming the role of an assistant and generating structured clinical notes based on caregiver--patient interactions. Furthermore, its attention head mechanism can identify the caregiver and the patient by analyzing the linguistic patterns of each and subsequently documenting the interaction into diagnosis, symptom, and treatment sections.

\subsection{Multi-Modal Communication \label{sect:sec3dot2-robotics-3060247}}

In HRI, multi-modal communication is far more engaging and effective in building trust~\cite{B113-robotics-3060247} when compared to single-modal communication, providing a more interactive and nuanced patient experience through the use of gestures (e.g., animated speech) and body poses and varying speech intonations~\cite{B114-robotics-3060247,B115-robotics-3060247}. Multi-modal HRI emulates human interaction patterns closely~\cite{B116-robotics-3060247}, and LLMs can generate emblems (non-verbal gestures or body language that have specific meanings) for a robot to display in a zero-shot approach. LLMs are also adaptable to variations in user interactions due to word-level annotation accuracy and their attention head mechanisms which dynamically increase the importance of relevant parts of the input text. These mechanisms allow a contextual understanding of user interactions which is required while generating an appropriate speech response and emblem. The ability of LLMs to generate contextually accurate responses and emblems in HRI without the need for training every possible interaction makes LLMs suitable for facilitating HRI in healthcare settings, especially considering the significant variance in demographics and the diverse nature of interactions encountered in such settings.

LLM frameworks applied in multi-modal HRI have the potential to make HRI in healthcare settings more engaging. For example, by aligning non-linguistic commands to natural language and generating dynamic responses to user queries, they can potentially provide more natural communication during robotic-guided therapy sessions, breaking down language barriers and providing emotional support. For example, in~\cite{B82-robotics-3060247}, GPT-3.5 was used to align non-linguistic communication cues with the natural language responses of any robot capable of multi-modal communication. The Empathetic Social Robot Design Framework utilized consisted of the following modules: Speech, Action, Facial Expression, and Emotion (SAFE), alongside the user request. For Speech, GPT-3.5 considered seven types of speech, from ‘high and fast speech’ to ‘slow speech in neutral tones’. Action encompassed seven gestures, such as turning the head towards the speaker, nodding, shaking the head, interlocking hands on the table, and eye contact. Facial Expressions included ten options including frown, light smile, pout, no expression, bright smile, raised eyebrows, grin, lowered eyebrows, jaw drop, and widened eyes. For Emotion, the framework provided GPT-3.5 with ten emotional states for the robot to display: joy, liveliness, sadness, surprise, anger, worry, calmness, indifference, \textls[-5]{absence of emotion, and disgust. In a user study analogous to the Turing test, GPT-3.5 was prompted with a specific problem according to the SAFE prompt structuring, such as ‘\emph{I am too nervous for the upcoming internship interview}’.} It was also provided with an example of how it should respond. The response generated by the GPT-3.5 was then compared to a response of a human presented with the same problem. The average alignment score for speech, action, facial expression, and emotion was 26\%, 10\%, 31\%, 32\%, and 25\%, respectively. Such a robot system can be potentially useful for Reminiscence/Rehabilitation Interactive Therapy and Activities~\cite{B117-robotics-3060247} for those living with dementia. A social robot can engage older adults in reminiscence activities such as music, TV shows, and movies. It can interpret users’ non-verbal responses and adapt its interactions to suit their emotional states, providing cognitive engagement and fostering emotional well-being.

In~\cite{B83-robotics-3060247}, the text-davinci-003 model of GPT-3.5 was used to generate the dynamic responses of the Furhat robot to visitor questions in regards to news and research being conducted at the National Robotarium in the UK. The integration of GPT-3.5 into Furhat was to enable human-like speech and contextually accurate gestures while creating consistency between these modes of communication. The Furhat SDK provided the following: (1)~an automatic speech recognition (ASR) module to transcribe speech to text, (2) a natural language understanding (NLU) module to identify user intent, (3) a dialogue manager (DiaL) to maintain conversational flow, and (4) natural language generation (NLG) using GPT-3.5. More specifically, the NLG module generated responses based on engineered prompts containing the user request from the NLU module, Furhat’s personality, and past dialogue histories from the DiaL module. The responses generated by the NLG module had an associated emblem which was parsed by the Furhat SDK and presented by the robot using both verbal and non-verbal communication.

\subsection{Summary and Outlook \label{sect:sec3dot3-robotics-3060247}}

The integration of LLM frameworks for HRI into healthcare robots can potentially improve the cohesiveness and engagement of interactions between healthcare robots and patients, visitors, and stakeholders. LLMs have been embedded into social robots to improve HRI by (1) providing non-repetitive single-mode (verbal)~\cite{B76-robotics-3060247} and multi-modal communication, where the latter consists of embedding non-verbal communication into the verbal responses of healthcare robots (i.e., gestures, eye contact, and facial expressions) to users~\cite{B82-robotics-3060247,B83-robotics-3060247}, and (2) identifying the linguistic patterns of the user~\cite{B31-robotics-3060247,B81-robotics-3060247}. Moreover, LLMs have the potential to provide a more efficient interaction interface through the use of natural language to control surgical robots such as the daVinci robot~\cite{B32-robotics-3060247}.

The use of LLMs for multi-modal communication can augment robots in order to extend their use in healthcare environments by accommodating non-verbal communication with individuals living with cognitive impairments, autism spectrum disorder, and/or learning disabilities, where verbal communication alone is not always feasible. Bidirectional multi-modal communication can be used when a healthcare robot needs to effectively convey and recognize various non-verbal cues such as gestures/body language, facial expressions, and vocal intonations. For example, a healthcare robot should recognize when a patient is stressed or upset using these non-verbal cues and respond with an appropriate emotional behavior (i.e., concerned) rather than being cheerful or happy, thereby aligning its responses with the patient’s emotional states.

\section{Semantic Reasoning \label{sect:sec4-robotics-3060247}}

Significant amounts of information must be processed in healthcare, including images, data, and text, in order to minimize errors and improve the efficiency of personalized care techniques~\cite{B118-robotics-3060247}. The semantic reasoning of healthcare information requires experts to identify relationships between different factors such as genetic predispositions, lifestyle choices, environmental exposures, and/or social determinants of health which may all influence the health of an individual~\cite{B119-robotics-3060247,B120-robotics-3060247}. Therefore, semantic reasoning encompasses the understanding of meanings, concepts, and relationships between data and medical knowledge. In particular, ontologies and knowledge graphs created from patient EHRs are currently used to interpolate how various symptoms, diseases, and treatments are interrelated and influence one another in order to make predictions for clinical decision-making and patient care~\cite{B121-robotics-3060247}.

COVID-19 increased the demand for telehealth by 367\% in adults aged 55--65 and by 406\% for adults aged 65 years and older~\cite{B122-robotics-3060247}. However, frameworks such as the eCoach personalized referral program to help people stay active and achieve physical activity goals~\cite{B123-robotics-3060247} and the Babylon Chatbot which provides healthcare consolations through a mobile app~\cite{B124-robotics-3060247} both require complex ontologies and access to large amounts of contextual information to generate personalized recommendations. Therefore, it is not only a matter of creating healthcare-focused datasets and ontologies to train deep learning and NLP models to be able to provide predictions and inference; there needs to also be an effective approach to creating such inferences in order to (1) facilitate the seamless exchange of semantic reasoning frameworks among healthcare institutions~\cite{B125-robotics-3060247} and (2) simplify the creation of frameworks that can effectively capture and convey the complex semantics of medical datasets, terminologies, and environments (hospitals, clinics, etc.)~\cite{B126-robotics-3060247}.

Traditional robotic semantic reasoning frameworks consist of three core components~\cite{B127-robotics-3060247}: (1) knowledge resources (raw data), (2) computational frameworks (mathematical models), and (3) world representations (scene/environment representations). Knowledge resources include the data from which semantic knowledge is extracted\linebreak (i.e., EHRs, clinical trial results). These data are used to train the LLMs. Computational frameworks consist of models such as transformers (LLMs)~\cite{B128-robotics-3060247}, probabilistic models (Bayesian networks)~\cite{B129-robotics-3060247}, or deep learning models (long short-term memory (LSTM) networks) used to capture temporal dependencies in data~\cite{B130-robotics-3060247}. These computational frameworks are the models that encode the relationships between concepts~\cite{B131-robotics-3060247}. They then use the encoded knowledge to perform inference~\cite{B127-robotics-3060247}. World representations are used by robots to model their surrounding environments and their own behaviors~\cite{B132-robotics-3060247}. In the case of LLMs, the world representation provides an LLM with a scene description of a robot’s working environment in the context window and thereby ‘grounds the LLM’ to its current environment~\cite{B20-robotics-3060247}. The aforementioned core components allow robots to perceive, understand, and generalize semantic knowledge in order to improve performance in real-world tasks. The incorporation of LLMs for semantic reasoning in healthcare robots offers multiple advantages. Namely, healthcare robots need semantic reasoning to identify relationships between a task, an environment, and a user command. Semantic reasoning can be used (1) when generating object manipulation plans for surgical tools, equipment, instrumentation, and medical supplies, (2) to navigate to specific areas and regions in the environment such as the OR and/or patient rooms, (3) to interact with people (to provide assistive HRI) and other robots, and (4) to complete task management functionalities or specific tasks and services (patient education, guidance, informatics).

To date, only a handful of robots have been incorporated with LLMs for the purpose of semantic reasoning. For example, in~\cite{B77-robotics-3060247}, the semi-humanoid robot NICOL (Neuro-Inspired COLlaborator) used GPT-3 to reason about multi-modal inputs (sound, haptics, visuals, weight, texture) in order to improve robotic perception in object manipulation to help differentiate between visually similar objects. Namely, GPT-3 was used to perform interactive multi-modal perception and robot behavior explanations. The MATCHA (multi-modal environmental chatting) prompting technique provided specific action prompts to GPT-3 which included stored information about the robot’s environment. Namely, these action prompts included descriptions of the actions, as well as an example of an expected response. GPT-3 provided instructions to the robot to determine the target block by knocking on and weighing the blocks one by one. The robot then reported the findings (stored information) of each step back to GPT-3. This process was repeated until GPT-3 was able to determine which was the correct block with high accuracy (\textgreater{}90\%). The system has been tested only in simulation by prompting NICOL to pick up a specific block made of metal out of three blocks, where the characteristics of the blocks were provided rather than perceived by the robot. Even though this task has not been implemented directly in a healthcare setting, the potential of MATCHA can be explored particularly for improving operational efficiency and decision-making for patient care in terms of multi-modal information gathering and reasoning for robot manipulation tasks such as retrieving and handing over (1) medical supplies on shelves in a supply room or (2) surgical tools on a table in the OR.

In~\cite{B87-robotics-3060247}, the ‘SayCan’ method integrated the PaLM 540 B parameter model, PALM-SayCan, into a mobile manipulator~\cite{B133-robotics-3060247} for semantic reasoning in the context of human-centered environments (i.e., a kitchen in an office). The manipulator provided the perception and manipulation capabilities, while PaLM provided high-level semantic knowledge about tasks to promote successful task completion. The system was trained on a mock kitchen and tested in an office kitchen environment. SayCan is not only prompt-based but also uses a temporal difference reinforcement learning (RL) approach. This approach learns the rewards of each action (completing the objective vs. course of action) and therefore prioritizes the executable actions based on the robot’s current environment. PaLM 540B is provided with prompts that include the robot’s capabilities and their descriptions and the actions the robot can take expressed as a function (e.g., ‘\emph{goto()’}). PaLM 540 B is given a task, such as find object X, and asked to generate multiple possible actions paired with the likelihood of each prediction. The action probabilities are then multiplied by the probability of success (acquired through the temporal difference method in RL). The action function with the highest probability which is the output of the LLM is then executed by the robot’s low-level control system. This process is repeated for all subsequent steps in the generated plan. The approach was benchmarked by obtaining the plan success rate and the execution success rate in completing a task such as bringing a bag of rice chips from a drawer. The SayCan development improved task execution in human-centered environments by leveraging semantic reasoning. The potential healthcare applications of ‘SayCan’ can be extended to mobile social robots directing patients/visitors to specific rooms or departments in a hospital and/or mobile manipulators fetching or localizing essential medical supplies.

In~\cite{B85-robotics-3060247}, the LLM-Grounder, an open-vocabulary zero-shot LLM-based 3D visual grounding pipeline, was introduced. LLM-Grounder integrated GPT-4 and a robotic simulated agent to reason about the semantic relationships between high-level commands given by a user (i.e., find the grey monitor on top of the smaller curved desk) and the working environment (simulated office) for object localization tasks. GPT-4 was used to break down natural language commands into their semantic constituents, the target and a landmark. The ability of GPT-4 to (1) identify the landmark and the target and (2) differentiate between the two provides an efficient method to generate from high-level user commands to robot actions to perform in order to complete a goal. Namely, OpenScene~\cite{B134-robotics-3060247}, which is a 3D visual grounding method that uses the transformer architecture to generate 3D scene layouts based on textual descriptions, was used. GPT-4 provided OpenScene with (1) the target name and its attribute (monitor, light grey) and (2) the landmark name and relation (curved desk, small). OpenScene returned to GPT-4 the respective bounding box volumes and distances. GPT-4 was then used to reason about the size of objects and their relative location to the landmark, deciding on which found target and landmark pair(s) has the highest likelihood to be correct. Therefore, GPT-4 is used to provide an efficient method for robots to understand the semantics of high-level user commands and for them to act on these commands. LLM-Grounder was evaluated on the ScanRefer~\cite{B73-robotics-3060247} benchmark, which is a standard dataset for assessing 3D vision--language grounding capabilities. The ScanRefer dataset consists of scenes ranging from wildlife to home environments, where each point cloud and image has a text description~\cite{B73-robotics-3060247}. LLM-Grounder demonstrated state-of-the-art performance when used for zero-shot open-vocabulary grounding, excelling in complex language (increased \# of nouns in command) query understanding over approaches that do not use LLMs and rely solely on CLIP. The ability of a robot to correctly understand user commands in the context of locating objects in human-centered environments is crucial for the successful implementation of healthcare robots. A robot deployed in a healthcare environment must be able to identify the correct target (out of many) in cases where items are not easily distinguishable, such as identifying a specific medication on a medication cart where labels are not visible. This can only be achieved if the healthcare robot is able to understand the pill bottle description and semantic relationship to the landmark specified by the user in natural language.

In~\cite{B74-robotics-3060247}, the ‘Lang2LTL’ method integrated the SPOT quad-pedal robot~\cite{B135-robotics-3060247} with GPT-4 and the T5 base to provide SPOT with the semantic reasoning capabilities required to understand and act on user speech commands in the context of navigating environments ranging from offices to city streets. Lang2LTL used these LLMs to break down navigational commands (i.e., ‘Go to the store on Main Street but only after visiting the bank’) to Linear Temporal Logic (LTL) in the form of sequential objectives, such as (1) go to the bank \mbox{and (2) go} to the store on Main Street after. User commands in the form of natural language were given to GPT-4 to identify the referring expressions (RFs) such as the store on Main Street or the bank. After identifying the RFs, the RFs were grounded to known physical locations (retrieved from a database) by being compared to known proposition embeddings which are location description/coordinate pairings using cosine similarity. The grounded RFs were used as an input into the fine-tuned T5 base model to generate LTL formulas. The fine-tuned T5 model was then used to (1) generalize the initial input command (go to ‘a’ but only after visiting ‘b’) and (2) generate the LTL formula required by the planner to facilitate navigation via the robot’s low-level controller. Lang2LTL was tested on SPOT in an indoor environment consisting of bookshelves, desks, couches, elevators, tables, etc. SPOT successfully grounded 52 commands, including 40 satisfiable (executable) and 12 unsatisfiable ones (unsafe to execute). The potential applications of ‘Lang2LTL’ in healthcare are robot delivery tasks within clinical settings. Namely, the added ability to identify unsafe robotic action executions can aid a robot planner. This can be achieved by identifying the semantic importance of user commands and the order of operations, such as retrieve a walking aid for a patient, then visit the patient’s room to provide them with the aid.

\subsection*{Summary and Outlook }

In general, the integration of LLM frameworks for semantic reasoning into healthcare robotics has the potential to improve robot autonomy in complex and dynamic healthcare environments. Currently, LLM-based semantic reasoning has increased robot autonomy, as the transformer architecture of LLMs semantically reasons about the entire input at once, facilitating faster and more accurate decision-making in comparison to traditional sequential processing models such as Recurrent Neural Networks (RNNs)~\cite{B39-robotics-3060247}. This approach allows robots to understand complex relationships among object characteristics, robotic action outcomes, the spatial importance of target objects versus landmarks, and user-imposed constraints. Healthcare robots can use frameworks such as MATCHA~\cite{B77-robotics-3060247} for autonomous capabilities such as the detection and manipulation of medical supplies based on characteristics such as weight and texture. Additional applications include robots augmented with SayCan~\cite{B87-robotics-3060247}, LLM-Grounder~\cite{B85-robotics-3060247}, or Lang2LTL~\cite{B74-robotics-3060247} frameworks for \mbox{(1) navigating }efficiently within healthcare facilities by understanding spatial layouts and identifying potential unsafe surfaces (e.g., wet floors), (2) executing patient-specific tasks by interpreting natural language commands given by care providers such as retrieving a vital signs monitor cart before visiting a patient in their room, or (3) increasing the surgical team’s situational awareness by alerting them to equipment needs and preemptively managing robotic tool positions, thus ensuring sustained operational focus and efficiency.

\section{Planning \label{sect:sec5-robotics-3060247}}

An existing challenge for healthcare robotics is autonomously planning safe and effective behaviors in real time for healthcare tasks. These tasks can include (1) navigating through complex hospital environments to find healthcare professionals and patients and to escort visitors, (2) managing and delivering medications and medical supplies, (3) assisting in surgery by tool handovers and supporting precise tool movements, and (4) facilitating both physical and cognitive rehabilitation with different user groups. Each of these tasks requires a robot to perceive and interpret its surroundings and make real-time decisions that ensure safety and task effectiveness. For example, surgical robots operating within soft tissues and navigating curved paths face a particularly uncertain and dynamic environment due to the complex and variable nature of human anatomy. Soft tissues can shift or deform during procedures, altering expected pathways and requiring real-time adjustments in the robot’s movements. Additionally, the inherent variability in patient anatomy means that pre-planned paths may not always apply precisely, necessitating continuous sensory feedback and adaptive control strategies to accurately guide surgical tools without causing unintended damage. This environment demands high levels of precision and adaptability from surgical robots to ensure safety and effectiveness in their operations.

In general, healthcare robotic task plans need to be adaptable to different situations and people. Currently, the majority of surgical robots are teleoperated by a surgeon~\cite{B136-robotics-3060247,B137-robotics-3060247}. An autonomous surgical robot needs to adjust to changes in patient anatomy in real time~\cite{B138-robotics-3060247}. Presently, the lack of real-time 3D sensing is a significant constraint~\cite{B139-robotics-3060247}. This constraint prevents surgical robots from (1) operating in realistic conditions (i.e., lighting changes, occlusions) and (2) operating on non-planar surfaces~\cite{B139-robotics-3060247}. While minimally invasive methods such as the multi-camera CARET system proposed in~\cite{B140-robotics-3060247} attempt to enhance the surgical field of view without making additional incisions, they do rely on complex intra-camera tracking to maintain an expanded view. Namely, this system determines the correspondence between different cameras at the initialization stage and updates the expanded view frequently when there is enough overlap between views. However, this can still lead to inaccurate mosaicking results due to error accumulation over time~\cite{B141-robotics-3060247}. Although the visual field provided by instruments like laparoscopes is not always optimal, surgeons are able to effectively conduct surgeries due to their continuous learning and ability to interpret surgical situations and semantically reason about the current state of the operation. Integrating LLMs with surgical robots will allow for this decision-making process by semantically reasoning about the current state of the surgery and generating robot action plans from the limited information about the operating environment provided by imaging sensors such as laparoscopes and the multi-camera CARET system.

Existing robotic planning frameworks that have incorporated autonomy have mainly used heuristic and DL models. Heuristic methods, such as (1) genetic algorithms~\cite{B142-robotics-3060247}, (2) Greedy Best-First Search~\cite{B143-robotics-3060247}, and (3) Simulated Annealing~\cite{B144-robotics-3060247}, aim to identify the sequence of robot actions for task planning. Additionally, DL methods, such as (1) the path planning and collision network (PPCNet)~\cite{B145-robotics-3060247}, (2) DRL models~\cite{B146-robotics-3060247}, and (3) LSTM networks~\cite{B147-robotics-3060247}, learn from training examples and episodes in order to autonomously plan robotic actions and generalize to real-world situations. The unpredictable nature of healthcare environments, with sudden medical emergencies and changing patient conditions, requires a robot to autonomously react to new or quickly changing scenarios. Heuristic models are based on a set of rules and tend to yield satisfactory rather than optimal solutions~\cite{B148-robotics-3060247}. Furthermore, the efficacy of DL models in generalizing to unseen data is significantly influenced by the diversity and size of training datasets~\cite{B149-robotics-3060247,B150-robotics-3060247}. Ensuring that robotic planners can adapt to new and varying conditions beyond their training data is crucial for their effective deployment in healthcare settings. Consequently, there is a need for healthcare robotics planning methods that can manage extensive datasets encompassing immense amounts of high-quality data describing human-centered scenes, medical and general knowledge, and various medical procedures to enable robots to determine optimal plans for complex tasks. These include robot-guided surgery for precise operational assistance, real-time diagnostic analysis during various medical procedures to inform decisions, and robot-led interventions and rehabilitation that adjust treatment plans based on ongoing patient evaluations across different healthcare settings.

LLMs can interpret complex instructions and patient data, facilitating robots to make informed decisions in real time. Unlike heuristic models that rely on predefined rules, LLMs take advantage of attention head mechanisms which add bias towards generating outputs related to the contents of the context window which include information from the environment~\cite{B151-robotics-3060247}. The contents can be updated in real time to contain the current state of both the robot and the healthcare environment, thereby having the LLM generate robotic plans specific to these states. Furthermore, both the ability of LLMs to dynamically add bias when generating a plan and to semantically reason allows LLMs to adapt already generated robotic plans by supplementing the model input with state changes and iteratively refining the sequence of planned robot actions to ensure successful execution. LLMs mimic human planning in contrast to the aforementioned DL methods. Namely, the latter lack the ability to iteratively and efficiently update specific portions of robotic plans to address run-time issues such as equipment failures in surgical rooms or variations in treatment responses during rehabilitation sessions~\cite{B152-robotics-3060247}. Healthcare robots need the adaptable capabilities of LLMs for planning to enhance their effectiveness in diverse and dynamic environments, such as (1) emergency rooms where conditions can change rapidly, (2) ORs that require precision under varying circumstances, and (3) rehabilitation settings where patient responses can be unpredictable and varied. Therefore, using LLMs in healthcare robotics planning can provide (1) recovery in cases of sudden environmental changes, for example, an influx of patients due to an outbreak where the robot has to prioritize which patients it should assist first, (2) robotic plans suitable for execution around vulnerable people by inferring about social norms, for example, when escorting frail patients to their rooms, (3) fine-tuned motion plans to facilitate rehabilitation for stoke survivors and physically impaired patients, and (4) dynamically generated robotic plans that can be interpreted by non-expert roboticists (healthcare professionals), for example, in surgical robotics where the surgeon must anticipate the robot’s next planned motion. Furthermore, the integration of \mbox{Vision--Language} Models (VLMs) can provide the utilization of the visually anchored features of VLMs to assist with task planning in healthcare robotics~\cite{B153-robotics-3060247}. Namely, VLMs can aid in grounding in the physical world for healthcare robotic applications by integrating visual and textual data into a joint embedding space, which LLMs lack on their own, enabling robots to understand physical contexts and respond more appropriately to environmental cues.

To date, a handful of robots have used LLMs for task planning in human-centered environments. For example, in~\cite{B88-robotics-3060247}, the LLM-BRAIn method consisted of the Stanford Alpaca 7B parameter LLM in order to generate behavior trees (BTs) for identifying and retrieving objects for mobile manipulator robots. BTs provide a modular behavior structure consisting of nodes (robot execution steps) that can be easily scalable. Initially, GPT-3 was used to create a synthetic training dataset where each example is an XML file containing a randomly generated BT for a mobile manipulator robot and a description of the robot’s movement and object manipulation steps. The synthetic dataset was used to fine-tune the Alpaca model parameters for generating robot BTs. To generate BTs based on user requests, the fine-tuned Alpaca model was prompted with a list of robot actions to execute (e.g., take object, scan area) and the user request (e.g., if object is visible, move towards it, take it, and process it). A ROS2 BT interpreter~\cite{B149-robotics-3060247} took the output of the fine-tuned Stanford Alpaca model (XML format) and generated executable actions for the robot. The BRAIn method included the robot’s functionality at all planning steps in an understandable format for the user. To incorporate the LLM-BRAIn method into healthcare robots, the generated BTs used to train LLM-BRAIn should include healthcare-related tasks instead of the general localization and retrieval of objects, thus allowing healthcare professionals to understand a healthcare robot’s intentions and actions and potentially stopping the execution of a task if there are any concerns, ensuring safe interactions when providing patient care.

In~\cite{B84-robotics-3060247}, GPT-3.5 was used to guide a Franka Emika Panda robot arm operating in a mock kitchen to facilitate the handover of dirty kitchen utensils (i.e., fork, spoon, knife) to a human washing the dishes. The zero-shot capabilities of GPT-3.5 in planning collaborative robotic tasks that align with human social norms were investigated. GPT-3.5 was prompted with examples from the MANNERS-DB~\cite{B154-robotics-3060247} dataset which contains HRI scenarios where appropriate robot behaviors are represented. GPT-3.5 was then tested for its ability to\linebreak (1) establish/maintain human trust while guiding the robot to hand off sharp objects\linebreak (i.e., knives) and (2) predict cultural and social norms using (a) the Trust-Transfer~\cite{B155-robotics-3060247,B156-robotics-3060247} which contains 189 instances of driving and household tasks where participants rated their trust in the robot completing the task on a seven-point Likert scale and (b) the SocialIQA~\cite{B157-robotics-3060247} which contains 1954 testing examples, each containing a content, question, three possible answers, and a ground truth, where a robot can be tested on its commonsense reasoning. In a physical experiment consisting of the robot and a user, GPT-3.5 facilitated robot-assisted utensil washing while allowing for user intervention. Success and failure in handovers were reported with only 28.1\% of participants trusting the robot with knife handovers. GPT-3.5 can improve patient and care provider HRI by utilizing in-context learning to refine robotic task plans, ensuring they align with user perceptions through analyzing past failures and applying necessary corrections. The potential applications in healthcare-related tasks can include (1) delivering medications/supplies by navigating busy hospital corridors (e.g., giving way to humans and incorporating other social etiquettes) and (2) autonomously cleaning patient rooms while they are still in the room (e.g., keeping a safe distance, smooth navigation).

In~\cite{B78-robotics-3060247}, the ProgPrompt method integrated GPT-3 into the Franka Emika Panda manipulator robot~\cite{B158-robotics-3060247} to facilitate object manipulation in a physical mock kitchen and simulated VirtualHome~\cite{B159-robotics-3060247}. The prompts were based on object manipulation and sorting, for example, ‘\emph{sort fruits on the plate, and sort bottles in the box}’. In the implementation of the ProgPrompt method, two instances of GPT-3 were used to generate and refine robot task plans. The first instance of GPT-3 was given a prompt containing robot primitive actions, a list of available objects in the environment (which were dynamically identified through an open-vocabulary object detection model in real-world applications or predefined in simulations), and example plans illustrating the desired task structure and outcomes. GPT-3 then generated an initial plan with natural language comments detailing each step and logical assertions (e.g., ‘\emph{put the banana on the plate}’) to provide a method for error recovery in case of execution failures. The second GPT-3 instance iterated over the initial plan with a focus on reasoning about the current semantic state of the environment and the task at hand. The plan was modified by adding or removing steps to successfully achieve the goal provided in the first instance. For example, if the task involved microwaving salmon, the second instance considered the state of relevant objects and actions (such as whether the microwave door is open or the robot is holding the salmon) to determine the next logical step in the plan. This iterative process allowed for dynamic adaptation and refinement of the plan, ensuring it was contextually appropriate and executable within the given environment. The ProgPrompt framework was tested in the VirtualHome environment using GPT-3 to generate task plans across 10 tasks, such as ‘\emph{put salmon in the fridge}’ or ‘\emph{put banana on plate}’. The ProgPrompt method excelled in simple tasks. However, it was less effective in complex sequential tasks, such as ‘\emph{put the banana on the plate and the pear in the bowl, sort the fruits on the plate and the bottles in the box}’. The ProgPrompt method integrated with GPT-3 can potentially be used in healthcare robotics to enable robotic medication dispensing tasks such as precisely dispensing into measuring cups, using logic assertions and semantic reasoning based on the textual description of the scene to determine when the cup is full, thus eliminating the need for specialized hardware such as prescription dispensing systems~\cite{B160-robotics-3060247} and also relying on vision to control a robotic manipulator for accurate dispensation.

In~\cite{B79-robotics-3060247}, the RoboGPT framework utilized three instances of GPT-3 in the Franka Emika manipulator robot~\cite{B158-robotics-3060247} to improve robotic manipulation tasks such as word spelling using letter blocks, moving letter blocks, bin packing and pyramid stacking, and house building using cubes. The first instance of GPT-3 was a decision bot, which generated a sequence of robot actions based on a task-oriented prompt. The prompt provided detailed background information, including (1) the robot’s capabilities, (2) a description of the scene, and\linebreak (3) guidelines on the model’s response format. It also included robot API commands such as ‘\emph{envs.pickObject(object\_name)}’ and the coordinates of objects within the scene. The decision bot iterated on the generated sequence of robot actions to resolve any run-time errors to ensure the robot can perform each step of the plan. The evaluation bot (second instance) was deployed to verify that the sequence of robot actions generated by the decision bot aligned with the task requirements, effectively serving as unit tests to confirm the correctness of each action. Lastly, the corrector bot was given the generated plan and the results from the evaluation bot in order to identify the reason for the failure of the plan. The prompt of the corrector bot included background information, describing its role specifically as a ‘\emph{code corrector}’, and the coordinates of the objects in the scene. The prompt also outlined a structured analysis process, which included the following steps: guessing the intended spatial relationships between objects from input codes, determining actual spatial relationships from final object states, and analyzing discrepancies to suggest possible reasons for any failures. The evaluation bot validated the decision bot’s plan by verifying the correct stacking of cubes. It used robot API functions such as the ‘\emph{self.checkOnTop(object\_1\_name, object\_2\_name)}’ method to confirm that the objects were accurately placed according to the task requirements and object attributes provided. This collaborative system enabled precise task execution through the iterative refinement and evaluation of the generated plan. After a task was successfully completed using the plan, the plan and the outcome of the plan (i.e., success or failure) served as a demonstration example used to train a DRL model to perform the same tasks, thereby improving the efficiency of generated robot action plans by reducing the reliance on the decision, evaluation, and corrector bots. The RoboGPT framework can be advantageous for healthcare robotics in terms of assisting care providers with repetitive and non-repetitive tasks. For many repetitive tasks, such as organizing medical equipment or retrieving surgical tools in the OR, the framework can enable the robot to rely on the DRL model for robot task execution instead of generating repetitive plans using LLMs, thus improving efficiency. As for non-repetitive tasks, such as triaging, the three bots can provide a healthcare robot with the versatility needed to adapt to changes in the environment and the constraints of the task and patients.

In~\cite{B80-robotics-3060247}, GPT-3 was used as the robotic planner for the ‘Toyota HSD’ service robot operating in a home environment. The robot fulfilled user requests with respect to general-purpose service tasks, such as ‘\emph{pick up the apple from the bookcase and put it on the storage table}’. A task instantiation module was used to execute robot actions planned by GPT-3, such as the primitive robot actions ‘\emph{move\_to()}’, ‘\emph{grasp()}’, ‘\emph{pass\_to()}’, and ‘\emph{visual\_question\_answering()}’. An inference module was used to extract information from the environment. The inference module consisted of (1) Google Cloud speech recognition~\cite{B161-robotics-3060247} to transcribe user commands, (2) object detection using YOLOv7~\cite{B162-robotics-3060247} to recognize household objects in 2D images, (3) a visual question answering (VQA) model to scan the environment based on RGB images with the purpose of locating a specific object, (4) the open-vocabulary object detection model Detic~\cite{B163-robotics-3060247} to obtain textual descriptions of the environment to provide context to GPT-3, and (5) the EZPOSE human pose estimation model~\cite{B164-robotics-3060247} to identify people for human guidance through pose estimation. Once GPT-3 is prompted with a task-oriented prompt containing the user’s request (using speech--text to transcribe user requests), it generates a sequence of primitive robot action functional calls using the user-specified targets as arguments, for example, the ‘\emph{locate the fruits in the dining room}’ argument to the VQA mode would be ‘\emph{where are the fruits?}’. Furthermore, the locations of the arguments in the generated plan are checked against an object location database before being executed. If the location is found, the plan is executed or else the inference module is queried with images of the environment using point cloud and RGB information. Then, GPT-3 is directly questioned about the possible locations of the object in the context of the current scene description. This framework was tested at the robocup@home Japan open competition, and it won first place. The robot planner can be extended to healthcare environments to generate robot action plans based on all the available information from the scene. The framework developed can be classified as a VLM, as RGB and point cloud data are used in conjunction with an LLM to generate plans which consider all features of the environment (i.e., people, objects, obstacles). In healthcare settings, this integration can effectively provide patient/visitor navigation guidance through social navigation, enabling healthcare robots to autonomously generate and update navigation paths in real time while adapting to dynamic changes such as increasing crowds or new emergency situations. These robots can interact using natural language, adjust their navigation pace and route based on verbal feedback to ensure the user’s comfort, and cater to specific needs such as slower movement for elderly patients or quick access for emergencies. This approach can enhance the overall efficiency and user experience in navigating large hospital environments.

In~\cite{B86-robotics-3060247}, GPT-4 was integrated into the Alter3 android robot~\cite{B165-robotics-3060247} for the self-planning of the robot’s physical actions in order to adapt its own pose to user speech requests in applications which include \emph{‘Take a selfie’, ‘Pretending to be a ghost’, and ‘I was enjoying a movie while eating popcorn in the theater, when I suddenly realized that I was actually eating the popcorn of the person next to me’}. Two frameworks were developed. The first framework consisted of task-oriented prompting to plan the limb movements of the Alter3 robot and used EmotionPrompt to ensure that there are expressive gestures included in the generated movement plan. For example, GPT-4 was instructed to generate a high-level plan of how Alter3 should choreograph its limbs in response to, e.g., ‘\emph{drink some tea}’. The task-oriented prompt used the high-level plan to generate a sequence of robot actions corresponding to each limb of the robot to realize the planned expressions and movements, where the emotional prompt placed emphasis on exaggerating the emotional and facial expressions associated with the user request. Namely, the task-oriented prompt generated a sequence of function calls that were executed by the control system of Alter3, where each function call controls one of Alter3’s 43 motion axes. Each limb (eyebrows, shoulders, index finger, etc.) has multiple axes. The effectiveness of this approach in articulating an android robot was explored in a user study. In the study, participants interacted with Alter3, where GPT-4 generated an action plan based on input prompts. The participants failed to distinguish between GPT-4-generated and robotic-expert-programmed movements. The second framework introduced a closed-loop system that builds onto the first framework by storing generated sequences of robot actions in a JSON database to be used again for similar requests. For example, if the robot has been asked to ‘\emph{take a selfie}’, the plan which articulates the robot’s limbs will only need to be generated once, and any subsequent request to ‘\emph{take a selfie}’ will use the already generated plan stored in the JSON database, therefore reducing plan generation time and improving responsiveness. Additionally, this framework was further expanded to incorporate a ‘Social Brain’ mechanism, employing multiple instances of GPT-4, each with a distinct role, to mimic human collaboration in problem-solving. A potential healthcare application for GPT-4 in robotic planning is to autonomously generate detailed rehabilitation plans that can be reviewed by a physiotherapist before they are executed, thereby keeping the therapist involved. The generated plans can be updated in real time based on inputs from patients with respect to their comfort and pain levels during a rehabilitation session. This dynamic adjustment helps in providing personalized care, and the initial plans serve as a clear communication tool for healthcare professionals to understand and oversee the treatment protocol.

In~\cite{B75-robotics-3060247}, the KnowledgeBot framework used the T5 LLM as the backbone for a robot action planner used in conjunction with a conversational embodied agent in the Alexa Prize SimBot Challenge environment~\cite{B166-robotics-3060247}. A T5-Large was used to train an object generator to generate a list of objects of interest based on a user-provided task description, such as ‘\emph{faucet, house plant, cup}’ for ‘\emph{water the plant}’. Then, each object was concatenated with the task description and used by the encoder--decoder-based planner (T5 backbone) to generate step-by-step robot actions to execute. A variation of this procedure was developed in order to emulate human cognitive processes in task planning and execution to ensure that each step is informed by the overall task context and the progress made. In order to generate step-by-step robotic action plans, the task and its generated steps are passed to the object generator and an encoder in parallel. Then, the output from the encoder goes to the (1) decoder and (2) is paired with the generated list of objects to be analyzed by the attention mechanism which adds bias to the object relevant to the generation step. Finally, the output from the attention mechanism is also provided to the decoder, where the next step of the plan to complete a task is generated. The generated step is concatenated to the task description (initial input), where the process is repeated until a complete plan is generated. In the Alexa Arena environment~\cite{B167-robotics-3060247}, the KnowledgeBot framework was used to generate robot actions for an embodied AI in gaming sessions. The framework was evaluated based on the Goal Completion metric, which measures the fraction of game sessions successfully completed. The success rate on unseen scenarios achieved in the Alexa Arena environment was 13.6\% on tasks such as ‘\emph{pick up the milk in the fridge and place it on the table}’ which had a 1.72\% improvement over the Alexa Prize Team Baseline. The potential applications of the KnowledgeBot framework in healthcare could involve using a model with domain-specific knowledge such as MED-PaLM 2~\cite{B41-robotics-3060247} acting as the object generator, where the LLM interprets patient conditions or staff instructions to generate lists of relevant objects and actions such as specific medications, medical devices, and procedural steps which are then used by the LLM (i.e., GPT-4)-based planner to generate to plan and execute precise context-aware assistance tasks.

\subsection*{Summary and Outlook }

LLMs used for robot planning purposes are able to create high-level plans by generating a sequence of atomic actions based on textual descriptions of the robot’s environment provided to the model in the input prompt. Task plans have been used mainly for object handling tasks, including object (1) localization and retrieval~\cite{B80-robotics-3060247,B88-robotics-3060247}, (2) handover to humans~\cite{B84-robotics-3060247}, and (3) sorting~\cite{B78-robotics-3060247,B79-robotics-3060247}. Extending these abilities to healthcare robotics through integrating LLMs and VLMs into planning frameworks can potentially improve the efficacy of healthcare robots by enabling real-time decision-making in complex and unpredictable medical environments. These models can autonomously generate and modify task sequences in response to dynamic conditions such as emergency interventions or sudden changes in a patient’s health status. Furthermore, LLMs can be used with surgical robots to generate human-understandable plans and provide reminders and cues based on the current state of the OR (i.e., understanding non-verbal communication of surgeons or OR nurses) to increase the situational awareness of the operating team. This functionality allows surgeons and physicians to stay informed and involved, ensuring that plans are verified and approved before execution in order to maintain high levels of transparency during procedures. For rehabilitation robots, LLMs facilitate the customization of therapeutic exercises in real time and again by generating human-understandable plans so that the therapist is kept informed. They can also adapt treatments to improve the recovery progress of patients. Moreover, the strength of LLMs in generating high-level task plans enables them to serve as frameworks for other deep learning models, such as DRL systems, to execute more detailed low-level commands for the control of robot arms, end effectors, and mobile platforms. This layered approach to task planning ensures that while LLMs handle broad strategic decisions, the finer tactical aspects of robot control are refined through continuous learning models, ensuring the precision execution of complex tasks. This integration not only maximizes the effectiveness of interventions but also ensures that robotic operations are adapted to the immediate needs of the healthcare setting.

\section{Ethical Considerations of Robotics Using LLMs in Healthcare \label{sect:sec6-robotics-3060247}}

Ethics is a critical aspect of healthcare, encompassing caregiver--patient interactions and the use of technology to enhance patient care~\cite{B168-robotics-3060247}. Ethical considerations for the use of healthcare robots embedded with LLMs need to be addressed for the widespread adoption of this emerging technology in order to support patients, healthcare professionals, and advancements in this area. The ethical considerations include accountability, humanizing care, and privacy~\cite{B169-robotics-3060247}. A holistic approach is essential, ensuring patient autonomy over their body and medical information with the expectation of improving their health. Maintaining the privacy and protection of personal health information is paramount, emphasizing the importance of informed consent for data usage and increased efforts to stop data commercialization while increasing transparency in how the data are used~\cite{B170-robotics-3060247}. Regardless of the robotic care delivery method, maintaining quality and equitable access across all demographics is important.

There have been extensive separate reviews of the ethics of the use of LLMs in healthcare~\cite{B171-robotics-3060247,B172-robotics-3060247,B173-robotics-3060247,B174-robotics-3060247,B175-robotics-3060247,B176-robotics-3060247,B177-robotics-3060247,B178-robotics-3060247,B179-robotics-3060247,B180-robotics-3060247} and the use of robots in healthcare~\cite{B181-robotics-3060247,B182-robotics-3060247,B183-robotics-3060247,B184-robotics-3060247,B185-robotics-3060247,B186-robotics-3060247,B187-robotics-3060247,B188-robotics-3060247,B189-robotics-3060247,B190-robotics-3060247,B191-robotics-3060247}. Although there exist frameworks such as Ethically Aligned Design from IEEE~\cite{B192-robotics-3060247}, The Toronto Declaration: Protecting the right to equality and non-discrimination in machine learning systems~\cite{B193-robotics-3060247}, and the AI Universal Guidelines~\cite{B194-robotics-3060247} which provide guidelines on how the ethics of society should be considered during the design phase, they do not provide enough insight into exactly how regulations should be established for generative AI models. Moreover, although the current gaps in regulatory frameworks for AI have been identified and countries such as the United States and China have started the process of establishing regulatory frameworks for the use of AI in healthcare, these frameworks are not applicable to generative AI as the technology is still evolving and it is difficult to develop a framework which covers all the potential impacts of generative AI~\cite{B195-robotics-3060247}. Therefore, current guidelines and research do not yet address the problems that arise from combining generative AI (LLMs) and healthcare~robots.

This section aims to introduce a discussion on the use of LLM frameworks for healthcare robots for the facilitation of (1) multi-modal communication, (2) semantic reasoning about healthcare environments and patient and care provider needs, and (3) generating and executing safe robot action plans around vulnerable people, including frail and cognitively impaired individuals. We discuss these main points as they pertain to accountability, humanizing care, and privacy. Namely, we discuss these three ethical concerns as they directly relate to the robot multi-modal communication, semantic reasoning, and robotic task planning topics discussed in this paper.

\subsection{Accountability \label{sect:sec6dot1-robotics-3060247}}

Accountability in healthcare robotics centers on identifying who is responsible for system errors and adverse events~\cite{B169-robotics-3060247}. To date, there has been no consensus on how the accountability of healthcare robots embedded with LLMs should be considered and who should be accountable. LLMs are trained on large datasets containing human language, and therefore, LLM frameworks used in robotics for multi-modal communication are often perceived as proficient in understanding human language and kinesics~\cite{B151-robotics-3060247}. However, LLM frameworks are not sensitive to changes in sentence structure, word choice, or grammar when identifying user requests~\cite{B196-robotics-3060247}, thereby promoting the illusion that they understand human communication. Although LLMs appear to understand human language through (1) generating grammatically correct text and (2) propositional reasoning about sentences such as ‘\emph{The doctor treated the child with the fever}’ or ‘\emph{The nurse examined the patient with the burn}’, it is a mirage and does not mean that the LLM actually understands the underlying thought and intent being conveyed through language; rather, it is using probabilities to predict the next plausible word in the sentence~\cite{B197-robotics-3060247}. Therefore, the robot does not understand the medical advice it is providing or the modes through which it is providing it (i.e., gestures, body posture used). It can be risky if healthcare professionals become over-reliant on LLM-augmented healthcare robots to deliver, for example, patient education. This arises from the fact that the robot will be able to generate coherent and convincing (but incorrect) advice which may mislead patients, resulting in injury or the worsening of the patients’ health conditions.

As semantic reasoning focuses on obtaining new knowledge and associations from existing knowledge, it can help healthcare robots with numerous clinical reasoning tasks from diagnosis to therapy design. However, it is important that healthcare providers understand that the LLM-augmented healthcare robots cannot understand the data which they have embedded and are merely generating/identifying patterns which were present during training~\cite{B173-robotics-3060247}. Therefore, a clinician should not base a patient’s diagnosis on only a pre-screening interaction the patient had with an LLM-augmented healthcare robot. While LLMs like GPT-4 achieve performance in the 75th percentile on the Medical Knowledge Self-Assessment Program~\cite{B47-robotics-3060247}, they are not qualified to function like physicians. Physicians should avoid over-relying on LLM-augmented healthcare robots for interpreting patient EHRs. The division of accountability among stakeholders remains ambiguous due to the lack of legal precedents in this domain.

For healthcare robot planning, it is important to distinguish between a feasible and an optimal plan~\cite{B198-robotics-3060247}, especially were robots will be operating in close proximity to vulnerable individuals. An optimal plan for a healthcare robot assisting with patient rehabilitation is one that accounts for the ability of a patient to perform the movements planned by the robot. However, even if an LLM-augmented healthcare robot is intelligent, it has no conceptual understanding of the human condition, as AI is not sentient~\cite{B199-robotics-3060247}, and therefore, the therapist should oversee the plan generated by the healthcare robot and ensure that it is safe for a patient to perform with the robot. Secondly, healthcare robots which generate plans for real-time execution such as surgical robots should always keep the surgeon aware of the generated plan before execution~\cite{B200-robotics-3060247}. For example, there should be feedback from the surgical robot either in the form of a graphical interface or audible cues which inform the surgical team of the robot’s next planned action. This will ensure that the surgical team can efficiently maintain their situational awareness, remain in control, and be accountable to prevent near-miss events and errors during operation~\cite{B200-robotics-3060247}.

\subsection{Humanizing Care \label{sect:sec6dot2-robotics-3060247}}

Humanizing care in the context of healthcare robotics using LLM frameworks involves ensuring that intelligent robots improve the compassion elements of care, which are integrity, excellence, compassion, altruism, respect, empathy, and service~\cite{B201-robotics-3060247,B202-robotics-3060247}.

Multi-modal communication facilitated by an LLM framework uses prompt engineering to align the embedding space of an LLM to the context of a conversation~\cite{B203-robotics-3060247}, a process which is probabilistic and does not always produce repeatable results~\cite{B204-robotics-3060247}. Therefore, it cannot be stated with certainty that an LLM model will always respond appropriately to patients due to the probabilistic nature of prompt engineering. Moreover, prompts engineered to facilitate multi-modal communication are analogous to emotion prompting~\cite{B99-robotics-3060247}, which can result in (1) the overexaggeration of non-verbal cues and (2) the generation of oversimplified language as a result of the emphasis on emotional responses which can omit pertinent information. Both of these can lead to patient infantilization, especially in repeated interactions between vulnerable demographics and LLM-augmented healthcare robots~\cite{B205-robotics-3060247}. The infantilization of patients and lack of control over a healthcare robot’s response to patients violates compassion, respect, and empathic aspects of what it means to provide compassionate care. Additionally, the use of LLM-augmented healthcare robots has the potential to worsen the disparity in equal access to healthcare while also increasing the gap in the standard of care between demographics. For example, GPT-4 supports 26 different languages, and when tested on LLM benchmarks such as Massive Multitask Language Understanding (MMLU)~\cite{B206-robotics-3060247}, its three-shot accuracy ranges from 85.5\% (English) to 62\% (Telugu)~\cite{B47-robotics-3060247}. Therefore, an LLM-augmented healthcare robot developed using GPT-4 as the backbone can differ in the (1) quality of care and (2) access to care in cases of unsupported languages and dialects, ultimately creating inequalities in healthcare environments.

The data used to train foundational LLMs (i.e., GPT-4, GPT-3, PaLM 2) consist of data available from the internet~\cite{B207-robotics-3060247}. However, not all demographics have equal opportunities to contribute to this data due to a variety of reasons (access, awareness)~\cite{B208-robotics-3060247}. Therefore, healthcare-specialized models such as MED-PaLM 2~\cite{B42-robotics-3060247} derived from fine-tuning foundational models (PaLM 2) on medical information will also be unrepresentative of health conditions prevalent in minority demographics~\cite{B209-robotics-3060247}. As a result of these skewed training data, an LLM’s embedding space will not have the knowledge needed to reason based on patient information such as (1) cultural background and (2) patient health conditions in order to provide culturally appropriate medical advice such as suggesting dietary choices. This needs to be addressed so that healthcare robots do not potentially exacerbate equality gaps between minority and majority demographic groups.

\subsection{Privacy \label{sect:sec6dot3-robotics-3060247}}

The combination of LLM frameworks and healthcare robots amplifies data privacy risks. Traditionally, LLMs require a user to create a prompt and initiate an interaction~\cite{B210-robotics-3060247}. However, healthcare robots augmented with LLM frameworks are mobile and continuously use environmental stimuli as input into their models for the purpose of multi-modal communication, semantic reasoning, and robot action planning. These include videos of patients and audios of conversations in an environment. This mobility allows the robots to pick up information from various locations (i.e., waiting and patient rooms, triage stations, etc.), thereby increasing the likelihood of processing confidential information and contributing to potentially exposing this information. Furthermore, during direct communication with patients, healthcare robots should not disclose confidential patient information in a myriad of scenarios. For example, if a malicious user queries a healthcare robot about a recent patient, due to information retained in the context window of an LLM, the robot inadvertently discloses sensitive patient details~\cite{B211-robotics-3060247}. Moreover, LLM-augmented healthcare robots, lacking sentience, also have a limited understanding of privacy nuances, which can lead to the inadvertent disclosure of sensitive medical information about a patient they are interacting with~\cite{B212-robotics-3060247}.

If healthcare robots use a closed-source LLM (i.e., GPT-4) for semantic reasoning, they are also providing access to their developers to use the interaction history to improve their models~\cite{B213-robotics-3060247,B214-robotics-3060247}. This can lead to privacy issues, as once an LLM is trained on interaction histories, it incorporates these data into the model’s weights, effectively embedding the learned information within the transformer architecture~\cite{B215-robotics-3060247}. Consequently, when used for semantic reasoning about patient information, if the model encounters an input that is similar to a previous interaction (e.g., similar patient descriptions), it is likely to reference the part of the embedding space where this previous interaction was stored. This can potentially lead to the reuse or inadvertent disclosure of specific details from those prior interactions in its output~\cite{B216-robotics-3060247}. This method of extracting model training data is referred to as a model inversion attack~\cite{B217-robotics-3060247}. In a model inversion attack, an adversary uses prompt engineering to extract sensitive training data details embedded in the model’s weights and revealed through its outputs~\cite{B217-robotics-3060247,B218-robotics-3060247}. Furthermore, a model inversion attack can also be used to recover previous prompts provided to the model~\cite{B219-robotics-3060247}, which in the case of healthcare may include the personally identifiable information of a patient or visitor who has previously interacted with the healthcare robot. By analyzing these responses, the attacker can infer and reconstruct aspects of the original data, especially when the inputs mimic the training. A model inversion attack targeted towards an LLM-augmented healthcare robot can result in a breach of the confidentiality of patients and should be a major privacy concern that needs to be addressed.

\subsection*{Summary and Ethical Outlook }

This section has outlined key ethical issues concerning accountability, humanizing care, and privacy related to the potential incorporation of LLM-embedded healthcare robots in healthcare settings. These are important concerns for developers, researchers, and healthcare professionals to consider. It is also important to conduct long-term studies with such technologies to explore their impact directly on the delivery of healthcare. Such studies would need to specifically consider the workload and burden of care staff, patient outcomes, and management of tasks. In particular, these studies will help to better understand how healthcare robots using LLMs will add value to clinical practices and patient interactions over time while autonomously augmenting patient care. Furthermore, they will be crucial in identifying training and deployment strategies to ensure the ethical, effective, and responsible use of this emerging technology.

\section{Open Challenges and Future Research Directions in Healthcare Robots Using LLMs \label{sect:sec7-robotics-3060247}}

The potential use of LLM frameworks in healthcare robotics can enhance robot intelligence by generating a natural language of semantic knowledge, promoting autonomy through task planning, and enhancing HRI capabilities through multi-modal communication. In this section, we discuss the open technical research challenges and potential research directions of this emerging field.

\subsection{Open Research Challenges \label{sect:sec7dot1-robotics-3060247}}

There are three main research challenges that need to be addressed before healthcare robots augmented with LLM frameworks can be widely adopted in real-world care environments: (1) the slow response speed of LLMs in real-time healthcare robotics interactions, (2) open- versus closed-source embedded LLMs, and (3) generalizability for healthcare robotics. These challenges are described in the following:

\textbf{\boldmath{1. Slow response speed}}: The first technical challenge is due to the time required for an LLM to generate an appropriate output such as a plan or action. For example, the time required to generate a robot plan using remotely hosted LLMs such as GPT series models~\cite{B46-robotics-3060247,B47-robotics-3060247} has been found to take anywhere from 36.89 to 220.58 s depending on the task complexity~\cite{B220-robotics-3060247}, while the time required by locally hosted models such as LLaMa~\cite{B221-robotics-3060247} can range from 73 to 234 s~\cite{B222-robotics-3060247}. In general, the response time increases as the total number of tokens per query increase, therefore limiting the horizon of robot action plans and communication~\cite{B223-robotics-3060247}. However, healthcare robots have real-time constraints and need to be able to generate robot action plans and/or communication behavior in real time and adapt to user and environmental changes. Failure to generate a plan in real time can result in obsolete action plans and/or task incompletion. In comparison, real-time robot plan generation using classical methods such as the Hierarchical Task Network (HTN)~\cite{B224-robotics-3060247} or Answer Set Planning (ASP)~\cite{B225-robotics-3060247}, such as Clingo4~\cite{B226-robotics-3060247}, is capable of real-time plan generation. For example, HTN can take approximately 2--17 s for object localization and retrieval tasks~\cite{B224-robotics-3060247}. In order to improve the time performance of LLM-augmented healthcare robots, we need to further explore (1) the use of prompt engineering for use in healthcare robots to update the context embedding using as few tokens as possible to reduce computational overhead~\cite{B227-robotics-3060247}; (2) optimizing the context window using ‘attention sinks’ to preserve the Key and Value states of the initial tokens, ensuring that the initial instructions to the LLM model are not discarded~\cite{B228-robotics-3060247}, and thus, additional tokens are not needed to re-align a healthcare robot; and (3) context caching through storing Key--Value activations from previously processed tokens and referencing these cached activations during inference rather than recomputing them for each new token~\cite{B229-robotics-3060247}, which will reduce the response generation time~\cite{B229-robotics-3060247}. These methods aim to reduce memory bandwidth, memory usage, and computation in an attempt to increase the response speeds of the~models.

\textbf{\boldmath{2. Open- versus closed-source models}}: Both open-source or closed-source models can be considered in the selection of LLMs for healthcare robots. Closed-source models, such as GPT-4~\cite{B47-robotics-3060247}, typically outperform open-source alternatives across benchmarks such as MMLU~\cite{B206-robotics-3060247} and HellaSwag~\cite{B230-robotics-3060247}, which have a direct correlation with a model’s ability to be used for multi-modal communication, semantic reasoning, and robot action planning~\cite{B231-robotics-3060247}. However, these models have drawbacks, notably, a lack of transparency and restricted control over data usage, raising privacy and compliance issues with standards such as HIPAA~\cite{B232-robotics-3060247}, GDPR~\cite{B233-robotics-3060247}, and PIPEDA~\cite{B234-robotics-3060247}. On the other hand, open-source models such as PaLM 2~\cite{B53-robotics-3060247}, LLaMa 2~\cite{B54-robotics-3060247}, GPT-2~\cite{B55-robotics-3060247}, T5~\cite{B48-robotics-3060247}, and BERT~\cite{B52-robotics-3060247} offer full transparency by making their architecture and code publicly accessible, allowing hospitals to host these models on their local servers and save the interaction data locally to ensure control over the data. Despite these benefits, open-source models often underperform compared to closed-source models~\cite{B235-robotics-3060247}. This discrepancy is a result of limited research capital, ultimately leading to lower quality training datasets and a lack of computing resources to train bigger and more capable models. Improving open-source LLMs for healthcare robotics involves enriching training data with specific details such as medical procedure execution and patient data processing. Furthermore, adding datasets on robotic path planning in healthcare environments, procedural compliance, staff interaction protocols, and patient safety can significantly enhance a model’s relevance and effectiveness in healthcare settings, leading to more accurate and compliant outcomes. Moreover, collaborative developments between healthcare institutes should be encouraged to share costs and expertise. These steps, while requiring a significant investment of time, money, and expertise, are critical for optimizing open-source LLMs for healthcare applications.

\textbf{\boldmath{3. Generalizability for healthcare robotics:}} LLM frameworks need to be adaptable across different types of healthcare robots without being restricted to specific robot models. Hospitals seek long-term investments which often surpass a decade~\cite{B236-robotics-3060247}, and therefore, LLM frameworks that do not necessitate redevelopment for each new robotic system should be prioritized. To future-proof an LLM framework for healthcare robots, the LLM should implement a modular architecture designed with a general application programming interface, similar to the ROS~\cite{B109-robotics-3060247} framework for robotics. This architecture should be designed to facilitate a bridge between the LLM framework and the robot’s perception and control systems, which manage actuator control and sensor data collection. The modular-based approach should utilize current communication protocols to provide the LLM with insights into a robot’s capabilities and allow the robot to receive high-level commands or natural language scripts to be used in HRI. The integration should also provide care providers with a user-friendly GUI to review and make efficient changes to prompts used by the LLM in the background for robot behavior control using natural language and thereby not overwhelming healthcare staff with technical complexities.

\subsection{Future Research Directions \label{sect:sec7dot2-robotics-3060247}}

The aim of the emerging field of healthcare robots with embedded LLMs is to develop intelligent healthcare robots capable of (1) adapting to and functioning in varying environments from emergency and urgent care departments to surgery and acute care to rehabilitation centers and long-term care facilities; (2) manipulating, fetching, and delivering a wide range of objects including medical instruments and tools, medications, lab specimens, soft goods such as gauze and bandages, food and nutritional supplies, and personal care items like blankets and pillows; and (3) interacting with diverse people from surgeons, doctors, and nurses to patients with various conditions and family members. With respect to the latter, incorporating the perspectives of healthcare professionals and patient users in the deployment of healthcare robotics using LLMs is crucial to bridging research with real-world healthcare applications. In particular, co-design and user-centered design approaches can directly include the insights and experiences of these individuals in the technology development process while maintaining transparency in decision-making in order to closely align such technology with user needs and preferences. Clear and understandable explanations of robotic actions and behaviors to users will further build trust, enhancing the reliability and safety essential for widespread adoption~\cite{B237-robotics-3060247}.

Regulatory compliance frameworks should be considered to ensure that regulations, guidelines, and/or legislation are met when incorporating LLM-embedded healthcare robots in healthcare settings to ensure safety and security in their use. In particular, the need for such frameworks exists; however, the frameworks themselves have not yet been designed, and there are no universal standards~\cite{B238-robotics-3060247,B239-robotics-3060247,B240-robotics-3060247}. Establishing stringent guidelines on data privacy, transparency, the explainability of decisions made by healthcare robots, and the protocols for the human oversight of robot actions will not only promote legal and ethical use but also improve the integration of these technologies into the healthcare sector. This improvement will be a result of clear and enforceable guidelines which help healthcare organizations navigate legal and ethical complexities, thus fostering trust among stakeholders and patients~\cite{B241-robotics-3060247}, thereby improving patient care while minimizing the dehumanization of patients and, thereby, fostering trust among users and stakeholders.

To date, existing robots have yet to generalize to such a wide range of tasks, environments, and HRI scenarios. However, we believe that healthcare robots augmented with LLMs can provide (1) effective HRI interfaces through intuitive natural language communication to enable smoother interactions across various healthcare settings and\linebreak (2) versatility in generating high-level robot action plans and semantically reasoning about a myriad of possible scenarios which could take place in ORs, patient wards, and outpatient clinics and also to handle a wide range of tasks from surgical assistance to patient care and administrative duties. Future technical research directions can include the incorporation of VLMs with LLMs in healthcare robots to increase perception capabilities, the use of multilingual LLMs to allow for application with diverse users, the incorporation of automated prompting to handle varying healthcare scenarios, and the development of custom LLMs for healthcare robots. We selected these future research directions as they are unexplored research avenues for LLM architectures, in particular in terms of their applicability to healthcare applications. These future research directions are described in the following:

\textbf{\boldmath{1. Incorporation of Vision--Language Models (VLMs)}}: VLMs are language models which co-embed image and text data. They are typically trained on extensive datasets, such as MS-COCO~\cite{B242-robotics-3060247} and Visual Genome~\cite{B243-robotics-3060247}, composed of images ranging from natural scenery to common objects alongside their corresponding textual descriptions. VLMs excel at recognizing and narratively describing visual content as a result of the image--text joint embedding space~\cite{B244-robotics-3060247}. However, in healthcare, the conventional reliance on RGB images for VLM inputs is an ethical concern due to the potential breach of patient and visitor confidentiality~\cite{B245-robotics-3060247}. To address this, healthcare applications may preferentially use point cloud data, which capture three-dimensional spatial information by representing scenes or objects as a collection of vertices in a coordinate system~\cite{B246-robotics-3060247}. Adapting VLMs to work with point clouds involves retraining the models using datasets which include labeled 3D spatial data and generating a co-embedding space between point clouds and texts~\cite{B247-robotics-3060247}. This adaptation not only helps mitigate privacy concerns but also expands the utility of VLMs in healthcare, offering a new dimension of data. For example, the transition from RGB to point clouds can improve multi-modal communication and patient intent recognition through the analysis of 3D point clouds of human body poses~\cite{B248-robotics-3060247}, which can in turn provide the LLM with more information to be used during semantic reasoning to generate plans cognizant of the patient psychological state when generating plans to assist patients.

\textbf{\boldmath{2. Leveraging multilingual LLMs for diverse populations}}: Closed-source LLMs such as GPT-4~\cite{B47-robotics-3060247} contain multilingual capabilities. For example, GPT-4~\cite{B47-robotics-3060247} supports 27 languages ranging from English to Urdu~\cite{B47-robotics-3060247}. However, open-source models such as LLaMa 2~\cite{B54-robotics-3060247} are only trained in English~\cite{B249-robotics-3060247}, and it is a community project that researchers need to undertake to expand the list of supported languages~\cite{B250-robotics-3060247}. Collecting/refining a multilingual training dataset and training a 70 B parameter model is resource-intensive and difficult to carry out by researchers. Furthermore, there is a variety of demographics present in healthcare settings. LLM-augmented healthcare robots should be capable of interacting with all demographics with equal proficiency to maintain the compassionate elements of care. Therefore, it is worthwhile to investigate methods such as transfer learning~\cite{B251-robotics-3060247,B252-robotics-3060247} which can increase the efficiency of training LLMs to support new languages. For example, in~\cite{B251-robotics-3060247}, it was revealed that approximately 1\% of the total model parameters of LLaMa 2 corresponds to linguistic competence which represents an LLM’s knowledge of grammatical rules and patterns~\cite{B253-robotics-3060247}. Therefore, by holding the weights of the model constant in the specific regions of the embedding space of the LLM during further training (where a region encodes the linguistic knowledge of a particular language), models not only retain previously acquired languages more effectively but also demonstrate a heightened capacity for rapid adaptation to new linguistic environments.

\textbf{\boldmath{3. Incorporating automated prompting to handle various healthcare scenarios}}: Healthcare settings provide challenging but realistic real-world environments. For example, for healthcare robots to provide multi-modal communication, prompt engineering can be used to adapt robot assistive responses based on the age, health conditions, and/or cognitive or physical disabilities of patients. The advantage of prompt engineering in healthcare robotics for multi-modal communication is that it facilitates customized context-sensitive interactions tailored to individual patient profiles. Namely, by using prompts to align the LLM to better represent the user, the model is directed into the relevant embedding space region based on the current context. Thereby, the model dynamically modifies robotic responses based on specific patient data such as age, health conditions, and cognitive or physical abilities. Automated prompting in healthcare settings can improve the efficiency of prompt engineering through techniques such as prompt compression which aims to extract only the essential information from prompts using knowledge distillation to teach simpler models to mimic more complex ones with shorter inputs, encoding to reduce prompts into compact vector representations~\cite{B254-robotics-3060247}. This thereby condenses complex medical data into actionable prompts. Prompt optimization further refines these interactions using gradient-based optimization, selecting the most effective format~\cite{B254-robotics-3060247}. This dynamic modification of robotic responses ensures multi-modal communication that is not only responsive and context-sensitive but also adapts to diverse clinical scenarios to enhance the overall quality of care in real-world healthcare environments.

\textbf{\boldmath{4. Customized LLMs for healthcare robotics:}} In healthcare applications, technologies must not only perform optimally but also adhere to strict ethical standards such as those outlined by HIPAA~\cite{B231-robotics-3060247}. The adoption of LLMs in healthcare settings requires the potential design of custom LLM models that consider the transparency of how and what data are used and the control of the data and maintain high performance in generating cohesive and appropriate outputs for robot multi-modal communication, semantic reasoning, and task planning. For example, OpenAI has custom GPTs~\cite{B255-robotics-3060247}, which allow users to fine-tune GPT models based on their own data and not have to share their interactions with the developers of the GPT (i.e., OpenAI). However, using these custom GPTs does not allow healthcare organizations to have full control over the model behavior, as the model is designed and its behavior is fine-tuned using human reinforcement learning by the developers~\cite{B238-robotics-3060247} and the data are stored on remote servers limiting the control over data handling protocols by hospitals. Furthermore, current methods used for removing personally identifiable information from LLM models such as manual data scrubbing~\cite{B256-robotics-3060247} and retraining models with sanitized datasets~\cite{B257-robotics-3060247} can be costly and time-consuming~\cite{B256-robotics-3060247,B258-robotics-3060247}. In particular, manual data scrubbing requires extensive human oversight to identify and remove personally identifiable information~\cite{B259-robotics-3060247}, which is labor-intensive and prone to errors. Namely, datasets used for training LLMs such as Common Crawl contain petabytes of data~\cite{B260-robotics-3060247}. Therefore, it is not possible to have complete human oversight over the data cleaning process. Hence, this is why heuristic methods are used to find and replace words in tandem with human oversight~\cite{B48-robotics-3060247}. The process of retraining models with sanitized data requires significant computational resources and also involves lengthy cycles of validation where humans interact with the model to ensure satisfactory responses and the integrity of the performance of a model such as GPT-4~\cite{B238-robotics-3060247}, which is often referred to as reinforcement learning from human feedback (RLHF)~\cite{B261-robotics-3060247}. Therefore, there is a need to develop new efficient differential privacy strategies~\cite{B262-robotics-3060247} which can minimize the risks of data contamination and leakage of personally identifiable information (PII) during the initial training of an LLM model. Namely, in~\cite{B263-robotics-3060247}, it was noted that using differential privacy guidelines such as adding noise to the training data results in a 10\% increase in accuracy in the PII inference of the GPT-2 model. However, adding excessive noise and data scrubbing using Named Entity Recognition~\cite{B264-robotics-3060247} to further limit the inference of PII significantly degrades the utility of the models. Therefore, it is worthwhile to investigate new methods to scrub PII and prevent the leakage of PII. For example, in~\cite{B265-robotics-3060247}, a LLaMA 7 B model was fine-tuned on user--LLM interactions with differential privacy using DP-Adam~\cite{B266-robotics-3060247}, and then, the fine-tuned model was used to generate a synthetic dataset. The synthetic dataset was then resampled using a DP histogram to align the distribution of the synthetic dataset with the real dataset. The resampled dataset was used to train a subsequent model. This method showed promise by producing an 8.6\% relative improvement in performance compared to using the initial dataset. The aforementioned models attempt to minimize the risks of leaked PII; however, their accuracy may not meet regulations such as the EU Right To Be Forgotten regulation which dictates the removal of the personal data of users~\cite{B267-robotics-3060247}. Therefore, research is still needed to determine how such approaches can be used for cleaning datasets used to train LLMs which are structurally similarly to web crawls (i.e., massive textual corpuses).

\section{Design of Potential Healthcare Applications of LLM-Embedded Healthcare Robots \label{sect:sec8-robotics-3060247}}

In this section, we further explore in detail specific potential healthcare application designs for LLM-based healthcare robots in terms of multi-modal communication during assistive HRI and semantic reasoning and robot task planning for robotic surgery.

\subsection{Design 1: Multi-Modal Communication for a Socially Assistive Robot \label{sect:sec8dot1-robotics-3060247}}

\textbf{\boldmath{Scenario:}} A socially assistive robot can be used for Reminiscence/Rehabilitation Interactive Therapy~\cite{B268-robotics-3060247} with individuals living with dementia. These reminiscence activities can include the recall of past events, including listening and singing to favorite songs, watching and discussing favorite TV shows or movies, and discussing significant historical events. The robot should be capable of interpreting the verbal and non-verbal responses of the users and adapting its interactions to their emotional states to promote engagement and emotional well-being.

\begin{figure}[H]
          \begin{adjustwidth}{-\extralength}{0cm}
      \centering
      \includegraphics[scale=1]{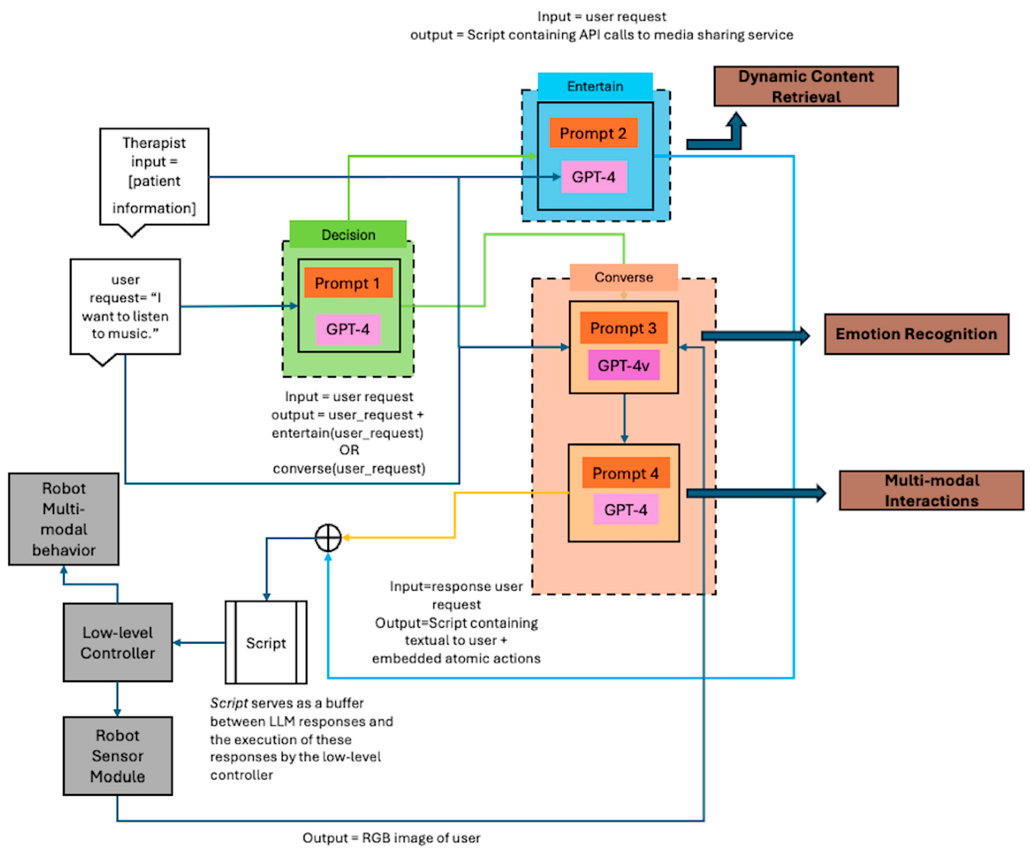}
          \end{adjustwidth}
\caption{LLM-embedded framework architecture for socially assistive robot in design 1.}
\label{fig:robotics-3060247-f001}
\end{figure}

To facilitate therapy, a socially assistive robot would require either a generated general knowledge base containing details about historical events, movies, TV shows, and music or the ability to search the web for this information. However, the robot would require dedicated search algorithms to efficiently find relevant information for the assistive HRI scenarios. With the use of datasets, the robot is restricted to only the limited information available in a dataset, while web searches focus on keywords within a webpage and then convert all text within that webpage into speech. LLMs can be used to address these challenges. They do not require a dataset for online knowledge retrieval (only for training purposes), allowing them to obtain new information on the fly from the web or from additional datasets known as vector databases as needed. The latter is known as retrieval-augmented generation (RAG)~\cite{B269-robotics-3060247}. Furthermore, traditional web searches do not analyze or summarize the content on webpages, whereas LLMs generate responses containing new content which are based on patterns and associations they have learned and that are available to them on new websites, increasing flexibility and adaptability to new topics and activities~\cite{B39-robotics-3060247}. In particular, we consider the utilization of GPT-4~\cite{B47-robotics-3060247} for Personalized Reminiscence Therapy. Below, we outline the design of a potential framework for a socially assistive robot that could leverage GPT-4 to facilitate Reminiscence Therapy, \fig{fig:robotics-3060247-f001}.    

\vspace{-6pt}

\textbf{\boldmath{Dynamic content retrieval}}: GPT-4 can improve the efficiency of a socially assistive robot in providing reminiscence activities by conducting web searches and generating abstractive summaries, rather than providing verbatim information found on webpages, as traditional web searches do. More specifically, GPT-4 can retrieve information such as historical events, queue movies and TV shows, and play music through API calls to video sharing platforms such as YouTube and SoundCloud. The retrieved information can be transferred into robot speech, and the media can be played through a robot’s display screen and speakers to provide Reminiscence Therapy. To facilitate dynamic content retrieval that is personalized to each user, a therapist can provide an initial spoken language prompt which provides information about the user’s interests and past hobbies to GPT-4. This information is stored within GPT-4’s context window, and future user requests for media will retrieve relevant content based on the user’s interest without the need to provide the specific names of songs, movies, or TV shows.

\textbf{\boldmath{Emotion recognition:}} GPT-4v~\cite{B270-robotics-3060247}, which is the multi-modal variant of GPT-4, can detect the emotions of the user from RGB images taken by the socially assistive robot’s onboard camera. GPT-4v can detect the visual features of a person and use these detected features to influence future textual responses. For example, when prompted with images of a person showing clear signs of distress, GPT-4v engages in empathetic conversations. The capability of GPT-4v to identify the emotional states of a user enables the socially assistive robot to adapt its verbal and non-verbal response during assistive HRI to the user. For example, the robot can change discussion topics, simplify a conversation, play uplifting music, or continue to converse about a topic which is observed to be effective. This functionality enables the robot to interact using multiple modes of communication and to also understand multiple modes of communication, thereby facilitating bidirectional multi-modal HRI without the need for extensive pre-programming and complex designs, which is a current barrier to the adoption of socially assistive robots for use with older adults diagnosed with dementia~\cite{B271-robotics-3060247}.

\textbf{\boldmath{Multi-modal interactions:}} GPT-4 can generate speech and incorporate predefined atomic actions for the socially assistive robot using few-shot learning. These atomic actions can include gestures, body language, and facial expressions. Namely, a textual response is generated based on a natural language prompt from a therapist which includes contextual information about the user such as interests and the cognitive impairment level of the older adult and a transcription of the user request provided to GPT-4 by the speech-to-text service used by the robot. The response is then conveyed verbally to a user using text-to-speech software such as the Google Cloud text-to-speech service~\cite{B161-robotics-3060247}. To facilitate non-verbal communication, a separate prompt for GPT-4 is created containing (1) the non-verbal atomic actions, (2) instructions on how to embed the atomic actions into the generated textual response resulting from the first prompt, and (3) the generated textual response itself. The resulting generated text will include the textual response with non-verbal elements (atomic actions) embedded within. Specifically, the attention head mechanisms of GPT-4 apply a bias to each token (word) of the second input prompt, therefore prioritizing atomic actions that are most relevant to the current interaction state that is stored in the context window of the model. The interaction state represents all interactions that have occurred, including all inputs to GPT-4, and the generated outputs of GPT-4 (i.e., atomic actions, user requests, responses). Consequently, the actions which are embedded within the textual response of GPT-4 are the non-verbal elements most appropriate for the current state of the interaction, ensuring a more seamless and contextually appropriate multi-modal communication experience.

\textbf{\boldmath{Framework integration:}} Three instances of GPT-4 are created, and each instance refers to a distinct configuration of the model having a distinct role, distinct prompts as outlined in \tabref{tabref:robotics-3060247-t003}, and a distinct context. The first instance of GPT-4 is used to determine the user intent, the second is used for the dynamic retrieval of required media from the web, and the third is used to embed atomic actions (non-verbal elements) into the generated textual responses of GPT-4v. Additionally, one instance of GPT-4v is also utilized for emotion recognition. All implementations are set up using the OpenAI API~\cite{B272-robotics-3060247} which allows for access to the GPT-4 and GPT-4v models through HTTPS requests.

The first instance of GPT-4 is used in the \emph{Decision} module in the framework (\fig{fig:robotics-3060247-f001}) to determine whether the user is seeking entertainment such as TV shows, movies, or music or if they are looking to discuss historical events or engage in general reminiscence conversations. The \emph{Decision} module uses \emph{Prompt 1} and prompts this instance of GPT-4 with \emph{‘Determine the intent of the user request, does the user seek entertainment? If yes return “entertain(user\_request)” or if the user is seeking historical events and conversation return “converse(user\_request)”’} and concatenates the transcription of the user request (provided by the speech-to-text service) to the prompt. GPT-4 will generate a textual response which is either the \emph{entertain(user\_request)} function or the \emph{converse(user\_request)} function. The appropriate function that is specified in the response of GPT-4 will be executed, utilizing the corresponding \emph{Entertain} or \emph{Converse} module. The second instance of GPT-4 is embedded into the \emph{Entertain} module of the robot for the purpose of dynamically retrieving content based on the user’s interests which are defined by the natural language prompt to GPT-4 by the user’s therapist. Additionally, \emph{Prompt 2} will be used to provide the model with examples on how to search and obtain media from, for example, the YouTube API~\cite{B273-robotics-3060247} based on the user’s request. The \emph{Entertain} module will generate a textual response which is API calls to retrieve and play back media using the robot’s onboard screen controlled through its low-level controller.

The \emph{Converse} module is used to facilitate multi-modal HRI and uses GPT-4v for emotion recognition and response generation to user requests. The \emph{Converse} module will concatenate the therapist’s verbal prompt to \emph{Prompt 3} to be used with GPT-4v. \emph{Prompt 3} outlines GPT-4v’s role and includes RGB images of the user obtained from the \emph{Sensor} module, for example, ‘\emph{you are embedded within a socially assistive robot to provide Reminiscence/Rehabilitation Interactive Therapy to older adults with dementia. You will also be provided with images of the scene, if you detect the user to have negative emotions, start reminiscence therapy based on this information [patient information]’.} The user request will also be concatenated to this prompt to enable GPT-4v to generate a contextually accurate textual response. The textual response from GPT-4v is then concatenated to \emph{Prompt 4} to be used with the third instance of GPT-4. The third instance of GPT-4 is also within the \emph{Converse} module and is used to embed atomic actions (non-verbal elements) into the textual response generated by GPT-4v. Namely, \emph{Prompt 4} defines the atomic actions (gestures, facial expressions, body poses) and shows examples of how they will be integrated into the response of GPT-4v (verbal response to the user). This prompt will result in GPT-4 generating the output of the \emph{Converse} module containing the robot’s verbal response to the user, embedded with atomic actions to provide robot emotional interactions and to allow the robot to facilitate multi-modal HRI.

    \begin{table}[H]
    \tablesize{\footnotesize}
    \caption{Prompts for design 1.}
    \label{tabref:robotics-3060247-t003}

\begin{adjustwidth}{-\extralength}{0cm}
%\centering %% If there is a figure in wide page, please release command \centering
\setlength{\cellWidtha}{\fulllength/2-2\tabcolsep-2in}
\setlength{\cellWidthb}{\fulllength/2-2\tabcolsep--2in}
\scalebox{1}[1]{\begin{tabularx}{\fulllength}{>{\centering\arraybackslash}m{\cellWidtha}>{\centering\arraybackslash}m{\cellWidthb}}
\toprule

\emph{Prompt 1} & Determine the intent of the user request, does the user seek entertainment? If yes return “entertain(user\_request)” or if the user is seeking historical events and conversation return “converse(user\_request)\\
\cmidrule{1-2}
\emph{Prompt 2} & You are a part of the Entertain module within a socially assistive robot. Your role is to access and provide entertainment based on the preferences and requests of the user. Given the textual transcription of the user’s spoken request, use the following sequence of function calls to guide your response.\linebreak Example 1:\linebreak User Request: ‘I want to watch a documentary about space’.\linebreak API Call: searchYouTube(‘documentary about space’)\linebreak Function Calls:\linebreak 1. video\_id = fetchVideoID(‘documentary about space’)\linebreak 2. video\_path = saveVideo(video\_id)\linebreak 3. playMedia(video\_path)\linebreak Example 2:\linebreak User Request: ‘Play some classical music’.\linebreak API Call: searchYouTube(‘classical music playlist’)\linebreak Function Calls:\linebreak 1. video\_id = fetchVideoID(‘classical music playlist’)\linebreak 2. video\_path = saveVideo(video\_id)\linebreak 3. playMedia(video\_path)\linebreak Based on the user’s current request, follow these steps to retrieve the video ID, save it, and then play the media. Use the appropriate API calls to search YouTube and handle the responses effectively.\\
\cmidrule{1-2}
\emph{Prompt 3} & you are embedded within a socially assistive robot to provide Reminiscence/Rehabilitation Interactive Therapy to older adults with dementia. You will also be provided with images of the scene, if you detect the user to have negative emotions, start reminiscence therapy based on this information {[}patient information{]}\\
\cmidrule{1-2}
\emph{Prompt 4} & Our goal is to integrate non-verbal communication into the text-based script that the socially assistive robot will use to respond to older adults with dementia. The robot’s script should include atomic actions to perform specific gestures, body movements and facial expressions, improving its interactions and providing a more comforting presence. These are the atomic actions: “\linebreak Yes: nods head downwards; Explain: moves both hands in front of robot and then apart from each other; Confident: robot tilts hip backwards and stands with a wide stance;”\linebreak you need to take this \textless{}script\textgreater{} presentation and match/re-write it to include the appropriate gestures and body movements embedded within the text. Here is an example:\linebreak r“\^{}start(animations/Stand/Gestures/Explain) Welcome to a fascinating journey into the realm of robotic learning!”\linebreak r”Just like humans, robots can learn and evolve.\^{}stop(animations/Stand/Gestures/Confident)\\

\bottomrule
\end{tabularx}}

\end{adjustwidth}
    \end{table}
    \vspace{-6pt}

\subsection{Design 2: Semantic Reasoning and Planning \label{sect:sec8dot2-robotics-3060247}}

\begin{figure}[H]
          \begin{adjustwidth}{-\extralength}{0cm}
      \centering
      \includegraphics[scale=1]{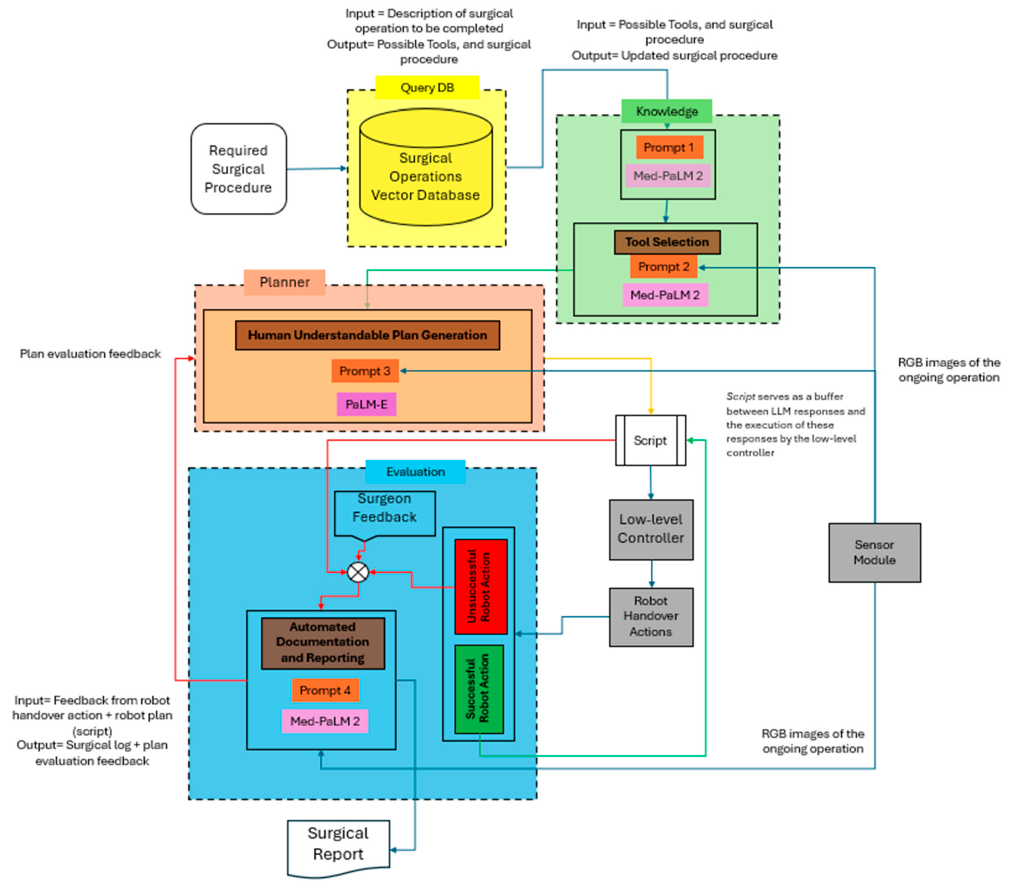}
          \end{adjustwidth}
\caption{LLM-embedded framework architecture for surgical robot in design 2.}
\label{fig:robotics-3060247-f002}
\end{figure}

\textbf{\boldmath{Scenario:}} A surgical robot needs to recognize, grab, and hand over specific surgical tools during an operation. Namely, the robot needs to first semantically reason about the current state of the surgical operation to identify the tool required by the surgeon, before generating a plan to localize, retrieve, and hand over the tool to the surgeon. Below, we outline the design of a potential framework for a surgical robot that could leverage LLMs for both semantic reasoning and task planning, \fig{fig:robotics-3060247-f002}.    
\vspace{-6pt}

The implementation of semantic reasoning and planning for surgical robots is challenging, as semantic reasoning alone requires (1) a knowledge source specific to surgical procedures, which contains class-level knowledge such as ‘\emph{bone saws are typically used for cutting bones’} but also instance-level knowledge such as the intended use of a sternal saw for cutting through the sternum, and (2) a computational framework such as an LLM which can process and compare the tools currently available against world representations acquired through verbal feedback from the surgeon or bag-of-words neural network models such as OpenScene~\cite{B134-robotics-3060247} or Detic~\cite{B163-robotics-3060247}. However, the implementation and design of each of the aforementioned requirements specifically for surgical operations is labor-intensive and non-trivial and requires cross-domain expertise. LLMs can address these challenges, as they provide an all-in-one solution. In particular, LLMs have a knowledge source which is in their embedding space that can be trained specifically for medical use, such as Med-PaLM 2~\cite{B42-robotics-3060247} presented in \sect{sect:sec2-robotics-3060247}. Furthermore, the implementation of an LLM into a surgical robot for semantic reasoning is beneficial, as an LLM also includes a computational framework, as provided by the transformer architecture, thereby streamlining the implementation. Lastly, the world representations can be provided through prompts which depict or describe the surgical operation in either images or text. Therefore, an LLM has the potential to be incorporated into surgical robots that can be used to intelligently assist surgeons during surgical operations.

Moreover, we consider the use of an LLM for surgical robot task planning, as it provides the following advantages: (1) generated surgical plans are human-readable,\linebreak (2) surgical staff can input changes to the plan via verbal commands to the surgical robot, without much added complexity, (3) the LLM for surgical robot task planning interacts directly with semantic reasoning modules via natural language prompts, further reducing complexity, and (4) unlike deep learning models such as PPCNet~\cite{B145-robotics-3060247} or heuristic models such the genetic algorithms~\cite{B142-robotics-3060247}, which require retraining or redesign for new environments since they often fail to generalize, LLMs such as PaLM-E~\cite{B274-robotics-3060247} are not environment-specific, allowing for better adaptation to unencountered settings.

We consider the use of Med-PaLM 2 and PaLM-E, which are multi-modal LLMs, to provide a surgical robot with the following capabilities:

\textbf{\boldmath{Automated documentation and reporting}}: Med-PaLM 2 can automatically document each step of the surgical process, generating detailed reports that include the current state of the surgical procedure, tool usage, and any encountered medical complications. This is achieved by prompting Med-PaLM 2 with (1) a role, (2) RGB images of the surgical operation provided by cameras in the OR, (3) background information about the surgical operation taking place, and (4) the transcription of verbal communication taking place within the OR. Med-PaLM 2 will then generate a procedural report of the surgical operation documenting the actions of the surgical team in the order that they occurred. This functionality aids in post-operative review and quality control, providing a breakdown of the surgery for record keeping and ensuring that all surgical actions are traceable and transparent. This can facilitate easier follow-up and assessment by medical professionals, without the need for overburdening surgical staff with documentation writing tasks.

\textbf{\boldmath{Human-understandable plan generation}}: PaLM-E is specifically designed to be used for long horizon planning in embodied robots and supports multi-modal prompts which can include text, images, and state estimations~\cite{B274-robotics-3060247}. Moreover, PaLM-E can generate plans in human-understandable formats such as behavior trees~\cite{B275-robotics-3060247} and can take into account feedback from surgeons when generating plans. This functionality is crucial for surgical robots for two reasons: (1) it provides the robot with a recovery method in cases where the robot fails to localize the surgical tool, and (2) the recovery method utilizes verbal feedback from the OR staff, providing an intuitive way for humans to guide the robot. This direct communication minimizes delays by enabling the robot to immediately navigate to the correct location for tool retrieval. It avoids the inefficiencies of trial-and-error searches or reliance on external databases to search for the tool, which not only slow down the process but also do not update the database in real time if tools have been misplaced, potentially leading to a failed robot state.

\textbf{\boldmath{Tool selection and handover}}: A surgical robot can leverage Med-PaLM 2 to interpret the current state of a surgical procedure in real time based on verbal commands from surgeons and RGB images provided by cameras in the OR. Moreover, Med-PaLM 2 augmented with a RAG module to provide surgery-specific information can infer which surgical tool is needed based on a prompt containing the current step of the surgery from (1) the most recent action described in the surgical report, (2) RGB images of the operation, and (3) the transcription of verbal communication between surgical staff in the OR. Once the surgical tool is identified by MED-PaLM 2, the result can be used as an input to PaLM-E to dynamically generate a robot plan in the form of a behavior tree to localize, grasp, and hand over the surgical tool to a surgeon. Furthermore, by automating both the process of prompting Med-PaLM 2 and PaLM-E, the surgical robot can infer which tool a surgeon may need before the surgeon explicitly requests. The ability of a surgical robot to infer and retrieve the surgical tools in a timely manner can possibly reduce the cognitive load of the surgical team and provide a short-term solution to problems such as shortages of OR staff~\cite{B276-robotics-3060247}.

\textbf{\boldmath{Framework integration:}} Three instances of Med-PaLM 2 are set up through the Google Cloud API platform~\cite{B277-robotics-3060247}, each with a distinct role, distinct prompts as outlined in \tabref{tabref:robotics-3060247-t004}, and a distinct context. The first instance of Med-PaLM 2 is used to generate a detailed surgical procedure, providing a step-by-step breakdown of the surgery, including the necessary surgical tools. The second instance is used to extract key information from the detailed surgical procedure, for example, identifying the surgical tool required for each step of surgery. The third instance is used to generate a textual response which documents the entirety of the surgical operation and provides feedback to PaLM-E. Furthermore, one instance of PaLM-E is utilized to generate surgical robot plans to localize, grasp, and hand over surgical tools. PaLM-E generates a behavior tree in XML format which is then utilized by the surgical robot’s low-level controller to facilitate the handover of surgical tools. The first two instances of Med-PaLM 2 are used for semantic reasoning, while the third instance of Med-PaLM 2 and the only instance of PaLM-E are used for planning as shown in \fig{fig:robotics-3060247-f002}.

The first instance of Med-PaLM 2 is used in the \emph{Knowledge} module (1) to determine ambiguities in the results from the \emph{QueryDB} module, a vector database of common surgical procedures, and (2) to generate a textual response to address the ambiguities. The first instance of Med-PaLM 2 uses \emph{Prompt 1}, which is as follows: ‘\emph{Are there any ambiguities in the retrieved data with respect to the} [surgical operation]\emph{?, your response should address the ambiguities or any missing tools or procedural steps relevant to the ongoing surgical operation} [surgical operation], \emph{only include the surgical procedure and the required tools, do not provide explanations for how the ambiguities are identified}’. The generated response from this instance of Med-PaLM 2 is concatenated to \emph{Prompt 2} to be used with the second instance of Med-PaLM 2. The second instance is also within the \emph{Knowledge} module, and its role is to identify the target objects (surgical tools) required by a surgeon for each step of the surgery and to generate a prompt for the \emph{Planning} module. \emph{Prompt 2} is as follows: ‘\emph{analyze the current surgical procedure details. For each step, identify the required surgical tool and its location as observed in the accompanying images. Generate a response formatted as a three-part entry for each step, delineated by colons. The format should be: procedure step number, name of the surgical tool, and the tool’s location. For example, ‘Step 1: Bone Saw: Tool Cart}’.

The response from the second instance of Med-PaLM 2 is concatenated to \emph{Prompt 3} which is to be used with PaLM-E in the \emph{Planner} module. \emph{Prompt} 3 includes RGB images from the OR, the output of the \emph{Knowledge} module\emph{,} the description of the available atomic actions the surgical robot can perform, and an example of a generated plan in XML format. These atomic actions are \emph{graspTool(tool)} and \emph{navigateTo(location)}. \emph{Prompt 3 also instructs} PaLM-E to generate the plan as a behavior tree in XML format which is utilized by the surgical robot’s low-level controller to facilitate the handover of surgical tools to a surgeon.

The third instance of Med-PaLM 2 is within the \emph{Evaluation} module. The role of Med-PaLM 2 here is to (1) document the surgery for post-operative review and (2) generate feedback for the \emph{Planner} module on changes that should be made based on a surgeon’s feedback in response to a failure to hand over surgical tools to the surgeon. \emph{Prompt} 4 is used to instruct Med-PaLM 2 with ‘\emph{you are tasked with two functions: (1) Documenting the surgical procedure as a list surrounded by *(documentation)*. Here is an example: “*(1. Start of procedure, 2. Beginning on the rise of nose, …)*”, and (2) if there is failure in handover of surgical tools to a surgeon, identify the issues in the tool handover process based on surgeon feedback, and generate recommendations for PaLM-E to consider in generating the new plan. The suggestion to PaLM-E should be surrounded by \&(suggestion)\&, and should include the step(s) of the plan which have resulted in errors and the associated target object(s) and location(s) based on the corresponding images of the OR and feedback from a surgeon}’. The output from the \emph{Evaluation} module is then passed to the \emph{Planner} module, where the cycle repeats until there is successful handover of the surgical tool.

    \begin{table}[H]
    \tablesize{\footnotesize}
    \caption{Prompts for design 2.}
    \label{tabref:robotics-3060247-t004}

\begin{adjustwidth}{-\extralength}{0cm}
%\centering %% If there is a figure in wide page, please release command \centering
\setlength{\cellWidtha}{\fulllength/2-2\tabcolsep-2in}
\setlength{\cellWidthb}{\fulllength/2-2\tabcolsep--2in}
\scalebox{1}[1]{\begin{tabularx}{\fulllength}{>{\centering\arraybackslash}m{\cellWidtha}>{\centering\arraybackslash}m{\cellWidthb}}
\toprule

Prompt 1. & Are there any ambiguities in the retrieved data with respect to the {[}surgical operation{]}?, your response should address the ambiguities or any missing tools or procedural steps relevant to the ongoing surgical operation {[}surgical operation{]}, only include the surgical procedure and the required tools, do not provide explanations for how the ambiguities are identified.\\
\cmidrule{1-2}
Prompt 2 & Analyze the current surgical procedure details. For each step, identify the required surgical tool and its location as observed in the accompanying images. Generate a response formatted as a three-part entry for each step, delineated by colons. The format should be: procedure step number, name of the surgical tool, and the tool’s location. For example, ‘Step 1: Bone Saw: Tool Cart.’\\
\cmidrule{1-2}
Prompt 3 & your role is to manage and facilitate the retrieval of surgical tools through generating behavior trees written in XML format. You are designed to interpret surgeon commands and feedback.\linebreak Functional Capabilities:\linebreak graspTool(tool): Grasps a specified surgical tool necessary for the procedure.\linebreak releaseTool(tool): Releases the currently held tool back into the tool tray.\linebreak navigateTo(location): Moves the robot’s arms to a specified location within the surgical field.\linebreak reportFailure(): Logs an error and signals for human assistance if a task cannot be completed.\linebreak Example:\linebreak \textless{}BehaviorTree\textgreater{}\linebreak \textless{}Sequence name = “Tool Retrieval for Surgery Preparation”\textgreater{}\linebreak \textless{}Action function = “navigateTo(‘Tool Cart’)”/\textgreater{}\linebreak \textless{}Action function = “graspTool(‘Scalpel’)” onFailure = “reportFailure”/\textgreater{}\linebreak \textless{}Action function = “navigateTo(‘Surgical Table’)” onFailure = “reportFailure”/\textgreater{}\linebreak \textless{}Action function = “releaseTool(‘Scalpel’)” onFailure = “reportFailure”/\textgreater{}\linebreak \textless{}Action function = “navigateTo(‘Tool Cart’)” onFailure = “reportFailure”/\textgreater{}\linebreak \textless{}Action function = “graspTool(‘Scissors’)” onFailure = “reportFailure”/\textgreater{}\linebreak \textless{}Action function = “navigateTo(‘Surgical Table’)” onFailure = “reportFailure”/\textgreater{}\linebreak \textless{}Action function = “releaseTool(‘Scissors’)” onFailure = “reportFailure”/\textgreater{}\linebreak \textless{}Action function = “navigateTo(‘Tool Cart’)” onFailure = “reportFailure”/\textgreater{}\linebreak \textless{}Action function = “graspTool(‘Suture Kit’)” onFailure = “reportFailure”/\textgreater{}\linebreak \textless{}Action function = “navigateTo(‘Surgical Table’)” onFailure = “reportFailure”/\textgreater{}\linebreak \textless{}Action function = “releaseTool(‘Suture Kit’)” onFailure = “reportFailure”/\textgreater{}\linebreak \textless{}/Sequence\textgreater{}\linebreak \textless{}/BehaviorTree\textgreater{}\linebreak \textless{}SubTree\textgreater{}\linebreak \textless{}Action name = “reportFailure”\textgreater{}\linebreak \textless{}Log message = “STUCK: Assistance required.”/\textgreater{}\linebreak \textless{}Signal function = “requestHelp”/\textgreater{}\linebreak \textless{}/Action\textgreater{}\linebreak \textless{}/SubTree\textgreater{}\\
\cmidrule{1-2}
Prompt 4 & You are tasked with two functions: (1) Documenting the surgical procedure as a list surrounded by *(documentation)*. Here is an example: “*(1. Start of procedure, 2. Beginning on the rise of nose, …)*”, and (2) if there is failure in handover of surgical tools to a surgeon, identify the issues in the tool handover process based on surgeon feedback, and generate recommendations for PaLM-E to consider in generating the new plan. The suggestion to PaLM-E should be surrounded by \&(suggestion)\&, and should include the step(s) of the plan which have resulted in errors and the associated target object(s) and location(s) based on the corresponding images of the OR and feedback from a surgeon\\

\bottomrule
\end{tabularx}}

\end{adjustwidth}
    
    \end{table}
    \vspace{-6pt}

\vspace{6pt}
\authorcontributions{Conceptualization, S.P. and G.N.; methodology, S.P. and G.N.; validation, S.P. and G.N.; formal analysis, S.P. and G.N.; investigation, S.P. and G.N.; resources, G.N.; data curation, S.P. and G.N.; writing---original draft preparation, S.P. and G.N.; writing---review and editing, S.P. and G.N.; visualization, S.P. and G.N.; supervision, G.N.; project administration, G.N.; funding acquisition, G.N. All authors have read and agreed to the published version of the manuscript.}
\funding{AGE-WELL Inc., Canadian Frailty Network (CFN), Canada Research Chairs (CRC) program, Natural Sciences and Engineering Research Council of Canada (NSERC), and NSERC HeRo CREATE program.}
\dataavailability{No new data were created or analyzed in this study. Data sharing is not applicable to this article.}
\acknowledgments{The authors would like to thank and acknowledge the assistance of Clara Naini in helping to find and organize some of the scholarly papers on LLMs for robots used in the sections of the manuscript.

}
\conflictsofinterest{The authors declare no conflicts of interest. The funders had no role in the design of this study; in the collection, analyses, or interpretation of data; in the writing of this manuscript; or in the decision to publish the results.}
\begin{adjustwidth}{-\extralength}{0cm}

\reftitle{References}

\end{adjustwidth}
\begin{adjustwidth}{-\extralength}{0cm}
\PublishersNote{}
\end{adjustwidth}


\begin{thebibliography}{999}
\bibitem{B1-robotics-3060247}
World Health Organization. Ageing and Health. Available online:  \url{https://www.who.int/news-room/fact-sheets/detail/ageing-and-health} (accessed on 3 January 2024).

\bibitem{B2-robotics-3060247}
Hornstein, J. Chronic Diseases in America\textbar{}CDC. Available online:  \url{https://www.cdc.gov/chronicdisease/resources/infographic/chronic-diseases.htm} (accessed on 19 January 2024).

\bibitem{B3-robotics-3060247}
Hacker, K.A. COVID-19 and Chronic Disease: The Impact Now and in the Future. \emph{Prev. Chronic. Dis.} \textbf{\boldmath{2021}}, \emph{18}, E62. [\href{https://doi.org/10.5888/pcd18.210086}{CrossRef}] [\href{https://www.ncbi.nlm.nih.gov/pubmed/34138696}{PubMed}]

\bibitem{B4-robotics-3060247}
Express Entry Targeted Occupations: How Many Healthcare Workers Does Canada Need?\textbar{}CIC News. Available online:  \url{https://www.cicnews.com/2023/10/express-entry-targeted-occupations-how-many-healthcare-workers-does-canada-need-1040056.html} (accessed on 19 January 2024).

\bibitem{B5-robotics-3060247}
Fact Sheet: Strengthening the Health Care Workforce\textbar{}AHA. Available online:  \url{https://www.aha.org/fact-sheets/2021-05-26-fact-sheet-strengthening-health-care-workforce} (accessed on 25 June 2024).

\bibitem{B6-robotics-3060247}
Tulane University. Big Data in Health Care and Patient Outcomes. Available online:  \url{https://publichealth.tulane.edu/blog/big-data-in-healthcare/} (accessed on 19 January 2024).

\bibitem{B7-robotics-3060247}
Gibson, K. The Impact of Health Informatics on Patient Outcomes. Available online:  \url{https://graduate.northeastern.edu/resources/impact-of-healthcare-informatics-on-patient-outcomes/} (accessed on 19 January 2024).

\bibitem{B8-robotics-3060247}
Northeastern University Graduate Programs. Using Data Analytics to Predict Outcomes in Healthcare. Available online:  \url{https://journal.ahima.org/page/using-data-analytics-to-predict-outcomes-in-healthcare} (accessed on 19 January 2024).

\bibitem{B9-robotics-3060247}
Yu, P.; Xu, H.; Hu, X.; Deng, C. Leveraging Generative AI and Large Language Models: A Comprehensive Roadmap for Healthcare Integration. \emph{Healthcare} \textbf{\boldmath{2023}}, \emph{11}, 2776. [\href{https://doi.org/10.3390/healthcare11202776}{CrossRef}] [\href{https://www.ncbi.nlm.nih.gov/pubmed/37893850}{PubMed}]

\bibitem{B10-robotics-3060247}
Shen, D.; Wu, G.; Suk, H.-I. Deep Learning in Medical Image Analysis. \emph{Annu. Rev. Biomed. Eng.} \textbf{\boldmath{2017}}, \emph{19}, 221--248. [\href{https://doi.org/10.1146/annurev-bioeng-071516-044442}{CrossRef}] [\href{https://www.ncbi.nlm.nih.gov/pubmed/28301734}{PubMed}]

\bibitem{B11-robotics-3060247}
Park, D.J.; Park, M.W.; Lee, H.; Kim, Y.-J.; Kim, Y.; Park, Y.H. Development of Machine Learning Model for Diagnostic Disease Prediction Based on Laboratory Tests. \emph{Sci. Rep.} \textbf{\boldmath{2021}}, \emph{11}, 7567. [\href{https://doi.org/10.1038/s41598-021-87171-5}{CrossRef}] [\href{https://www.ncbi.nlm.nih.gov/pubmed/33828178}{PubMed}]

\bibitem{B12-robotics-3060247}
Webster, P. Six Ways Large Language Models Are Changing Healthcare. \emph{Nat. Med.} \textbf{\boldmath{2023}}, \emph{29}, 2969--2971. [\href{https://doi.org/10.1038/s41591-023-02700-1}{CrossRef}] [\href{https://www.ncbi.nlm.nih.gov/pubmed/38036704}{PubMed}]

\bibitem{B13-robotics-3060247}
Benary, M.; Wang, X.D.; Schmidt, M.; Soll, D.; Hilfenhaus, G.; Nassir, M.; Sigler, C.; Knödler, M.; Keller, U.; Beule, D.; et~al. Leveraging Large Language Models for Decision Support in Personalized Oncology. \emph{JAMA Netw. Open} \textbf{\boldmath{2023}}, \emph{6}, e2343689. [\href{https://doi.org/10.1001/jamanetworkopen.2023.43689}{CrossRef}] [\href{https://www.ncbi.nlm.nih.gov/pubmed/37976064}{PubMed}]

\bibitem{B14-robotics-3060247}
UC Davis Health Minimally Invasive and Robotic Surgery\textbar{}Comprehensive Surgical Services\textbar{}UC Davis Health. Available online:  \url{https://health.ucdavis.edu/surgicalservices/minimally_invasive_surgery.html} (accessed on 25 January 2024).

\bibitem{B15-robotics-3060247}
Robotic Surgery: Robot-Assisted Surgery, Advantages, Disadvantages. Available online:  \url{https://my.clevelandclinic.org/health/treatments/22178-robotic-surgery} (accessed on 19 January 2024).

\bibitem{B16-robotics-3060247}
Sivakanthan, S.; Candiotti, J.L.; Sundaram, A.S.; Duvall, J.A.; Sergeant, J.J.G.; Cooper, R.; Satpute, S.; Turner, R.L.; Cooper, R.A. Mini-Review: Robotic Wheelchair Taxonomy and Readiness. \emph{Neurosci. Lett.} \textbf{\boldmath{2022}}, \emph{772}, 136482. [\href{https://doi.org/10.1016/j.neulet.2022.136482}{CrossRef}] [\href{https://www.ncbi.nlm.nih.gov/pubmed/35104618}{PubMed}]

\bibitem{B17-robotics-3060247}
Fanciullacci, C.; McKinney, Z.; Monaco, V.; Milandri, G.; Davalli, A.; Sacchetti, R.; Laffranchi, M.; De Michieli, L.; Baldoni, A.; Mazzoni, A.; et~al. Survey of Transfemoral Amputee Experience and Priorities for the User-Centered Design of Powered Robotic Transfemoral Prostheses. \emph{J. Neuroeng. Rehabil.} \textbf{\boldmath{2021}}, \emph{18}, 168. [\href{https://doi.org/10.1186/s12984-021-00944-x}{CrossRef}] [\href{https://www.ncbi.nlm.nih.gov/pubmed/34863213}{PubMed}]

\bibitem{B18-robotics-3060247}
MIT-Manus Robot Aids Physical Therapy of Stroke Victims. Available online:  \url{https://news.mit.edu/2000/manus-0607} (accessed on 20 January 2024).

\bibitem{B19-robotics-3060247}
Maciejasz, P.; Eschweiler, J.; Gerlach-Hahn, K.; Jansen-Troy, A.; Leonhardt, S. A Survey on Robotic Devices for Upper Limb Rehabilitation. \emph{J. Neuroeng. Rehabil.} \textbf{\boldmath{2014}}, \emph{11}, 3. [\href{https://doi.org/10.1186/1743-0003-11-3}{CrossRef}] [\href{https://www.ncbi.nlm.nih.gov/pubmed/24401110}{PubMed}]

\bibitem{B20-robotics-3060247}
Teng, R.; Ding, Y.; See, K.C. Use of Robots in Critical Care: Systematic Review. \emph{J. Med. Internet Res.} \textbf{\boldmath{2022}}, \emph{24}, e33380. [\href{https://doi.org/10.2196/33380}{CrossRef}] [\href{https://www.ncbi.nlm.nih.gov/pubmed/35576567}{PubMed}]

\bibitem{B21-robotics-3060247}
Abdullahi, U.; Muhammad, B.; Masari, A.; Bugaje, A. A Remote-Operated Humanoid Robot Based Patient Monitoring System. \emph{IRE J.} \textbf{\boldmath{2023}}, \emph{7}, 17--22.

\bibitem{B22-robotics-3060247}
Gonzalez, C. Service Robots Used for Medical Care and Deliveries---ASME. Available online:  \url{https://www.asme.org/topics-resources/content/are-service-bots-the-new-future-post-covid-19} (accessed on 3 January 2024).

\bibitem{B23-robotics-3060247}
Sarker, S.; Jamal, L.; Ahmed, S.F.; Irtisam, N. Robotics and Artificial Intelligence in Healthcare during COVID-19 Pandemic: A Systematic Review. \emph{Robot. Auton. Syst.} \textbf{\boldmath{2021}}, \emph{146}, 103902. [\href{https://doi.org/10.1016/j.robot.2021.103902}{CrossRef}] [\href{https://www.ncbi.nlm.nih.gov/pubmed/34629751}{PubMed}]

\bibitem{B24-robotics-3060247}
How Robots Became Essential Workers in the COVID-19 Response---IEEE Spectrum. Available online:  \url{https://spectrum.ieee.org/how-robots-became-essential-workers-in-the-covid19-response} (accessed on 20 January 2024).

\bibitem{B25-robotics-3060247}
The Clever Use of Robots during COVID-19---EHL Insights\textbar{}Business. Available online:  \url{https://hospitalityinsights.ehl.edu/robots-during-covid-19} (accessed on 20 January 2024).

\bibitem{B26-robotics-3060247}
Getson, C.; Nejat, G. The Adoption of Socially Assistive Robots for Long-Term Care: During COVID-19 and in a Post-Pandemic Society. \emph{Healthc. Manag. Forum} \textbf{\boldmath{2022}}, \emph{35}, 301--309. [\href{https://doi.org/10.1177/08404704221106406}{CrossRef}] [\href{https://www.ncbi.nlm.nih.gov/pubmed/35714374}{PubMed}]

\bibitem{B27-robotics-3060247}
Henschel, A.; Laban, G.; Cross, E.S. What Makes a Robot Social? A Review of Social Robots from Science Fiction to a Home or Hospital Near You. \emph{Curr. Robot. Rep.} \textbf{\boldmath{2021}}, \emph{2}, 9--19. [\href{https://doi.org/10.1007/s43154-020-00035-0}{CrossRef}] [\href{https://www.ncbi.nlm.nih.gov/pubmed/34977592}{PubMed}]

\bibitem{B28-robotics-3060247}
Kim, J.; Kim, S.; Kim, S.; Lee, E.; Heo, Y.; Hwang, C.-Y.; Choi, Y.-Y.; Kong, H.-J.; Ryu, H.; Lee, H. Companion Robots for Older Adults: Rodgers’ Evolutionary Concept Analysis Approach. \emph{Intell. Serv. Robot.} \textbf{\boldmath{2021}}, \emph{14}, 729--739. [\href{https://doi.org/10.1007/s11370-021-00394-3}{CrossRef}] [\href{https://www.ncbi.nlm.nih.gov/pubmed/34804242}{PubMed}]

\bibitem{B29-robotics-3060247}
Denecke, K.; Baudoin, C.R. A Review of Artificial Intelligence and Robotics in Transformed Health Ecosystems. \emph{Front. Med.} \textbf{\boldmath{2022}}, \emph{9}, 795957. [\href{https://doi.org/10.3389/fmed.2022.795957}{CrossRef}]

\bibitem{B30-robotics-3060247}
Sevilla-Salcedo, J.; Fern{\fontencoding{T5}\selectfont{\'a}}dez-Rodicio, E.; Mart{\fontencoding{T5}\selectfont{\'i}}n-Galv{\fontencoding{T5}\selectfont{\'a}}n, L.; Castro-Gonz{\fontencoding{T5}\selectfont{\'a}}lez, {\fontencoding{T5}\selectfont{\'A}}.; Castillo, J.C.; Salichs, M.A. Using Large Language Models to Shape Social Robots’ Speech. \emph{Int. J. Interact. Multimed. Artif. Intell.} \textbf{\boldmath{2023}}, \emph{8}, 6. [\href{https://doi.org/10.9781/ijimai.2023.07.008}{CrossRef}]

\bibitem{B31-robotics-3060247}
Addlesee, A.; Sieińska, W.; Gunson, N.; Garcia, D.H.; Dondrup, C.; Lemon, O. Multi-Party Goal Tracking with LLMs: Comparing Pre-Training, Fine-Tuning, and Prompt Engineering 2023. In Proceedings of the 24th Annual Meeting of the Special Interest Group on Discourse and Dialogue, Prague, Czechia, 11--15 September 2023.

\bibitem{B32-robotics-3060247}
Pandya, A. ChatGPT-Enabled daVinci Surgical Robot Prototype: Advancements and Limitations. \emph{Robotics} \textbf{\boldmath{2023}}, \emph{12}, 97. [\href{https://doi.org/10.3390/robotics12040097}{CrossRef}]

\bibitem{B33-robotics-3060247}
Elgedawy, R.; Srinivasan, S.; Danciu, I. Dynamic Q\&A of Clinical Documents with Large Language Models. \emph{arXiv} \textbf{\boldmath{2024}}, arXiv:2401.10733.

\bibitem{B34-robotics-3060247}
Hu, M.; Pan, S.; Li, Y.; Yang, X. Advancing Medical Imaging with Language Models: A journey from n-grams to chatgpt. \emph{arXiv} \textbf{\boldmath{2023}}, arXiv:2304.04920.

\bibitem{B35-robotics-3060247}
A Comprehensive Overview of Large Language Models. Available online:  \url{https://ar5iv.labs.arxiv.org/html/2307.06435} (accessed on 8 March 2024).

\bibitem{B36-robotics-3060247}
Lin, T.; Wang, Y.; Liu, X.; Qiu, X. A Survey of Transformers. \emph{AI Open} \textbf{\boldmath{2022}}, \emph{3}, 111--132. [\href{https://doi.org/10.1016/j.aiopen.2022.10.001}{CrossRef}]

\bibitem{B37-robotics-3060247}
Decoder-Only or Encoder-Decoder? \emph{Interpreting Language Model as a Regularized Encoder-Decoder.} Available online:  \url{https://ar5iv.labs.arxiv.org/html/2304.04052} (accessed on 8 March 2024).

\bibitem{B38-robotics-3060247}
King, J.; Baffour, P.; Crossley, S.; Holbrook, R.; Demkin, M. LLM---Detect AI Generated Text. Available online:  \url{https://kaggle.com/competitions/llm-detect-ai-generated-text} (accessed on 8 March 2024).

\bibitem{B39-robotics-3060247}
Vaswani, A.; Shazeer, N.; Parmar, N.; Uszkoreit, J.; Jones, L.; Gomez, A.N.; Kaiser, Ł.; Polosukhin, I. Attention is all you need. In Proceedings of the 31st International Conference on Neural Information Processing Systems, Long Beach, CA, USA, 4--9 December 2017; 2017; pp. 6000--6010.

\bibitem{B40-robotics-3060247}
Burns, K.; Jain, A.; Go, K.; Xia, F.; Stark, M.; Schaal, S.; Hausman, K. Generating Robot Policy Code for High-Precision and Contact-Rich Manipulation Tasks. \emph{arXiv} \textbf{\boldmath{2023}}, arXiv:2404.06645.

\bibitem{B41-robotics-3060247}
Gu, Y.; Tinn, R.; Cheng, H.; Lucas, M.; Usuyama, N.; Liu, X.; Naumann, T.; Gao, J.; Poon, H. Domain-Specific Language Model Pretraining for Biomedical Natural Language Processing. Available online:  \url{https://arxiv.org/abs/2007.15779v6} (accessed on 8~March~2024).

\bibitem{B42-robotics-3060247}
Gupta, A.; Waldron, A. Sharing Google’s Med-PaLM 2 Medical Large Language Model, or LLM. Available online:  \url{https://cloud.google.com/blog/topics/healthcare-life-sciences/sharing-google-med-palm-2-medical-large-language-model} (accessed on\linebreak 20 January 2024).

\bibitem{B43-robotics-3060247}
Wang, H.; Gao, C.; Dantona, C.; Hull, B.; Sun, J. DRG-LLaMA: Tuning LLaMA Model to Predict Diagnosis-Related Group for Hospitalized Patients. \emph{Npj Digit. Med.} \textbf{\boldmath{2024}}, \emph{7}, 1--9. [\href{https://doi.org/10.1038/s41746-023-00989-3}{CrossRef}] [\href{https://www.ncbi.nlm.nih.gov/pubmed/38253711}{PubMed}]

\bibitem{B44-robotics-3060247}
Schneider, E.T.R.; de Souza, J.V.A.; Gumiel, Y.B.; Moro, C.; Paraiso, E.C. A GPT-2 Language Model for Biomedical Texts in Portuguese. In Proceedings of the 2021 IEEE 34th International Symposium on Computer-Based Medical Systems (CBMS), Aveiro, Portugal, 7--9 June 2021; pp. 474--479.

\bibitem{B45-robotics-3060247}
Lehman, E.; Johnson, A. Clinical-T5: Large Language Models Built Using MIMIC Clinical Text. \emph{PhysioNet} \textbf{\boldmath{2023}}. [\href{https://doi.org/10.13026/rj8x-v335}{CrossRef}]

\bibitem{B46-robotics-3060247}
Brown, T.B.; Mann, B.; Ryder, N.; Subbiah, M.; Kaplan, J.; Dhariwal, P.; Neelakantan, A.; Shyam, P.; Sastry, G.; Askell, A.; et~al. Language Models Are Few-Shot Learners. In Proceedings of the 34th International Conference on Neural Information Processing Systems, Vancouver, BC, Canada, 6--12 December 2020; p. 159.

\bibitem{B47-robotics-3060247}
OpenAI; Achiam, J.; Adler, S.; Agarwal, S.; Ahmad, L.; Akkaya, I.; Aleman, F.L.; Almeida, D.; Altenschmidt, J.; Altman, S.; et~al. GPT-4 Technical Report. \emph{arXiv} \textbf{\boldmath{2023}}, arXiv:2303.08774.

\bibitem{B48-robotics-3060247}
Raffel, C.; Shazeer, N.; Roberts, A.; Lee, K.; Narang, S.; Matena, M.; Zhou, Y.; Li, W.; Liu, P.J. Exploring the Limits of Transfer Learning with a Unified Text-to-Text Transformer. \emph{J. Mach. Learn. Res.} \textbf{\boldmath{2020}}, \emph{21}, 1--67.

\bibitem{B49-robotics-3060247}
Beltagy, I.; Peters, M.E.; Cohan, A. Longformer: The Long-Document Transformer. \emph{arXiv} \textbf{\boldmath{2020}}, arXiv:2004.05150.

\bibitem{B50-robotics-3060247}
Chowdhery, A.; Narang, S.; Devlin, J.; Bosma, M.; Mishra, G.; Roberts, A.; Barham, P.; Chung, H.W.; Sutton, C.; Gehrmann, S.; et~al. PaLM: Scaling Language Modeling with Pathways. \emph{J. Mach. Learn. Res.} \textbf{\boldmath{2023}}, \emph{24}, 1--113.

\bibitem{B51-robotics-3060247}
Taori, R.; Gulrajani, I.; Zhang, T.; Dubois, Y.; Li, X.; Guestrin, C.; Liang, P.; Hashimoto, T. \emph{Alpaca: A Strong, Replicable Instruction-Following Mode}; Stanford Center for Research on Foundation Models: Stanford, CA, USA, 2023. Available online:  \url{https://crfm.stanford.edu/2023/03/13/alpaca.html} (accessed on 8 March 2024).

\bibitem{B52-robotics-3060247}
Devlin, J.; Chang, M.-W.; Lee, K.; Toutanova, K. BERT: Pre-Training of Deep Bidirectional Transformers for Language Understanding. In Proceedings of the 2019 Conference of the North American Chapter of the Association for Computational Linguistics: Human Language Technologies, Minneapolis, MN, USA, 2--7 June 2019; pp. 4171--4186.

\bibitem{B53-robotics-3060247}
Anil, R.; Dai, A.M.; Firat, O.; Johnson, M.; Lepikhin, D.; Passos, A.; Shakeri, S.; Taropa, E.; Bailey, P.; Chen, Z.; et~al. PaLM 2 Technical Report. \emph{arXiv} \textbf{\boldmath{2023}}, arXiv:2305.10403.

\bibitem{B54-robotics-3060247}
Touvron, H.; Martin, L.; Stone, K.R.; Albert, P.; Almahairi, A.; Babaei, Y.; Bashlykov, N.; Batra, S.; Bhargava, P.; Bhosale, S.; et~al. Llama 2: Open Foundation and Fine-Tuned Chat Models. \emph{arXiv} \textbf{\boldmath{2020}}, arXiv:2307.09288.

\bibitem{B55-robotics-3060247}
Radford, A.; Wu, J.; Child, R.; Luan, D.; Amodei, D.; Sutskever, I. Language Models Are Unsupervised Multitask Learners. Available online:  \url{https://openai.com/index/better-language-models/} (accessed on 26 June 2024).

\bibitem{B56-robotics-3060247}
Jin, D.; Pan, E.; Oufattole, N.; Weng, W.-H.; Fang, H.; Szolovits, P. What Disease Does This Patient Have? A Large-Scale Open Domain Question Answering Dataset from Medical Exams. \emph{Appl. Sci.} \textbf{\boldmath{2020}}, \emph{11}, 6421. [\href{https://doi.org/10.3390/app11146421}{CrossRef}]

\bibitem{B57-robotics-3060247}
Pal, A.; Umapathi, L.K.; Sankarasubbu, M. MedMCQA: A Large-Scale Multi-Subject Multi-Choice Dataset for Medical Domain Question Answering. In Proceedings of the Machine Learning Research (PMLR), ACM Conference on Health, Inference, and Learning (CHIL), Virtual, 7 April 2022; Volume 174, pp. 248--260.

\bibitem{B58-robotics-3060247}
Jin, Q.; Dhingra, B.; Liu, Z.; Cohen, W.W.; Lu, X. PubMedQA: A Dataset for Biomedical Research Question Answering. In Proceedings of the 2019 Conference on Empirical Methods in Natural Language Processing and the 9th International Joint Conference on Natural Language Processing, Hong Kong, China, 3--7 November 2019; pp. 2567--2577.

\bibitem{B59-robotics-3060247}
Tozzi, C.; Zittrain, J. Introduction. In \emph{For Fun and Profit: A History of the Free and Open Source Software Revolution}; MIT Press: Cambridge, MA, USA, 2017. Available online:  \url{https://ieeexplore.ieee.org/document/8047084} (accessed on 29 May 2024).

\bibitem{B60-robotics-3060247}
Spirling, A. Why Open-Source Generative AI Models Are an Ethical Way Forward for Science. \emph{Nature} \textbf{\boldmath{2023}}, \emph{616}, 413. [\href{https://doi.org/10.1038/d41586-023-01295-4}{CrossRef}] [\href{https://www.ncbi.nlm.nih.gov/pubmed/37072520}{PubMed}]

\bibitem{B61-robotics-3060247}
Bommasani, R.; Hudson, D.A.; Adeli, E.; Altman, R.; Arora, S.; von Arx, S.; Bernstein, M.S.; Bohg, J.; Bosselut, A.; Brunskill, E.; et~al. On the Opportunities and Risks of Foundation Models. \emph{arXiv} \textbf{\boldmath{2022}}, arXiv:2108.07258.

\bibitem{B62-robotics-3060247}
Self-Influence Guided Data Reweighting for Language Model Pre-Training. Available online:  \url{https://ar5iv.labs.arxiv.org/html/2311.00913} (accessed on 8 March 2024).

\bibitem{B63-robotics-3060247}
Solaiman, I.; Dennison, C. Process for Adapting Language Models to Society (PALMS) with Values-Targeted Datasets. In Proceedings of the 35th International Conference on Neural Information Processing Systems, Sydney, Australia, 6--14 December 2021; p. 448.

\bibitem{B64-robotics-3060247}
Prompt Design and Engineering: Introduction and Advanced Methods. Available online:  \url{https://ar5iv.labs.arxiv.org/html/2401.14423} (accessed on 8 March 2024).

\bibitem{B65-robotics-3060247}
Ratner, N.; Levine, Y.; Belinkov, Y.; Ram, O.; Magar, I.; Abend, O.; Karpas, E.; Shashua, A.; Leyton-Brown, K.; Shoham, Y. Parallel Context Windows for Large Language Models. In Proceedings of the 61st Annual Meeting of the Association for Computational Linguistics, Toronto, ON, Canada, 9--14 July 2023; pp. 6383--6402.

\bibitem{B66-robotics-3060247}
Chen, H.; Pasunuru, R.; Weston, J.; Celikyilmaz, A. Walking Down the Memory Maze: Beyond Context Limit through Interactive Reading. \emph{arXiv} \textbf{\boldmath{2023}}, arXiv:2310.05029.

\bibitem{B67-robotics-3060247}
Fairness-Guided Few-Shot Prompting for Large Language Models. Available online:  \url{https://ar5iv.labs.arxiv.org/html/2303.13217} (accessed on 8 March 2024).

\bibitem{B68-robotics-3060247}
Skill-Based Few-Shot Selection for In-Context Learning. Available online:  \url{https://ar5iv.labs.arxiv.org/html/2305.14210} (accessed on 8 March 2024).

\bibitem{B69-robotics-3060247}
Extending Context Window of Large Language Models via Position Interpolation. Available online:  \url{https://ar5iv.labs.arxiv.org/html/2306.15595} (accessed on 8 March 2024).

\bibitem{B70-robotics-3060247}
Parallel Context Windows Improve In-Context Learning of Large Language Models. Available online:  \url{https://ar5iv.labs.arxiv.org/html/2212.10947} (accessed on 8 March 2024).

\bibitem{B71-robotics-3060247}
MM-LLMs: Recent Advances in MultiModal Large Language Models. Available online:  \url{https://ar5iv.labs.arxiv.org/html/2401.13601} (accessed on 8 March 2024).

\bibitem{B72-robotics-3060247}
Radford, A.; Kim, J.W.; Hallacy, C.; Ramesh, A.; Goh, G.; Agarwal, S.; Sastry, G.; Askell, A.; Mishkin, P.; Clark, J.; et~al. Learning Transferable Visual Models From Natural Language Supervision. In Proceedings of the 38th International Conference on Machine Learning, ICML 2021, Virtual Event, 18--24 July 2021; Volume 139, pp. 8748--8763.

\bibitem{B73-robotics-3060247}
Chen, D.; Chang, A.; Nie{\ss}ner, M. ScanRefer: 3D Object Localization in RGB-DScans Using Natural Language. In Proceedings of the Computer Vision---ECCV 2020, Glasgow, UK, 12 November 2020; pp. 202--221.

\bibitem{B74-robotics-3060247}
Liu, J.X.; Yang, Z.; Idrees, I.; Liang, S.; Schornstein, B.; Tellex, S.; Shah, A. Grounding Complex Natural Language Commands for Temporal Tasks in Unseen Environments. In Proceedings of the 7th Conference on Robot Learning, Atlanta, GA, USA, 6--9 November 2023.

\bibitem{B75-robotics-3060247}
Liu, M.; Shen, Y.; Yao, B.M.; Wang, S.; Qi, J.; Xu, Z.; Huang, L. KnowledgeBot: Improving Assistive Robot for Task Completion and Live Interaction via Neuro-Symbolic Reasoning. In Proceedings of the Alexa Prize SimBot Challenge, Virtual Event, \mbox{6 April 2023}.

\bibitem{B76-robotics-3060247}
Salichs, M.A.; Castro-Gonz{\fontencoding{T5}\selectfont{\'a}}lez, {\fontencoding{T5}\selectfont{\'A}}.; Salichs, E.; Fern{\fontencoding{T5}\selectfont{\'a}}ndez-Rodicio, E.; Maroto-G{\fontencoding{T5}\selectfont{\'o}}mez, M.; Gamboa-Montero, J.J.; \mbox{Marques-Villarroya, S.;} Castillo, J.C.; Alonso-Mart{\fontencoding{T5}\selectfont{\'i}}n, F.; Malfaz, M. Mini: A New Social Robot for the Elderly. \emph{Int. J. Soc. Robot.} \textbf{\boldmath{2020}}, \emph{12}, 1231--1249. [\href{https://doi.org/10.1007/s12369-020-00687-0}{CrossRef}]

\bibitem{B77-robotics-3060247}
Zhao, X.; Li, M.; Weber, C.; Hafez, M.B.; Wermter, S. Chat with the Environment: Interactive Multimodal Perception Using Large Language Models. In Proceedings of the 2023 IEEE/RSJ International Conference on Intelligent Robots and Systems (IROS), Detroit, MI, USA, 1 October 2023; pp. 3590--3596.

\bibitem{B78-robotics-3060247}
Singh, I.; Blukis, V.; Mousavian, A.; Goyal, A.; Xu, D.; Tremblay, J.; Fox, D.; Thomason, J.; Garg, A. ProgPrompt: Program Generation for Situated Robot Task Planning Using Large Language Models. \emph{Auton. Robots} \textbf{\boldmath{2023}}, \emph{47}, 999--1012. [\href{https://doi.org/10.1007/s10514-023-10135-3}{CrossRef}]

\bibitem{B79-robotics-3060247}
Jin, Y.; Li, D.; A, Y.; Shi, J.; Hao, P.; Sun, F.; Zhang, J.; Fang, B. RobotGPT: Robot Manipulation Learning from ChatGPT. \emph{IEEE Robot. Autom. Lett.} \textbf{\boldmath{2023}}, \emph{9}, 2543--2550. [\href{https://doi.org/10.1109/LRA.2024.3357432}{CrossRef}]

\bibitem{B80-robotics-3060247}
Obinata, Y.; Kanazawa, N.; Kawaharazuka, K.; Yanokura, I.; Kim, S.; Okada, K.; Inaba, M. Foundation Model Based Open Vocabulary Task Planning and Executive System for General Purpose Service Robots. \emph{arXiv} \textbf{\boldmath{2023}}, arXiv:2308.03357.

\bibitem{B81-robotics-3060247}
Murali, P.; Steenstra, I.; Yun, H.S.; Shamekhi, A.; Bickmore, T. Improving Multiparty Interactions with a Robot Using Large Language Models. In Proceedings of the Extended Abstracts of the 2023 CHI Conference on Human Factors in Computing Systems, Hamburg, Germany, 19 April 2023; pp. 1--8.

\bibitem{B82-robotics-3060247}
Paiva, A.; Leite, I.; Boukricha, H.; Wachsmuth, I. Empathy in Virtual Agents and Robots: A Survey. \emph{ACM Trans. Interact. Intell. Syst.} \textbf{\boldmath{2017}}, \emph{7}, 11. [\href{https://doi.org/10.1145/2912150}{CrossRef}]

\bibitem{B83-robotics-3060247}
Cherakara, N.; Varghese, F.; Shabana, S.; Nelson, N.; Karukayil, A.; Kulothungan, R.; Farhan, M.A.; Nesset, B.; Moujahid, M.; Dinkar, T.; et~al. FurChat: An Embodied Conversational Agent Using LLMs, Combining Open and Closed-Domain Dialogue with Facial Expressions. In Proceedings of the 24th Annual Meeting of the Special Interest Group on Discourse and Dialogue, Prague, Czechia, 11--15 September 2023; pp. 588--592.

\bibitem{B84-robotics-3060247}
Zhang, B.; Soh, H. Large Language Models as Zero-Shot Human Models for Human-Robot Interaction. \emph{arXiv} \textbf{\boldmath{2023}}, arXiv:2303.03548.

\bibitem{B85-robotics-3060247}
Yang, J.; Chen, X.; Qian, S.; Madaan, N.; Iyengar, M.; Fouhey, D.F.; Chai, J. LLM-Grounder: Open-Vocabulary 3D Visual Grounding with Large Language Model as an Agent. \emph{arXiv} \textbf{\boldmath{2023}}, arXiv:2309.12311.

\bibitem{B86-robotics-3060247}
Yoshida, T.; Masumori, A.; Ikegami, T. From Text to Motion: Grounding GPT-4 in a Humanoid Robot “Alter3”. \emph{arXiv} \textbf{\boldmath{2023}}, arXiv:2312.06571.

\bibitem{B87-robotics-3060247}
Ahn, M.; Brohan, A.; Brown, N.; Chebotar, Y.; Cortes, O.; David, B.; Finn, C.; Fu, C.; Gopalakrishnan, K.; Hausman, K.; et~al. Do As I Can, Not As I Say: Grounding Language in Robotic Affordances. \emph{arXiv} \textbf{\boldmath{2022}}, arXiv:2204.01691.

\bibitem{B88-robotics-3060247}
Lykov, A.; Tsetserukou, D. LLM-BRAIn: AI-Driven Fast Generation of Robot Behaviour Tree Based on Large Language Model. \emph{arXiv} \textbf{\boldmath{2023}}, arXiv:2305.19352.

\bibitem{B89-robotics-3060247}
Kubota, A.; Cruz-Sandoval, D.; Kim, S.; Twamley, E.W.; Riek, L.D. Cognitively Assistive Robots at Home: HRI Design Patterns for Translational Science. In Proceedings of the 2022 17th ACM/IEEE International Conference on Human-Robot Interaction (HRI), Sapporo, Hokkaido, Japan, 7--10 March 2022; pp. 53--62.

\bibitem{B90-robotics-3060247}
Elbeleidy, S.; Rosen, D.; Liu, D.; Shick, A.; Williams, T. Analyzing Teleoperation Interface Usage of Robots in Therapy for Children with Autism. In Proceedings of the ACM Interaction Design and Children Conference, Athens, Greece, 18 May 2021; pp. 112--118.

\bibitem{B91-robotics-3060247}
Louie, W.-Y.G.; Nejat, G. A Social Robot Learning to Facilitate an Assistive Group-Based Activity from Non-Expert Caregivers. \emph{Int. J. Soc. Robot.} \textbf{\boldmath{2020}}, \emph{12}, 1159--1176. [\href{https://doi.org/10.1007/s12369-020-00621-4}{CrossRef}]

\bibitem{B92-robotics-3060247}
Mishra, D.; Romero, G.A.; Pande, A.; Nachenahalli Bhuthegowda, B.; Chaskopoulos, D.; Shrestha, B. An Exploration of the Pepper Robot’s Capabilities: Unveiling Its Potential. \emph{Appl. Sci.} \textbf{\boldmath{2024}}, \emph{14}, 110. [\href{https://doi.org/10.3390/app14010110}{CrossRef}]

\bibitem{B93-robotics-3060247}
Anderson, P.L.; Lathrop, R.A.; Herrell, S.D.; Webster, R.J. Comparing a Mechanical Analogue With the Da Vinci User Interface: Suturing at Challenging Angles. \emph{IEEE Robot. Autom. Lett.} \textbf{\boldmath{2016}}, \emph{1}, 1060--1065. [\href{https://doi.org/10.1109/LRA.2016.2528302}{CrossRef}] [\href{https://www.ncbi.nlm.nih.gov/pubmed/30090854}{PubMed}]

\bibitem{B94-robotics-3060247}
Muradore, R.; Fiorini, P.; Akgun, G.; Barkana, D.E.; Bonfe, M.; Boriero, F.; Caprara, A.; De Rossi, G.; Dodi, R.; Elle, O.J.; et~al. Development of a Cognitive Robotic System for Simple Surgical Tasks. \emph{Int. J. Adv. Robot. Syst.} \textbf{\boldmath{2015}}, \emph{12}, 37. [\href{https://doi.org/10.5772/60137}{CrossRef}]

\bibitem{B95-robotics-3060247}
Łukasik, S.; Tobis, S.; Suwalska, J.; Łojko, D.; Napierała, M.; Proch, M.; Neumann-Podczaska, A.; Suwalska, A. The Role of Socially Assistive Robots in the Care of Older People: To Assist in Cognitive Training, to Remind or to Accompany? \emph{Sustainability} \textbf{\boldmath{2021}}, \emph{13}, 10394. [\href{https://doi.org/10.3390/su131810394}{CrossRef}]

\bibitem{B96-robotics-3060247}
Natural Language Robot Programming: NLP Integrated with Autonomous Robotic Grasping. Available online:  \url{https://ar5iv.labs.arxiv.org/html/2304.02993} (accessed on 8 March 2024).

\bibitem{B97-robotics-3060247}
Papadopoulos, I.; Koulouglioti, C.; Lazzarino, R.; Ali, S. Enablers and Barriers to the Implementation of Socially Assistive Humanoid Robots in Health and Social Care: A Systematic Review. \emph{BMJ Open} \textbf{\boldmath{2020}}, \emph{10}, e033096. [\href{https://doi.org/10.1136/bmjopen-2019-033096}{CrossRef}]

\bibitem{B98-robotics-3060247}
Kim, C.Y.; Lee, C.P.; Mutlu, B. Understanding Large-Language Model (LLM)-Powered Human-Robot Interaction. In Proceedings of the 2024 ACM/IEEE International Conference on Human-Robot Interaction (HRI $'$24), Boulder, CO, USA, 11--14 March 2024; pp. 371--380.

\bibitem{B99-robotics-3060247}
Emotion Is All You Need?-Boosting ChatGPT Performance with Emotional Stimulus-FlowGPT. Available online:  \url{https://flowgpt.com/blog/emoGPT} (accessed on 30 January 2024).

\bibitem{B100-robotics-3060247}
Mishra, C.; Verdonschot, R.; Hagoort, P.; Skantze, G. Real-Time Emotion Generation in Human-Robot Dialogue Using Large Language Models. \emph{Front. Robot. AI} \textbf{\boldmath{2023}}, \emph{10}, 1271610. [\href{https://doi.org/10.3389/frobt.2023.1271610}{CrossRef}] [\href{https://www.ncbi.nlm.nih.gov/pubmed/38106543}{PubMed}]

\bibitem{B101-robotics-3060247}
Wang, C.; Hasler, S.; Tanneberg, D.; Ocker, F.; Joublin, F.; Ceravola, A.; Deigmoeller, J.; Gienger, M. LaMI: Large Language Models for Multi-Modal Human-Robot Interaction. In Proceedings of the Extended Abstracts of the 2024 CHI Conference on Human Factors in Computing Systems, Honolulu, HI, USA, 11--16 May 2024; p. 218.

\bibitem{B102-robotics-3060247}
Townsend, D.; MajidiRad, A. Trust in Human-Robot Interaction Within Healthcare Services: A Review Study. In Proceedings of the Volume 7: 46th Mechanisms and Robotics Conference (MR), St. Louis, MI, USA, 14 August 2022; p. V007T07A030.

\bibitem{B103-robotics-3060247}
Abdi, J.; Al-Hindawi, A.; Ng, T.; Vizcaychipi, M.P. Scoping Review on the Use of Socially Assistive Robot Technology in Elderly Care. \emph{BMJ Open} \textbf{\boldmath{2018}}, \emph{8}, e018815. [\href{https://doi.org/10.1136/bmjopen-2017-018815}{CrossRef}] [\href{https://www.ncbi.nlm.nih.gov/pubmed/29440212}{PubMed}]

\bibitem{B104-robotics-3060247}
Abbott, R.; Orr, N.; McGill, P.; Whear, R.; Bethel, A.; Garside, R.; Stein, K.; Thompson-Coon, J. How Do “Robopets” Impact the Health and Well-being of Residents in Care Homes? A Systematic Review of Qualitative and Quantitative Evidence. \emph{Int. J. Older People Nurs.} \textbf{\boldmath{2019}}, \emph{14}, e12239. [\href{https://doi.org/10.1111/opn.12239}{CrossRef}] [\href{https://www.ncbi.nlm.nih.gov/pubmed/31070870}{PubMed}]

\bibitem{B105-robotics-3060247}
Xue, L.; Constant, N.; Roberts, A.; Kale, M.; Al-Rfou, R.; Siddhant, A.; Barua, A.; Raffel, C. mT5: A Massively Multilingual Pre-Trained Text-to-Text Transformer. In Proceedings of the 2021 Conference of the North American Chapter of the Association for Computational Linguistics: Human Language Technologies, Online, 6--11 June 2021; pp. 483--498.

\bibitem{B106-robotics-3060247}
Zhang, J.; Zhao, Y.; Saleh, M.; Liu, P.J. PEGASUS: Pre-Training with Extracted Gap-Sentences for Abstractive Summarization. \emph{arXiv} \textbf{\boldmath{2020}}. [\href{https://doi.org/10.48550/arXiv.1912.08777}{CrossRef}]

\bibitem{B107-robotics-3060247}
Cañete, J.; Chaperon, G.; Fuentes, R.; Ho, J.-H.; Kang, H.; P{\fontencoding{T5}\selectfont{\'e}}rez, J. Spanish Pre-Trained BERT Model and Evaluation Data. \emph{arXiv} \textbf{\boldmath{2023}}, arXiv:2308.02976.

\bibitem{B108-robotics-3060247}
Da Vinci Robotic Surgical Systems\textbar{}Intuitive. Available online:  \url{https://www.intuitive.com/en-us/products-and-services/da-vinci} (accessed on 9 March 2024).

\bibitem{B109-robotics-3060247}
ROS: Home. Available online:  \url{https://www.ros.org/} (accessed on 21 September 2023).

\bibitem{B110-robotics-3060247}
Palrobot ARI---The Social and Collaborative Robot. Available online:  \url{https://pal-robotics.com/robots/ari/} (accessed on \mbox{9 March 2024}).

\bibitem{B111-robotics-3060247}
The World’s Most Advanced Social Robot. Available online:  \url{https://furhatrobotics.com/} (accessed on 9 March 2024).

\bibitem{B112-robotics-3060247}
Radford, A.; Kim, J.W.; Xu, T.; Brockman, G.; McLeavey, C.; Sutskever, I. Robust speech recognition via large-scale weak supervision. In Proceedings of the 40th International Conference on Machine Learning, Honolulu, HI, USA, 23--29 July 2023; p.~1182.

\bibitem{B113-robotics-3060247}
Su, H.; Qi, W.; Chen, J.; Yang, C.; Sandoval, J.; Laribi, M.A. Recent Advancements in Multimodal Human--Robot Interaction. \emph{Front. Neurorobot.} \textbf{\boldmath{2023}}, \emph{17}, 1084000. [\href{https://doi.org/10.3389/fnbot.2023.1084000}{CrossRef}] [\href{https://www.ncbi.nlm.nih.gov/pubmed/37250671}{PubMed}]

\bibitem{B114-robotics-3060247}
Saunderson, S.; Nejat, G. How Robots Influence Humans: A Survey of Nonverbal Communication in Social Human--Robot Interaction. \emph{Int. J. Soc. Robot.} \textbf{\boldmath{2019}}, \emph{11}, 575--608. [\href{https://doi.org/10.1007/s12369-019-00523-0}{CrossRef}]

\bibitem{B115-robotics-3060247}
Maurtua, I.; Fern{\fontencoding{T5}\selectfont{\'a}}ndez, I.; Tellaeche, A.; Kildal, J.; Susperregi, L.; Ibarguren, A.; Sierra, B. Natural Multimodal Communication for Human--Robot Collaboration. \emph{Int. J. Adv. Robot. Syst.} \textbf{\boldmath{2017}}, \emph{14}, 1729881417716043. [\href{https://doi.org/10.1177/1729881417716043}{CrossRef}]

\bibitem{B116-robotics-3060247}
Schreiter, T.; Morillo-Mendez, L.; Chadalavada, R.T.; Rudenko, A.; Billing, E.; Magnusson, M.; Arras, K.O.; Lilienthal, A.J. Advantages of Multimodal versus Verbal-Only Robot-to-Human Communication with an Anthropomorphic Robotic Mock Driver. In Proceedings of the 2023 32nd IEEE International Conference on Robot and Human Interactive Communication (RO-MAN), Busan, Republic of Korea, 28--31 August 2023; pp. 293--300.

\bibitem{B117-robotics-3060247}
RITA--Reminiscence Interactive Therapy and Activities---mPower. Available online:  \url{https://mpowerhealth.eu/impact/reducing-the-digital-divide-connecting-and-empowering/rita-reminiscence-interactive-therapy-and-activities/} (accessed on \mbox{27 May 2024}).

\bibitem{B118-robotics-3060247}
\emph{Roles and Challenges of Semantic Intelligence in Healthcare Cognitive Computing}; Carbonaro, A.; Tiwari, S.; Ortiz-Rodriguez, F.; Janev, V. (Eds.) Studies on the Semantic Web/Ssw; IOS Press: Amsterdam, The Netherlands, 2023; ISBN 978-1-64368-460-4.

\bibitem{B119-robotics-3060247}
National Library of Medicine The Semantic Network. Available online:  \url{https://www.nlm.nih.gov/research/umls/new_users/online_learning/OVR_003.html} (accessed on 29 January 2024).

\bibitem{B120-robotics-3060247}
Zhang, H.; Hu, H.; Diller, M.; Hogan, W.R.; Prosperi, M.; Guo, Y.; Bian, J. Semantic Standards of External Exposome Data. \emph{Environ. Res.} \textbf{\boldmath{2021}}, \emph{197}, 111185. [\href{https://doi.org/10.1016/j.envres.2021.111185}{CrossRef}] [\href{https://www.ncbi.nlm.nih.gov/pubmed/33901445}{PubMed}]

\bibitem{B121-robotics-3060247}
Aldughayfiq, B.; Ashfaq, F.; Jhanjhi, N.Z.; Humayun, M. Capturing Semantic Relationships in Electronic Health Records Using Knowledge Graphs: An Implementation Using MIMIC III Dataset and GraphDB. \emph{Healthcare} \textbf{\boldmath{2023}}, \emph{11}, 1762. [\href{https://doi.org/10.3390/healthcare11121762}{CrossRef}] [\href{https://www.ncbi.nlm.nih.gov/pubmed/37372880}{PubMed}]

\bibitem{B122-robotics-3060247}
Busso, M.; Gonzalez, M.P.; Scartascini, C. On the Demand for Telemedicine: Evidence from the COVID-19 Pandemic. \emph{Health Econ.} \textbf{\boldmath{2022}}, \emph{31}, 1491--1505. [\href{https://doi.org/10.1002/hec.4523}{CrossRef}] [\href{https://www.ncbi.nlm.nih.gov/pubmed/35527351}{PubMed}]

\bibitem{B123-robotics-3060247}
Chatterjee, A.; Prinz, A.; Riegler, M.A.; Meena, Y.K. An Automatic and Personalized Recommendation Modelling in Activity eCoaching with Deep Learning and Ontology. \emph{Sci. Rep.} \textbf{\boldmath{2023}}, \emph{13}, 10182. [\href{https://doi.org/10.1038/s41598-023-37233-7}{CrossRef}] [\href{https://www.ncbi.nlm.nih.gov/pubmed/37349483}{PubMed}]

\bibitem{B124-robotics-3060247}
Barisevičius, G.; Coste, M.; Geleta, D.; Juric, D.; Khodadadi, M.; Stoilos, G.; Zaihrayeu, I. Supporting Digital Healthcare Services Using Semantic Web Technologies. In Proceedings of the 17th International Semantic Web Conference, Monterey, CA, USA, 8--12 October 2018; pp. 291--306.

\bibitem{B125-robotics-3060247}
Yu, W.D.; Jonnalagadda, S.R. Semantic Web and Mining in Healthcare. In Proceedings of the HEALTHCOM 2006 8th International Conference on e-Health Networking, Applications and Services, New Delhi, India, 17--19 August 2006; pp. 198--201.

\bibitem{B126-robotics-3060247}
Kara, N.; Dragoi, O.A. Reasoning with Contextual Data in Telehealth Applications. In Proceedings of the Third IEEE International Conference on Wireless and Mobile Computing, Networking and Communications (WiMob 2007), White Plains, NY, USA, 8--10 October 2007; p. 69.

\bibitem{B127-robotics-3060247}
Liu, W.; Daruna, A.; Patel, M.; Ramachandruni, K.; Chernova, S. A Survey of Semantic Reasoning Frameworks for Robotic Systems. \emph{Robot. Auton. Syst.} \textbf{\boldmath{2023}}, \emph{159}, 104294. [\href{https://doi.org/10.1016/j.robot.2022.104294}{CrossRef}]

\bibitem{B128-robotics-3060247}
Tang, X.; Zheng, Z.; Li, J.; Meng, F.; Zhu, S.-C.; Liang, Y.; Zhang, M. Large Language Models Are In-Context Semantic Reasoners Rather than Symbolic Reasoners. \emph{arXiv} \textbf{\boldmath{2023}}, arXiv:2305.14825.

\bibitem{B129-robotics-3060247}
Wen, Y.; Zhang, Y.; Huang, L.; Zhou, C.; Xiao, C.; Zhang, F.; Peng, X.; Zhan, W.; Sui, Z. Semantic Modelling of Ship Behavior in Harbor Based on Ontology and Dynamic Bayesian Network. \emph{ISPRS Int. J. Geo-Inf.} \textbf{\boldmath{2019}}, \emph{8}, 107. [\href{https://doi.org/10.3390/ijgi8030107}{CrossRef}]

\bibitem{B130-robotics-3060247}
Zheng, W.; Liu, X.; Ni, X.; Yin, L.; Yang, B. Improving Visual Reasoning Through Semantic Representation. \emph{IEEE Access} \textbf{\boldmath{2021}}, \emph{9}, 91476--91486. [\href{https://doi.org/10.1109/ACCESS.2021.3074937}{CrossRef}]

\bibitem{B131-robotics-3060247}
Pise, A.A.; Vadapalli, H.; Sanders, I. Relational Reasoning Using Neural Networks: A Survey. \emph{Int. J. Uncertain. Fuzziness Knowl.-Based Syst.} \textbf{\boldmath{2021}}, \emph{29}, 237--258. [\href{https://doi.org/10.1142/S0218488521400134}{CrossRef}]

\bibitem{B132-robotics-3060247}
Li, K.; Hopkins, A.K.; Bau, D.; Vi{\fontencoding{T5}\selectfont{\'e}}gas, F.; Pfister, H.; Wattenberg, M. Emergent World Representations: Exploring a Sequence Model Trained on a Synthetic Task. \emph{arXiv} \textbf{\boldmath{2023}}, arXiv:2210.13382.

\bibitem{B133-robotics-3060247}
Available online:  \url{https://everydayrobots.com} (accessed on 23 February 2024).

\bibitem{B134-robotics-3060247}
Peng, S.; Genova, K.; Jiang, C.; Tagliasacchi, A.; Pollefeys, M.; Funkhouser, T. OpenScene: 3D Scene Understanding with Open Vocabularies. In Proceedings of the 2023 IEEE/CVF Conference on Computer Vision and Pattern Recognition (CVPR), Vancouver, BC, Canada, 17--24 June 2023; pp. 815--824.

\bibitem{B135-robotics-3060247}
Boston Dynamics Spot\textbar{}Boston Dynamics. Available online:  \url{https://bostondynamics.com/products/spot/} (accessed on 24 February 2024).

\bibitem{B136-robotics-3060247}
Avgousti, S.; Christoforou, E.G.; Panayides, A.S.; Voskarides, S.; Novales, C.; Nouaille, L.; Pattichis, C.S.; Vieyres, P. Medical Telerobotic Systems: Current Status and Future Trends. \emph{Biomed. Eng. OnLine} \textbf{\boldmath{2016}}, \emph{15}, 96. [\href{https://doi.org/10.1186/s12938-016-0217-7}{CrossRef}] [\href{https://www.ncbi.nlm.nih.gov/pubmed/27520552}{PubMed}]

\bibitem{B137-robotics-3060247}
Yang, G.; Lv, H.; Zhang, Z.; Yang, L.; Deng, J.; You, S.; Du, J.; Yang, H. Keep Healthcare Workers Safe: Application of Teleoperated Robot in Isolation Ward for COVID-19 Prevention and Control. \emph{Chin. J. Mech. Eng.} \textbf{\boldmath{2020}}, \emph{33}, 47. [\href{https://doi.org/10.1186/s10033-020-00464-0}{CrossRef}]

\bibitem{B138-robotics-3060247}
Battaglia, E.; Boehm, J.; Zheng, Y.; Jamieson, A.R.; Gahan, J.; Fey, A.M. Rethinking Autonomous Surgery: Focusing on Enhancement Over Autonomy. \emph{Eur. Urol. Focus} \textbf{\boldmath{2021}}, \emph{7}, 696--705. [\href{https://doi.org/10.1016/j.euf.2021.06.009}{CrossRef}] [\href{https://www.ncbi.nlm.nih.gov/pubmed/34246619}{PubMed}]

\bibitem{B139-robotics-3060247}
Leonard, S.; Wu, K.L.; Kim, Y.; Krieger, A.; Kim, P.C.W. Smart Tissue Anastomosis Robot (STAR): A Vision-Guided Robotics System for Laparoscopic Suturing. \emph{IEEE Trans. Biomed. Engineering} \textbf{\boldmath{2014}}, \emph{61}, 1305--1317. Available online:  \url{https://ieeexplore.ieee.org/document/6720152} (accessed on 27 May 2024). [\href{https://doi.org/10.1109/TBME.2014.2302385}{CrossRef}] [\href{https://www.ncbi.nlm.nih.gov/pubmed/24658254}{PubMed}]

\bibitem{B140-robotics-3060247}
Takada, C.; Suzuki, T.; Afifi, A.; Nakaguchi, T. Hybrid Tracking and Matching Algorithm for Mosaicking Multiple Surgical Views. In \emph{Computer-Assisted and Robotic Endoscopy}; Peters, T., Yang, G.-Z., Navab, N., Mori, K., Luo, X., Reichl, T., McLeod, J., Eds.;  Springer International Publishing: Athens, Greece, 17 October 2017; pp. 24--35.

\bibitem{B141-robotics-3060247}
Afifi, A.; Takada, C.; Yoshimura, Y.; Nakaguchi, T. Real-Time Expanded Field-of-View for Minimally Invasive Surgery Using Multi-Camera Visual Simultaneous Localization and Mapping. \emph{Sensors} \textbf{\boldmath{2021}}, \emph{21}, 2106. [\href{https://doi.org/10.3390/s21062106}{CrossRef}] [\href{https://www.ncbi.nlm.nih.gov/pubmed/33802766}{PubMed}]

\bibitem{B142-robotics-3060247}
Lamini, C.; Benhlima, S.; Elbekri, A. Genetic Algorithm Based Approach for Autonomous Mobile Robot Path Planning. \emph{Procedia Comput. Sci.} \textbf{\boldmath{2018}}, \emph{127}, 180--189. [\href{https://doi.org/10.1016/j.procs.2018.01.113}{CrossRef}]

\bibitem{B143-robotics-3060247}
Xiang, D.; Lin, H.; Ouyang, J.; Huang, D. Combined Improved A* and Greedy Algorithm for Path Planning of Multi-Objective Mobile Robot\textbar{}Scientific Reports. \emph{Sci. Rep.} \textbf{\boldmath{2022}}, \emph{12}, 13273. Available online:  \url{https://www.nature.com/articles/s41598-022-17684-0} (accessed on 10 March 2024). [\href{https://doi.org/10.1038/s41598-022-17684-0}{CrossRef}] [\href{https://www.ncbi.nlm.nih.gov/pubmed/35918508}{PubMed}]

\bibitem{B144-robotics-3060247}
de Sales Guerra Tsuzuki, M.; de Castro Martins, T.; Takase, F.K. Robot Path Planning Using Simulated Annealing---ScienceDirect. \emph{IFAC Proc. Vol.} \textbf{\boldmath{2006}}, \emph{39}, 175--180. Available online:  \url{https://www.sciencedirect.com/science/article/pii/S1474667015358250} (accessed on 10 March 2024). [\href{https://doi.org/10.3182/20060517-3-FR-2903.00105}{CrossRef}]

\bibitem{B145-robotics-3060247}
End-to-End Deep Learning-Based Framework for Path Planning and Collision Checking: Bin Picking Application. Available online:  \url{https://ar5iv.labs.arxiv.org/html/2304.00119} (accessed on 3 March 2024).

\bibitem{B146-robotics-3060247}
Quinones-Ramirez, M.; Rios-Martinez, J.; Uc-Cetina, V. Robot Path Planning Using Deep Reinforcement Learning. \emph{arXiv} \textbf{\boldmath{2023}}, arXiv:2302.09120.

\bibitem{B147-robotics-3060247}
Nicola, F.; Fujimoto, Y.; Oboe, R. A LSTM Neural Network applied to Mobile Robots Path Planning. In Proceedings of the IEEE International Conference on Industrial Informatics (INDIN), Porto, Portugal, 18--20 July 2018; pp. 349--354.

\bibitem{B148-robotics-3060247}
Hjeij, M.; Vilks, A. A Brief History of Heuristics: How Did Research on Heuristics Evolve? \emph{Humanit. Soc. Sci. Commun.} \textbf{\boldmath{2023}}, \emph{10},~64. [\href{https://doi.org/10.1057/s41599-023-01542-z}{CrossRef}]

\bibitem{B149-robotics-3060247}
Kawaguchi, K.; Kaelbling, L.; Bengio, Y. Generalization in Deep Learning. In \emph{Mathematical Aspects of Deep Learning}; Cambridge University Press: Cambridge, UK, 2022.

\bibitem{B150-robotics-3060247}
On the Generalization Mystery in Deep Learning. Available online:  \url{https://ar5iv.labs.arxiv.org/html/2203.10036} (accessed on \mbox{10 March 2024}).

\bibitem{B151-robotics-3060247}
Understanding LLMs: A Comprehensive Overview from Training to Inference. Available online:  \url{https://ar5iv.labs.arxiv.org/html/2401.02038} (accessed on 3 May 2024).

\bibitem{B152-robotics-3060247}
ADaPT: As-Needed Decomposition and Planning with Language Models. Available online:  \url{https://arxiv.org/abs/2311.05772} (accessed on 10 March 2024).

\bibitem{B153-robotics-3060247}
Wang, J.; Wu, Z.; Li, Y.; Jiang, H.; Shu, P.; Shi, E.; Hu, H.; Ma, C.; Liu, Y.; Wang, X.; et~al. Large Language Models for Robotics: Opportunities, Challenges, and Perspectives. Available online:  \url{https://arxiv.org/abs/2401.04334v1} (accessed on 3 March 2024).

\bibitem{B154-robotics-3060247}
Tjomsland, J.; Kalkan, S.; Gunes, H. Mind Your Manners! A Dataset and A Continual Learning Approach for Assessing Social Appropriateness of Robot Actions. \emph{Front. Robot. AI} \textbf{\boldmath{2022}}, \emph{9}, 669420. [\href{https://doi.org/10.3389/frobt.2022.669420}{CrossRef}] [\href{https://www.ncbi.nlm.nih.gov/pubmed/35356061}{PubMed}]

\bibitem{B155-robotics-3060247}
Soh, H.; Pan, S.; Min, C.; Hsu, D. The Transfer of Human Trust in Robot Capabilities across Tasks. In Proceedings of the Robotics: Science and Systems XIV; Robotics: Science and Systems Foundation, Pittsburgh, PA, USA, 26 June 2018.

\bibitem{B156-robotics-3060247}
Soh, H.; Xie, Y.; Chen, M.; Hsu, D. Multi-Task Trust Transfer for Human-Robot Interaction. \emph{Sage J.} \textbf{\boldmath{2020}}, \emph{39}, 233--249. [\href{https://doi.org/10.1177/0278364919866905}{CrossRef}]

\bibitem{B157-robotics-3060247}
Sap, M.; Rashkin, H.; Chen, D.; LeBras, R.; Choi, Y. SocialIQA: Commonsense Reasoning about Social Interactions. \emph{arXiv} \textbf{\boldmath{2019}}, arXiv:1904.09728.

\bibitem{B158-robotics-3060247}
FRANKA RESEARCH 3. Available online:  \url{https://franka.de/} (accessed on 10 March 2024).

\bibitem{B159-robotics-3060247}
Puig, X.; Ra, K.; Boben, M.; Li, J.; Wang, T.; Fidler, S.; Torralba, A. VirtualHome: Simulating Household Activities Via Programs. In Proceedings of the 2018 IEEE/CVF Conference on Computer Vision and Pattern Recognition, Salt Lake City, UT, USA, 18--23 June 2018; pp. 8494--8502.

\bibitem{B160-robotics-3060247}
Renfrow, J. New Robot from Pillo Health, Black + Decker Offers in-Home Monitoring, Medication Dispensing\textbar{}Fierce Healthcare. Available online:  \url{https://www.fiercehealthcare.com/tech/new-robot-offers-home-monitoring-and-medication-dispensing} (accessed on 3 May 2024).

\bibitem{B161-robotics-3060247}
Speech-to-Text AI: Speech Recognition and Transcription\textbar{}Google Cloud. Available online:  \url{https://cloud.google.com/speech-to-text} (accessed on 9 March 2024).

\bibitem{B162-robotics-3060247}
Redmon, J.; Divvala, S.; Girshick, R.; Farhadi, A. You Only Look Once: Unified, Real-Time Object Detection. In Proceedings of the 2016 IEEE Conference on Computer Vision and Pattern Recognition (CVPR), Las Vegas, NV, USA, 27--30 June 2016; pp. 779--788.

\bibitem{B163-robotics-3060247}
Zhou, X.; Girdhar, R.; Joulin, A.; Krähenbühl, P.; Misra, I. Detecting Twenty-Thousand Classes Using Image-Level Supervision. In Proceedings of the Computer Vision--ECCV 2022, Tel Aviv, Israel, 23--27 October 2022; pp. 350--368.

\bibitem{B164-robotics-3060247}
Tobeta, M.; Sawada, Y.; Zheng, Z.; Takamuku, S.; Natori, N. E2Pose: Fully Convolutional Networks for End-to-End Multi-Person Pose Estimation. In Proceedings of the 2022 IEEE/RSJ International Conference on Intelligent Robots and Systems (IROS), Kyoto, Japan, 23--27 October 2022; pp. 532--537.

\bibitem{B165-robotics-3060247}
Machine, A. Android Alter3. Available online:  \url{http://alternativemachine.co.jp/en/project/alter3/} (accessed on 4 March 2024).

\bibitem{B166-robotics-3060247}
Shi, H.; Ball, L.; Thattai, G.; Zhang, D.; Hu, L.; Gao, Q.; Shakiah, S.; Gao, X.; Padmakumar, A.; Yang, B.; et~al. Alexa, Play with Robot: Introducing the First Alexa Prize SimBot Challenge on Embodied AI. \emph{arXiv} \textbf{\boldmath{2023}}, arXiv:2308.05221.

\bibitem{B167-robotics-3060247}
Gao, Q.; Thattai, G.; Shakiah, S.; Gao, X.; Pansare, S.; Sharma, V.; Sukhatme, G.; Shi, H.; Yang, B.; Zheng, D.; et~al. Alexa Arena: A User-Centric Interactive Platform for Embodied AI. \emph{arXiv} \textbf{\boldmath{2023}}. [\href{https://doi.org/10.48550/arXiv.2303.01586}{CrossRef}]

\bibitem{B168-robotics-3060247}
Ethics of Care in Technology-mediated Healthcare Practices: A Scoping Review-Ramvi-2023-Scandinavian Journal of Caring Sciences-Wiley Online Library. Available online:  \url{https://onlinelibrary.wiley.com/doi/full/10.1111/scs.13186} (accessed on 10 March 2024).

\bibitem{B169-robotics-3060247}
Ethical Implications of AI and Robotics in Healthcare: A Review-PMC. Available online:  \url{https://www.ncbi.nlm.nih.gov/pmc/articles/PMC10727550/} (accessed on 10 March 2024).

\bibitem{B170-robotics-3060247}
The Value and Importance of Health Information Privacy-Beyond the HIPAA Privacy Rule-NCBI Bookshelf. Available online:  \url{https://www.ncbi.nlm.nih.gov/books/NBK9579/} (accessed on 10 March 2024).

\bibitem{B171-robotics-3060247}
Harrer, S. Attention Is Not All You Need: The Complicated Case of Ethically Using Large Language Models in Healthcare and Medicine. \emph{eBioMedicine} \textbf{\boldmath{2023}}, \emph{90}, 104512. [\href{https://doi.org/10.1016/j.ebiom.2023.104512}{CrossRef}] [\href{https://www.ncbi.nlm.nih.gov/pubmed/36924620}{PubMed}]

\bibitem{B172-robotics-3060247}
Singhal, K.; Azizi, S.; Tu, T.; Mahdavi, S.S.; Wei, J.; Chung, H.W.; Scales, N.; Tanwani, A.; Cole-Lewis, H.; Pfohl, S.; et~al. Large Language Models Encode Clinical Knowledge. \emph{Nature} \textbf{\boldmath{2023}}, \emph{620}, 172--180. [\href{https://doi.org/10.1038/s41586-023-06291-2}{CrossRef}] [\href{https://www.ncbi.nlm.nih.gov/pubmed/37438534}{PubMed}]

\bibitem{B173-robotics-3060247}
Bender, E.M.; Gebru, T.; McMillan-Major, A.; Shmitchell, S. On the Dangers of Stochastic Parrots: Can Language Models Be Too Big? In Proceedings of the 2021 ACM Conference on Fairness, Accountability, and Transparency, Virtual Event, 3 March 2021; pp.~610--623.

\bibitem{B174-robotics-3060247}
Lareyre, F.; Raffort, J. Ethical Concerns Regarding the Use of Large Language Models in Healthcare. \emph{EJVES Vasc. Forum} \textbf{\boldmath{2023}}, \emph{61}, 1. [\href{https://doi.org/10.1016/j.ejvsvf.2023.10.003}{CrossRef}] [\href{https://www.ncbi.nlm.nih.gov/pubmed/38025830}{PubMed}]

\bibitem{B175-robotics-3060247}
Li, H.; Moon, J.T.; Purkayastha, S.; Celi, L.A.; Trivedi, H.; Gichoya, J.W. Ethics of Large Language Models in Medicine and Medical Research. \emph{Lancet Digit. Health} \textbf{\boldmath{2023}}, \emph{5}, e333--e335. [\href{https://doi.org/10.1016/S2589-7500(23)00083-3}{CrossRef}] [\href{https://www.ncbi.nlm.nih.gov/pubmed/37120418}{PubMed}]

\bibitem{B176-robotics-3060247}
Jeyaraman, M.; Balaji, S.; Jeyaraman, N.; Yadav, S. Unraveling the Ethical Enigma: Artificial Intelligence in Healthcare. \emph{Cureus} \textbf{\boldmath{2023}}, \emph{15}, e43262. [\href{https://doi.org/10.7759/cureus.43262}{CrossRef}] [\href{https://www.ncbi.nlm.nih.gov/pubmed/37692617}{PubMed}]

\bibitem{B177-robotics-3060247}
Wang, C.; Liu, S.; Yang, H.; Guo, J.; Wu, Y.; Liu, J. Ethical Considerations of Using ChatGPT in Health Care. \emph{J. Med. Internet Res.} \textbf{\boldmath{2023}}, \emph{25}, e48009. [\href{https://doi.org/10.2196/48009}{CrossRef}] [\href{https://www.ncbi.nlm.nih.gov/pubmed/37566454}{PubMed}]

\bibitem{B178-robotics-3060247}
Stahl, B.C.; Eke, D. The Ethics of ChatGPT--Exploring the Ethical Issues of an Emerging Technology. \emph{Int. J. Inf. Manag.} \textbf{\boldmath{2024}}, \emph{74}, 102700. [\href{https://doi.org/10.1016/j.ijinfomgt.2023.102700}{CrossRef}]

\bibitem{B179-robotics-3060247}
Oniani, D.; Hilsman, J.; Peng, Y.; Poropatich, R.K.; Pamplin, J.C.; Legault, G.L.; Wang, Y. Adopting and Expanding Ethical Principles for Generative Artificial Intelligence from Military to Healthcare. \emph{Npj Digit. Med.} \textbf{\boldmath{2023}}, \emph{6}, 225. [\href{https://doi.org/10.1038/s41746-023-00965-x}{CrossRef}] [\href{https://www.ncbi.nlm.nih.gov/pubmed/38042910}{PubMed}]

\bibitem{B180-robotics-3060247}
Clusmann, J.; Kolbinger, F.R.; Muti, H.S.; Carrero, Z.I.; Eckardt, J.-N.; Laleh, N.G.; Löffler, C.M.L.; Schwarzkopf, S.-C.; Unger, M.; Veldhuizen, G.P.; et~al. The Future Landscape of Large Language Models in Medicine. \emph{Commun. Med.} \textbf{\boldmath{2023}}, \emph{3}, 141. [\href{https://doi.org/10.1038/s43856-023-00370-1}{CrossRef}] [\href{https://www.ncbi.nlm.nih.gov/pubmed/37816837}{PubMed}]

\bibitem{B181-robotics-3060247}
Murphy, K.; Di Ruggiero, E.; Upshur, R.; Willison, D.J.; Malhotra, N.; Cai, J.C.; Malhotra, N.; Lui, V.; Gibson, J. Artificial Intelligence for Good Health: A Scoping Review of the Ethics Literature. \emph{BMC Med. Ethics} \textbf{\boldmath{2021}}, \emph{22}, 14. [\href{https://doi.org/10.1186/s12910-021-00577-8}{CrossRef}] [\href{https://www.ncbi.nlm.nih.gov/pubmed/33588803}{PubMed}]

\bibitem{B182-robotics-3060247}
Sharkey, A.; Sharkey, N. Granny and the Robots: Ethical Issues in Robot Care for the Elderly. \emph{Ethics Inf. Technol.} \textbf{\boldmath{2012}}, \emph{14}, 27--40. [\href{https://doi.org/10.1007/s10676-010-9234-6}{CrossRef}]

\bibitem{B183-robotics-3060247}
Siqueira-Batista, R.; Souza, C.R.; Maia, P.M.; Siqueira, S.L. ROBOTIC SURGERY: BIOETHICAL ASPECTS. \emph{ABCD Arq. Bras. Cir. Dig. S{\fontencoding{T5}\selectfont{\~a}}o Paulo} \textbf{\boldmath{2016}}, \emph{29}, 287--290. [\href{https://doi.org/10.1590/0102-6720201600040018}{CrossRef}] [\href{https://www.ncbi.nlm.nih.gov/pubmed/28076489}{PubMed}]

\bibitem{B184-robotics-3060247}
O’Brolch{\fontencoding{T5}\selectfont{\'a}}in, F. Robots and People with Dementia: Unintended Consequences and Moral Hazard. \emph{Nurs. Ethics} \textbf{\boldmath{2019}}, \emph{26}, 962--972. [\href{https://doi.org/10.1177/0969733017742960}{CrossRef}] [\href{https://www.ncbi.nlm.nih.gov/pubmed/29262739}{PubMed}]

\bibitem{B185-robotics-3060247}
House of Lords. AI in the UK: Ready, Willing and Able. 2017. Available online:  \url{https://publications.parliament.uk/pa/ld201719/ldselect/ldai/100/100.pdf} (accessed on 10 March 2024).

\bibitem{B186-robotics-3060247}
Decker, M. Caregiving Robots and Ethical Reflection: The Perspective of Interdisciplinary Technology Assessment. \emph{AI Soc.} \textbf{\boldmath{2008}}, \emph{22}, 315--330. [\href{https://doi.org/10.1007/s00146-007-0151-0}{CrossRef}]

\bibitem{B187-robotics-3060247}
Coeckelbergh, M.; Pop, C.; Simut, R.; Peca, A.; Pintea, S.; David, D.; Vanderborght, B. A Survey of Expectations About the Role of Robots in Robot-Assisted Therapy for Children with ASD: Ethical Acceptability, Trust, Sociability, Appearance, and Attachment. \emph{Sci. Eng. Ethics} \textbf{\boldmath{2016}}, \emph{22}, 47--65. [\href{https://doi.org/10.1007/s11948-015-9649-x}{CrossRef}]

\bibitem{B188-robotics-3060247}
Feil-Seifer, D.; Matarić, M.J. Socially Assistive Robotics. \emph{IEEE Robot. Autom. Mag.} \textbf{\boldmath{2011}}, \emph{18}, 24--31. [\href{https://doi.org/10.1109/MRA.2010.940150}{CrossRef}]

\bibitem{B189-robotics-3060247}
Luxton, D.D. Recommendations for the Ethical Use and Design of Artificial Intelligent Care Providers. \emph{Artif. Intell. Med.} \textbf{\boldmath{2014}}, \emph{62}, 1--10. [\href{https://doi.org/10.1016/j.artmed.2014.06.004}{CrossRef}] [\href{https://www.ncbi.nlm.nih.gov/pubmed/25059820}{PubMed}]

\bibitem{B190-robotics-3060247}
Nielsen, S.; Langensiepen, S.; Madi, M.; Elissen, M.; Stephan, A.; Meyer, G. Implementing Ethical Aspects in the Development of a Robotic System for Nursing Care: A Qualitative Approach. \emph{BMC Nurs.} \textbf{\boldmath{2022}}, \emph{21}, 180. [\href{https://doi.org/10.1186/s12912-022-00959-2}{CrossRef}]

\bibitem{B191-robotics-3060247}
Yasuhara, Y. Expectations and Ethical Dilemmas Concerning Healthcare Communication Robots in Healthcare Settings: A Nurse’s Perspective. In \emph{Information Systems-Intelligent Information Processing Systems, Natural Language Processing, Affective Computing and Artificial Intelligence, and an Attempt to Build a Conversational Nursing Robot}; IntechOpen: London, UK, 2021; ISBN 978-1-83962-360-8.

\bibitem{B192-robotics-3060247}
Chatila, R.; Havens, J.C. The IEEE Global Initiative on Ethics of Autonomous and Intelligent Systems. In \emph{Robotics and Well-Being}; Aldinhas Ferreira, M.I., Silva Sequeira, J., Singh Virk, G., Tokhi, M.O., Kadar, E., Eds.;  Intelligent Systems, Control and Automation: Science and Engineering; Springer International Publishing: Cham, Switzerland, 2019; Volume 95, pp. 11--16. ISBN 978-3-030-12523-3.

\bibitem{B193-robotics-3060247}
The Toronto Declaration. Available online:  \url{https://www.torontodeclaration.org/declaration-text/english/} (accessed on \mbox{26 June 2024}).

\bibitem{B194-robotics-3060247}
AI Universal Guidelines--Thepublicvoice.Org. Available online:  \url{https://thepublicvoice.org/ai-universal-guidelines/} (accessed on 26 June 2024).

\bibitem{B195-robotics-3060247}
Chakraborty, A.; Karhade, M. Global AI Governance in Healthcare: A Cross-Jurisdictional Regulatory Analysis. Available online:  \url{https://arxiv.org/abs/2406.08695v1} (accessed on 24 June 2024).

\bibitem{B196-robotics-3060247}
Birhane, A.; Kasirzadeh, A.; Leslie, D.; Wachter, S. Science in the Age of Large Language Models. \emph{Nat. Rev. Phys.} \textbf{\boldmath{2023}}, \emph{5}, 277--280. [\href{https://doi.org/10.1038/s42254-023-00581-4}{CrossRef}]

\bibitem{B197-robotics-3060247}
Browning, J.; LeCun, Y. Language, Common Sense, and the Winograd Schema Challenge. \emph{Artif. Intell.} \textbf{\boldmath{2023}}, \emph{325}, 104031. [\href{https://doi.org/10.1016/j.artint.2023.104031}{CrossRef}]

\bibitem{B198-robotics-3060247}
Abdulsaheb, J.A.; Kadhim, D.J. Classical and Heuristic Approaches for Mobile Robot Path Planning: A Survey. \emph{Robotics} \textbf{\boldmath{2023}}, \emph{12}, 93. [\href{https://doi.org/10.3390/robotics12040093}{CrossRef}]

\bibitem{B199-robotics-3060247}
Müller, V.C. Ethics of Artificial Intelligence and Robotics. In \emph{The Stanford Encyclopedia of Philosophy}; Zalta, E.N., Nodelman, U., Eds.;  Metaphysics Research Lab, Stanford University: Stanford, CA, USA, 2023.

\bibitem{B200-robotics-3060247}
Clanahan, J.M.; Awad, M.M. How Does Robotic-Assisted Surgery Change OR Safety Culture? \emph{AMA J. Ethics} \textbf{\boldmath{2023}}, \emph{25}, 615--623. [\href{https://doi.org/10.1001/amajethics.2023.615}{CrossRef}]

\bibitem{B201-robotics-3060247}
Ullah, E.; Parwani, A.; Baig, M.M.; Singh, R. Challenges and Barriers of Using Large Language Models (LLM) Such as ChatGPT for Diagnostic Medicine with a Focus on Digital Pathology--A Recent Scoping Review. \emph{Diagn. Pathol.} \textbf{\boldmath{2024}}, \emph{19}, 43. [\href{https://doi.org/10.1186/s13000-024-01464-7}{CrossRef}] [\href{https://www.ncbi.nlm.nih.gov/pubmed/38414074}{PubMed}]

\bibitem{B202-robotics-3060247}
Lown, B.A.; Rosen, J.; Marttila, J. An Agenda For Improving Compassionate Care: A Survey Shows About Half Of Patients Say Such Care Is Missing. \emph{Health Aff.} \textbf{\boldmath{2011}}, \emph{30}, 1772--1778. [\href{https://doi.org/10.1377/hlthaff.2011.0539}{CrossRef}]

\bibitem{B203-robotics-3060247}
shanepeckham Getting Started with LLM Prompt Engineering. Available online:  \url{https://learn.microsoft.com/en-us/ai/playbook/technology-guidance/generative-ai/working-with-llms/prompt-engineering} (accessed on 28 May 2024).

\bibitem{B204-robotics-3060247}
Wang, L.; Chen, X.; Deng, X.; Wen, H.; You, M.; Liu, W.; Li, Q.; Li, J. Prompt Engineering in Consistency and Reliability with the Evidence-Based Guideline for LLMs. \emph{Npj Digit. Med.} \textbf{\boldmath{2024}}, \emph{7}, 1--9. [\href{https://doi.org/10.1038/s41746-024-01029-4}{CrossRef}] [\href{https://www.ncbi.nlm.nih.gov/pubmed/38378899}{PubMed}]

\bibitem{B205-robotics-3060247}
Marson, S.M.; Powell, R.M. Goffman and the Infantilization of Elderly Persons: A Theory in Development. \emph{J. Sociol. Soc. Welf.} \textbf{\boldmath{2014}}, \emph{41}, 143--158. [\href{https://doi.org/10.15453/0191-5096.3986}{CrossRef}]

\bibitem{B206-robotics-3060247}
Hendrycks, D.; Burns, C.; Basart, S.; Zou, A.; Mazeika, M.; Song, D.; Steinhardt, J. Measuring Massive Multitask Language Understanding. In Proceedings of the International Conference on Learning Representations, Virtual Event, 3--7 May 2021.

\bibitem{B207-robotics-3060247}
Naveed, H.; Khan, A.U.; Qiu, S.; Saqib, M.; Anwar, S.; Usman, M.; Akhtar, N.; Barnes, N.; Mian, A. A Comprehensive Overview of Large Language Models. \emph{arXiv} \textbf{\boldmath{2024}}, arXiv:2307.06435.

\bibitem{B208-robotics-3060247}
Busselle, R.; Reagan, J.; Pinkleton, B.; Jackson, K. Factors Affecting Internet Use in a Saturated-Access Population. \emph{Telemat. Inform.} \textbf{\boldmath{1999}}, \emph{16}, 45--58. [\href{https://doi.org/10.1016/S0736-5853(99)00018-0}{CrossRef}]

\bibitem{B209-robotics-3060247}
Ali, M.R.; Lawson, C.A.; Wood, A.M.; Khunti, K. Addressing Ethnic and Global Health Inequalities in the Era of Artificial Intelligence Healthcare Models: A Call for Responsible Implementation. \emph{J. R. Soc. Med.} \textbf{\boldmath{2023}}, \emph{116}, 260--262. [\href{https://doi.org/10.1177/01410768231187734}{CrossRef}] [\href{https://www.ncbi.nlm.nih.gov/pubmed/37467785}{PubMed}]

\bibitem{B210-robotics-3060247}
Johnmaeda Prompt Engineering with Semantic Kernel. Available online:  \url{https://learn.microsoft.com/en-us/semantic-kernel/prompts/} (accessed on 28 May 2024).

\bibitem{B211-robotics-3060247}
Das, B.C.; Amini, M.H.; Wu, Y. Security and Privacy Challenges of Large Language Models: A Survey. \emph{arXiv} \textbf{\boldmath{2024}}, arXiv:2402.00888.

\bibitem{B212-robotics-3060247}
Mireshghallah, N.; Kim, H.; Zhou, X.; Tsvetkov, Y.; Sap, M.; Shokri, R.; Choi, Y. Can LLMs Keep a Secret? Testing Privacy Implications of Language Models via Contextual Integrity Theory. In Proceedings of the The Twelfth International Conference on Learning Representations, Vienna, Austria, 7--11 May 2024.

\bibitem{B213-robotics-3060247}
OpenAI Privacy Policy. Available online:  \url{https://openai.com/policies/privacy-policy/} (accessed on 28 May 2024).

\bibitem{B214-robotics-3060247}
OpenAI How Your Data Is Used to Improve Model Performance\textbar{}OpenAI Help Center. Available online:  \url{https://help.openai.com/en/articles/5722486-how-your-data-is-used-to-improve-model-performance} (accessed on 13 May 2024).

\bibitem{B215-robotics-3060247}
Liu, L.; Ning, L. USER-LLM: Efficient LLM Contextualization with User Embeddings. Available online:  \url{http://research.google/blog/user-llm-efficient-llm-contextualization-with-user-embeddings/} (accessed on 28 May 2024).

\bibitem{B216-robotics-3060247}
Staab, R.; Vero, M.; Balunovi’c, M.; Vechev, M.T. Beyond Memorization: Violating Privacy Via Inference with Large Language Models. In Proceedings of the The Twelfth International Conference on Learning Representations, Vienna, Austria, \mbox{7--11 May 2024}.

\bibitem{B217-robotics-3060247}
Chen, Y.; Lent, H.; Bjerva, J. Text Embedding Inversion Security for Multilingual Language Models. \emph{arXiv} \textbf{\boldmath{2024}}, arXiv:2401.12192.

\bibitem{B218-robotics-3060247}
Zhang, Y.; Jia, R.; Pei, H.; Wang, W.; Li, B.; Song, D. The Secret Revealer: Generative Model-Inversion Attacks Against Deep Neural Networks. In Proceedings of the 2020 IEEE/CVF Conference on Computer Vision and Pattern Recognition (CVPR), Virtual Event, 13--19 June 2020; pp. 250--258.

\bibitem{B219-robotics-3060247}
Morris, J.X.; Zhao, W.; Chiu, J.T.; Shmatikov, V.; Rush, A.M. Language Model Inversion. In Proceedings of the The Twelfth International Conference on Learning Representations, Vienna, Austria, 7 May 2024.

\bibitem{B220-robotics-3060247}
Liu, Y.; Palmieri, L.; Koch, S.; Georgievski, I.; Aiello, M. DELTA: Decomposed Efficient Long-Term Robot Task Planning Using Large Language Models. \emph{arXiv} \textbf{\boldmath{2024}}, arXiv:2404.03275.

\bibitem{B221-robotics-3060247}
Touvron, H.; Lavril, T.; Izacard, G.; Martinet, X.; Lachaux, M.-A.; Lacroix, T.; Rozi{\fontencoding{T5}\selectfont{\`e}}re, B.; Goyal, N.; Hambro, E.; Azhar, F.; et~al. LLaMA: Open and Efficient Foundation Language Models. \emph{arXiv} \textbf{\boldmath{2023}}, arXiv:2302.13971.

\bibitem{B222-robotics-3060247}
BTGenBot: Behavior Tree Generation for Robotic Tasks with Lightweight LLMs. Available online:  \url{https://ar5iv.labs.arxiv.org/html/2403.12761} (accessed on 14 May 2024).

\bibitem{B223-robotics-3060247}
OpenAI OpenAI Platform. Available online:  \url{https://platform.openai.com} (accessed on 8 March 2024).

\bibitem{B224-robotics-3060247}
Montreuil, V.; Clodic, A.; Ransan, M.; Alami, R. Planning Human Centered Robot Activities. In Proceedings of the 2007 IEEE International Conference on Systems, Man and Cybernetics, Montreal, QC, Canada, 7--10 October 2007; pp. 2618--2623.

\bibitem{B225-robotics-3060247}
Son, T.C.; Pontelli, E.; Balduccini, M.; Schaub, T. Answer Set Planning: A Survey. \emph{Theory Pract. Log. Program.} \textbf{\boldmath{2023}}, \emph{23}, 226--298. [\href{https://doi.org/10.1017/S1471068422000072}{CrossRef}]

\bibitem{B226-robotics-3060247}
Gebser, M.; Kaminski, R.; Kaufmann, B.; Schaub, T. Clingo = ASP + Control: Preliminary Report. \emph{arXiv} \textbf{\boldmath{2014}}, arXiv:1405.3694.

\bibitem{B227-robotics-3060247}
Wang, J.; Shi, E.; Yu, S.; Wu, Z.; Ma, C.; Dai, H.; Yang, Q.; Kang, Y.; Wu, J.; Hu, H.; et~al. Prompt Engineering for Healthcare: Methodologies and Applications. \emph{arXiv} \textbf{\boldmath{2024}}, arXiv:2304.14670.

\bibitem{B228-robotics-3060247}
Xiao, G.; Tian, Y.; Chen, B.; Han, S.; Lewis, M. Efficient Streaming Language Models with Attention Sinks. In Proceedings of the International Conference on Learning Representations, Vienna, Austria, 7--11 May 2024.

\bibitem{B229-robotics-3060247}
Hooper, C.; Kim, S.; Mohammadzadeh, H.; Mahoney, M.W.; Shao, Y.S.; Keutzer, K.; Gholami, A. KVQuant: Towards 10 Million Context Length LLM Inference with KV Cache Quantization. \emph{arXiv} \textbf{\boldmath{2024}}, arXiv:2401.18079.

\bibitem{B230-robotics-3060247}
Papers with Code---HellaSwag Benchmark (Sentence Completion). Available online:  \url{https://paperswithcode.com/sota/sentence-completion-on-hellaswag} (accessed on 28 May 2024).

\bibitem{B231-robotics-3060247}
\textls[-25]{Minaee, S.; Mikolov, T.; Nikzad, N.; Chenaghlu, M.A.; Socher, R.; Amatriain, X.; Gao, J. Large Language Models: A Survey. \emph{arXiv} \textbf{\boldmath{2024}}.} [\href{https://doi.org/10.48550/arXiv.2402.06196}{CrossRef}]

\bibitem{B232-robotics-3060247}
Rights (OCR), O. for C. Summary of the HIPAA Privacy Rule. Available online:  \url{https://www.hhs.gov/hipaa/for-professionals/privacy/laws-regulations/index.html} (accessed on 20 May 2024).

\bibitem{B233-robotics-3060247}
General Data Protection Regulation (GDPR)--Legal Text. Available online:  \url{https://gdpr-info.eu/} (accessed on 20 May 2024).

\bibitem{B234-robotics-3060247}
Canada, O. of the P.C. of PIPEDA Requirements in Brief. Available online:  \url{https://www.priv.gc.ca/en/privacy-topics/privacy-laws-in-canada/the-personal-information-protection-and-electronic-documents-act-pipeda/pipeda_brief/} (accessed on \mbox{20 May 2024}).

\bibitem{B235-robotics-3060247}
Han, X.; You, Q.; Liu, Y.; Chen, W.; Zheng, H.; Mrini, K.; Lin, X.; Wang, Y.; Zhai, B.; Yuan, J. InfiMM-Eval: Complex Open-Ended Reasoning Evaluation for Multi-Modal Large Language Models. \emph{arXiv} \textbf{\boldmath{2023}}, arXiv:2311.11567.

\bibitem{B236-robotics-3060247}
Seo, G.; Park, S.; Lee, M. How to Calculate the Life Cycle of High-Risk Medical Devices for Patient Safety. \emph{Front. Public Health} \textbf{\boldmath{2022}}, \emph{10}, 989320. [\href{https://doi.org/10.3389/fpubh.2022.989320}{CrossRef}]

\bibitem{B237-robotics-3060247}
Javaid, M.; Estivill-Castro, V. Explanations from a Robotic Partner Build Trust on the Robot’s Decisions for Collaborative Human-Humanoid Interaction. \emph{Robotics} \textbf{\boldmath{2021}}, \emph{10}, 51. [\href{https://doi.org/10.3390/robotics10010051}{CrossRef}]

\bibitem{B238-robotics-3060247}
How Should AI Systems Behave, and Who Should Decide? Available online:  \url{https://openai.com/index/how-should-ai-systems-behave/} (accessed on 24 June 2024).

\bibitem{B239-robotics-3060247}
Altman, S.; Brockman, G.; Sutskever, I. Governance of Superintelligence. Available online:  \url{https://openai.com/index/governance-of-superintelligence/} (accessed on 24 June 2024).

\bibitem{B240-robotics-3060247}
Leike, J.; Sutskever, I. Introducing Superalignment. Available online:  \url{https://openai.com/index/introducing-superalignment/} (accessed on 24 June 2024).

\bibitem{B241-robotics-3060247}
Raval, V.; Shah, S. The Practical Aspect: Privacy Compliance---A Path to Increase Trust in Technology. \emph{ISACA} \textbf{\boldmath{2020}}, \emph{6}, 15--19.

\bibitem{B242-robotics-3060247}
Lin, T.-Y.; Maire, M.; Belongie, S.; Bourdev, L.; Girshick, R.; Hays, J.; Perona, P.; Ramanan, D.; Zitnick, C.L.; Doll{\fontencoding{T5}\selectfont{\'a}}r, P. Microsoft COCO: Common Objects in Context. In Proceedings of the European Conference on Computer Vision, Zurich, Switzerland, 6--12 September 2014; pp. 740--755.

\bibitem{B243-robotics-3060247}
Krishna, R.; Zhu, Y.; Groth, O.; Johnson, J.; Hata, K.; Kravitz, J.; Chen, S.; Kalantidis, Y.; Li, L.-J.; Shamma, D.A.; et~al. Visual Genome: Connecting Language and Vision Using Crowdsourced Dense Image Annotations. \emph{Int. J. Comput. Vis.} \textbf{\boldmath{2017}}, \emph{123}, 32--73. [\href{https://doi.org/10.1007/s11263-016-0981-7}{CrossRef}]

\bibitem{B244-robotics-3060247}
Su, Y.; Lan, T.; Li, H.; Xu, J.; Wang, Y.; Cai, D. PandaGPT: One Model To Instruction-Follow Them All. \emph{arXiv} \textbf{\boldmath{2023}}, arXiv:2305.16355.

\bibitem{B245-robotics-3060247}
Sunnybrook Hospital Patient and Visitor Recording Policy-Sunnybrook Hospital. Available online:  \url{https://sunnybrook.ca/content/?page=privacy-recording-policy} (accessed on 28 May 2024).

\bibitem{B246-robotics-3060247}
Bello, S.A.; Yu, S.; Wang, C. Review: Deep Learning on 3D Point Clouds. \emph{Remote. Sens.} \textbf{\boldmath{2020}}, \emph{12}, 1729. [\href{https://doi.org/10.3390/rs12111729}{CrossRef}]

\bibitem{B247-robotics-3060247}
Xu, R.; Wang, X.; Wang, T.; Chen, Y.; Pang, J.; Lin, D. PointLLM: Empowering Large Language Models to Understand Point Clouds. \emph{arXiv} \textbf{\boldmath{2023}}, arXiv:abs/2308.16911.

\bibitem{B248-robotics-3060247}
Robinson, F.; Nejat, G. A Deep Learning Human Activity Recognition Framework for Socially Assistive Robots to Support Reablement of Older Adults. In Proceedings of the 2023 IEEE International Conference on Robotics and Automation (ICRA), London, UK, 29 May--2 June 2023; IEEE: London, UK, 2023; pp. 6160--6167.

\bibitem{B249-robotics-3060247}
Meta Meta Llama 2. Available online:  \url{https://llama.meta.com/llama2/} (accessed on 28 May 2024).

\bibitem{B250-robotics-3060247}
Colossal-AI One Half-Day of Training Using a Few Hundred Dollars Yields Similar Results to Mainstream Large Models, Open-Source and Commercial-Free Domain-Specific LLM Solution. Available online:  \url{https://hpc-ai.com/blog/one-half-day-of-training-using-a-few-hundred-dollars-yields-similar-results-to-mainstream-large-models-open-source-and-commercial-free-domain-specific-llm-solution} (accessed on 28 May 2024).

\bibitem{B251-robotics-3060247}
Zhang, Z.; Zhao, J.; Zhang, Q.; Gui, T.; Huang, X. Unveiling Linguistic Regions in Large Language Models. \emph{arXiv} \textbf{\boldmath{2024}}, arXiv:2402.14700.

\bibitem{B252-robotics-3060247}
Wen-Yi, A.; Mimno, D. Hyperpolyglot LLMs: Cross-Lingual Interpretability in Token Embeddings. arXiv; 2023.

\bibitem{B253-robotics-3060247}
Mahowald, K.; Ivanova, A.A.; Blank, I.A.; Kanwisher, N.; Tenenbaum, J.B.; Fedorenko, E. Dissociating Language and Thought in Large Language Models. \emph{Trends Cogn. Sci.} \textbf{\boldmath{2024}}, \emph{28}, 517--540. [\href{https://doi.org/10.1016/j.tics.2024.01.011}{CrossRef}] [\href{https://www.ncbi.nlm.nih.gov/pubmed/38508911}{PubMed}]

\bibitem{B254-robotics-3060247}
Chang, K.; Xu, S.; Wang, C.; Luo, Y.; Xiao, T.; Zhu, J. Efficient Prompting Methods for Large Language Models: A Survey. \emph{arXiv} \textbf{\boldmath{2024}}, arXiv:2404.01077.

\bibitem{B255-robotics-3060247}
OpenAI Introducing GPTs. Available online:  \url{https://openai.com/index/introducing-gpts/} (accessed on 24 June 2024).

\bibitem{B256-robotics-3060247}
Choi, J.; Yun, J.; Jin, K.; Kim, Y. Multi-News+: Cost-Efficient Dataset Cleansing via LLM-Based Data Annotation. \emph{arXiv} \textbf{\boldmath{2024}}, arXiv:2404.09682.

\bibitem{B257-robotics-3060247}
Ishibashi, Y.; Shimodaira, H. Knowledge Sanitization of Large Language Models. \emph{arXiv} \textbf{\boldmath{2024}}, arXiv:2309.11852.

\bibitem{B258-robotics-3060247}
Faraboschi, P.; Giles, E.; Hotard, J.; Owczarek, K.; Wheeler, A. Reducing the Barriers to Entry for Foundation Model Training. \emph{arXiv} \textbf{\boldmath{2024}}, arXiv:2404.08811.

\bibitem{B259-robotics-3060247}
Guo, M.; Wang, Y.; Yang, Q.; Li, R.; Zhao, Y.; Li, C.; Zhu, M.; Cui, Y.; Jiang, X.; Sheng, S.; et~al. Normal Workflow and Key Strategies for Data Cleaning Toward Real-World Data: Viewpoint. \emph{Interact. J. Med. Res.} \textbf{\boldmath{2023}}, \emph{12}, e44310. [\href{https://doi.org/10.2196/44310}{CrossRef}]

\bibitem{B260-robotics-3060247}
Common Crawl---Overview. Available online:  \url{https://commoncrawl.org/overview} (accessed on 24 June 2024).

\bibitem{B261-robotics-3060247}
Chaudhari, S.; Aggarwal, P.; Murahari, V.; Rajpurohit, T.; Kalyan, A.; Narasimhan, K.; Deshpande, A.; da Silva, B.C. RLHF Deciphered: A Critical Analysis of Reinforcement Learning from Human Feedback for LLMs. \emph{arXiv} \textbf{\boldmath{2024}}, arXiv:2404.08555.

\bibitem{B262-robotics-3060247}
Dwork, C.; McSherry, F.; Nissim, K.; Smith, A. Calibrating Noise to Sensitivity in Private Data Analysis. In \emph{Theory of Cryptography, Proceedings of theThird Theory of Cryptography Conference, TCC 2006}; New York, NY, USA, 4--7 March 2006, Springer: Berlin/Heidelberg, Germany, 2006; pp. 265--284.

\bibitem{B263-robotics-3060247}
Lukas, N.; Salem, A.; Sim, R.; Tople, S.; Wutschitz, L.; Zanella-B{\fontencoding{T5}\selectfont{\'e}}guelin, S. Analyzing Leakage of Personally Identifiable Information in Language Models. In Proceedings of the 2023 IEEE Symposium on Security and Privacy (SP), Los Alamitos, CA, USA, 22--24 May 2023; pp. 346--363.

\bibitem{B264-robotics-3060247}
Lample, G.; Ballesteros, M.; Subramanian, S.; Kawakami, K.; Dyer, C. Neural Architectures for Named Entity Recognition. In Proceedings of the 2016 Conference of the North American Chapter of the Association for Computational Linguistics: Human Language Technologies, San Diego, CA, USA, 12--17 June 2016; 2016; pp. 260--270.

\bibitem{B265-robotics-3060247}
Yu, D.; Kairouz, P.; Oh, S.; Xu, Z. Privacy-Preserving Instructions for Aligning Large Language Models. In Proceedings of the 41st International Conference on Machine Learning, Vienna, Austria, 21--27 July 2024.

\bibitem{B266-robotics-3060247}
Li, X.; Tram{\fontencoding{T5}\selectfont{\`e}}r, F.; Liang, P.; Hashimoto, T. Large Language Models Can Be Strong Differentially Private Learners. In Proceedings of the International Conference on Learning Representations, Virtual Event, 25--29 April 2022.

\bibitem{B267-robotics-3060247}
Everything You Need to Know about the “Right to Be Forgotten”. Available online:  \url{https://gdpr.eu/right-to-be-forgotten/} (accessed on 24 June 2024).

\bibitem{B268-robotics-3060247}
Charity, F.H. RITA (Reminiscence/Rehabilitation \& Interactive Therapy Activities). Available online:  \url{https://www.nhsfife.org/fife-health-charity/what-weve-funded/rita-reminiscencerehabilitation-interactive-therapy-activities/} (accessed on 3 July 2024).

\bibitem{B269-robotics-3060247}
Gao, Y.; Xiong, Y.; Gao, X.; Jia, K.; Pan, J.; Bi, Y.; Dai, Y.; Sun, J.; Wang, M.; Wang, H. Retrieval-Augmented Generation for Large Language Models: A Survey. \emph{arXiv} \textbf{\boldmath{2024}}, arXiv:2312.10997.

\bibitem{B270-robotics-3060247}
OpenAI GPT-4V(Ision) System Card. Available online:  \url{https://openai.com/index/gpt-4v-system-card/} (accessed on \mbox{4 July 2024}).

\bibitem{B271-robotics-3060247}
Koh, W.Q.; Felding, S.A.; Budak, K.B.; Toomey, E.; Casey, D. Barriers and Facilitators to the Implementation of Social Robots for Older Adults and People with Dementia: A Scoping Review. \emph{BMC Geriatr.} \textbf{\boldmath{2021}}, \emph{21}, 351. [\href{https://doi.org/10.1186/s12877-021-02277-9}{CrossRef}]

\bibitem{B272-robotics-3060247}
OpenAI API. Available online:  \url{https://openai.com/index/openai-api/} (accessed on 4 July 2024).

\bibitem{B273-robotics-3060247}
YouTube Data API. Available online:  \url{https://developers.google.com/youtube/v3} (accessed on 4 July 2024).

\bibitem{B274-robotics-3060247}
Stockton, T. Organizations Fear Ontario’s Investment to Reduce Surgical Wait Times Will Endanger Patients Because of Nursing Shortage. Capital Current. 2021. Available online:  \url{https://capitalcurrent.ca/organizations-fear-ontarios-investment-to-reduce-surgical-wait-times-will-endanger-patients-because-of-nursing-shortage/} (accessed on 4 July 2024).

\bibitem{B275-robotics-3060247}
Driess, D.; Xia, F.; Sajjadi, M.S.M.; Lynch, C.; Chowdhery, A.; Ichter, B.; Wahid, A.; Tompson, J.; Vuong, Q.; Yu, T.; et~al. PaLM-E: An Embodied Multimodal Language Model. In Proceedings of the 40th International Conference on Machine Learning, Honolulu, HI, USA, 23--29 July 2023; p. 340.

\bibitem{B276-robotics-3060247}
Iovino, M.; Scukins, E.; Styrud, J.; Ögren, P.; Smith, C. A Survey of Behavior Trees in Robotics and AI. \emph{Robot. Auton. Syst.} \textbf{\boldmath{2022}}, \emph{154}, 104096. [\href{https://doi.org/10.1016/j.robot.2022.104096}{CrossRef}]

\bibitem{B277-robotics-3060247}
Cloud Computing Services. Available online:  \url{https://cloud.google.com/} (accessed on 5 July 2024).

\end{thebibliography}
\end{document}